\documentclass[11pt]{article}

\usepackage[final]{acl}

\usepackage{times}
\usepackage{latexsym}
\usepackage[T1]{fontenc}
\usepackage[utf8]{inputenc}
\usepackage{microtype}
\usepackage{inconsolata}

\usepackage{graphicx}
\usepackage{adjustbox}

\usepackage{amsmath}
\usepackage{booktabs}
\usepackage{multirow}
\usepackage{tabularx}
\usepackage{xltabular}
\usepackage{array}
\usepackage{ragged2e}
\usepackage{enumitem}

\usepackage[table]{xcolor}
\usepackage[most]{tcolorbox}
\usepackage{caption}
\usepackage{hyperref}
\usepackage{pdfpages}
\usepackage{placeins}

\definecolor{turnblue}{RGB}{225,239,250}
\definecolor{turncream}{RGB}{253,244,231}
\definecolor{linegray}{RGB}{190,190,190}

\newcolumntype{N}{>{\centering\arraybackslash\color{gray}}p{0.42cm}}
\newcolumntype{S}{>{\centering\arraybackslash}p{0.52cm}}
\newcolumntype{Y}{>{\raggedright\arraybackslash}X}

\newtcolorbox{promptbox}[2][]{
  colback=gray!5!white,
  colframe=gray!80!black,
  coltitle=black,
  fonttitle=\bfseries\sffamily,
  title={#2},
  boxrule=0.5pt,
  arc=3pt,
  left=8pt, right=8pt, top=8pt, bottom=8pt,
  breakable,
  #1
}

\newcommand{\system}{\textsc{DigitalCoach}}
\newcommand{\placeholder}[1]{\textit{\textcolor{gray}{[#1]}}}
\newcommand{\action}[1]{\textit{\textcolor{blue}{[#1]}}}
\newcommand{\duration}{28.1}
\newcommand{\avgduration}{22.5 min}
\newcommand{\minduration}{3 min 53 sec}
\newcommand{\maxduration}{1 hour 16 min 16 sec}
\newcommand{\turns}{22,752}
\newcommand{\sessions}{72}
\newcommand{\nevents}{39,609}
\newcommand{\nfiles}{36,724}

\newcolumntype{C}{>{\centering\arraybackslash}X}

\usepackage{pifont}
\usepackage{makecell}

\usepackage{xcolor}
\usepackage{pifont}
\newcommand{\cmark}{\textcolor{green!55!black}{\ding{51}}}
\newcommand{\xmark}{\textcolor{red!55!black}{\ding{55}}}

\usepackage{silence}
\newif\ifshownnew
\shownnewfalse       

\newcommand{\newtext}[1]{%
    \ifshownnew
        \textcolor{blue}{#1}%
    \else
        #1%
    \fi
}
\renewcommand{\paragraph}[1]{\vspace{0.25em}\noindent\textbf{#1}}

\makeatletter
\def\@secpenalty{-300}
\makeatother
\makeatletter
\renewcommand{\@afterheading}{%
  \@nobreakfalse
  \everypar{%
    \if@nobreak
      \@nobreakfalse
      \clubpenalty 50
    \else
      \clubpenalty \@clubpenalty
      \everypar{}%
    \fi}}
\makeatother

\newif\ifshowcomments
\showcommentsfalse  

\definecolor{meng}{RGB}{245, 129, 66}
\definecolor{amy}{RGB}{30, 130, 130}
\definecolor{david}{RGB}{180, 120, 20}

\ifshowcomments
  \newcommand{\meng}[1]{\textcolor{meng}{\textbf{Meng:} #1}}
  \newcommand{\amy}[1]{\textcolor{amy}{\textbf{Amy:} #1}}
  \newcommand{\anya}[1]{\textcolor{blue}{\textbf{Anya:} #1}}
  \newcommand{\tobi}[1]{\textcolor{blue}{\textbf{Tobi:} #1}}
  \newcommand{\david}[1]{\textcolor{david}{\textbf{David:} #1}}
  \newcommand{\as}[1]{\textcolor{green}{\textbf{Alane:} #1}}
\else
  \newcommand{\meng}[1]{}
  \newcommand{\amy}[1]{}
  \newcommand{\anya}[1]{}
  \newcommand{\tobi}[1]{}
  \newcommand{\david}[1]{}
  \newcommand{\as}[1]{}
\fi

\title{\system{}: Communication and Grounding Gaps in\\ Human and Agentic Computer Use Coaching}

\author{
\textbf{Meng Chen}$^{1}$,
\textbf{Anya Ji}$^{1}$,
\textbf{Tsung-Han Wu}$^{1}$,
\textbf{Tobias Maringgele}$^{2}$\thanks{Work done at UC Berkeley as a visiting student.}, \\
\textbf{David M. Chan}$^{1}$,
\textbf{Alane Suhr}$^{1}$\thanks{Equal supervision.},
\textbf{Amy Pavel}$^{1}$\footnotemark[2] \\
$^{1}$University of California, Berkeley \quad
$^{2}$Technical University of Munich
}

\begin{document}
\maketitle

\begin{abstract}
Agents are increasingly capable of automating software tasks, but can they teach humans how to use software themselves?
We introduce \textbf{\system{}}, a multimodal dataset of 72 human expert-novice computer use coaching sessions consisting of 22,752 dialogue turns grounded in 28.1 hours of screen and input event recordings across 5 software applications.
We then use \system{} to evaluate whether state-of-the-art models can teach humans how to use computers.
Our automated evaluation shows that models differ from humans in how they coach: models provide more direct instructions, but fewer explanations, error diagnoses and knowledge-check questions.  
When we fix the coaching method, models produce utterances that are similar to human references, but poorly grounded in visual context.
Our interactive evaluation confirms that model coaches cause learners to passively follow instructions without deeper engagement and fall short in visual grounding. \system{} lays a foundation for collaborative and proactive computer use coaching agents. \newtext{Data and code are available at \url{https://project-digital-coach.vercel.app}}.
\end{abstract}

\begin{figure}[t]
    \centering
    \includegraphics[width=\columnwidth]{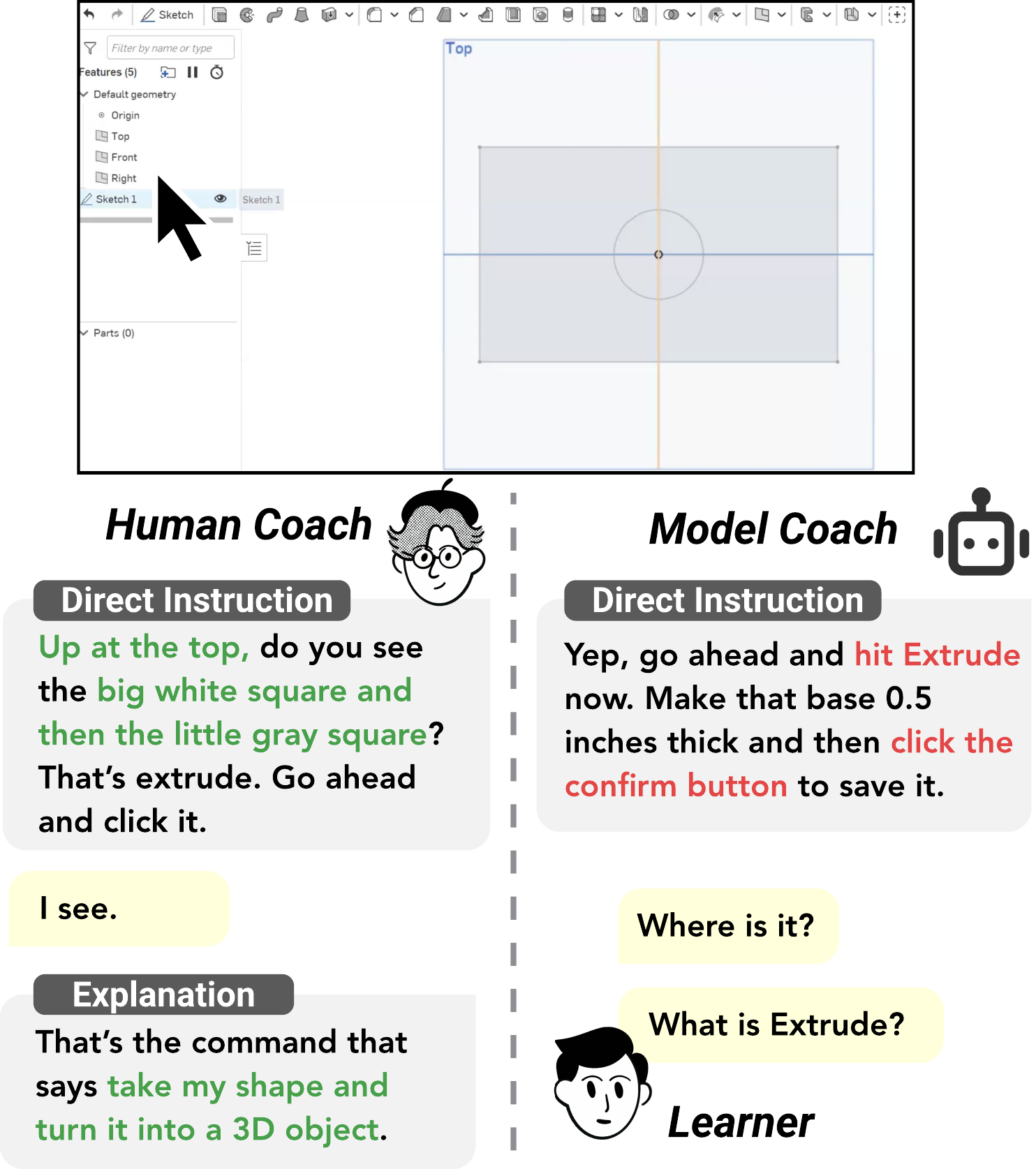}
    \caption{
     An example from \system{}. Human coach gives guidance grounded in user's screen and with explanations, while model coach gives instructions without explaining how or why. 
    }
    \label{fig:teaser}
\end{figure}

\section{Introduction}

\noindent Professional software tools for creativity  (\textit{e.g.}, Blender), engineering (\textit{e.g.}, OnShape), and analysis (\textit{e.g.}, Excel) are powerful but hard to learn. 
Agents now let novices use natural language instructions to easily execute creativity, engineering, and analysis tasks end-to-end without using the graphical user interfaces (GUIs) of the software~\cite{wang_genartist_2024,yin_vision-as-inverse-graphics_2026, Gao2024EmpoweringBDA}.
However, GUIs remain important to professionals as they afford expressivity, precision, and speed that cannot be achieved with natural language alone. 
For example, skilled Blender users can use the hotkey ``G'' then drag an object to position it, rather than using natural language to describe what object to move and where to move it.
For novices who want to develop such software expertise, the automation afforded by agents bypasses learning. 
Thus, we explore a new role for agents: teaching novices the skills they need to use these interfaces effectively.


Novices traditionally learn from expert demonstrations in webpage or video tutorials, but these materials are one-size-fits-all and require learners to manually align their own progress with the tutorial. To counter this, Human-Computer Interaction (HCI) research has prototyped systems~\cite{gmeiner_prototyping_2025, huh_vid2coach_2025} that transform tutorials into task assistants. But can current models generate adaptive, grounded guidance that supports skill acquisition beyond task completion?

We introduce \textbf{\system{}}, a dataset for (1) characterizing multimodal language-based interactions during human expert--novice computer use coaching and (2) analyzing how well models emulate human expert coaching. \system{} contains \sessions{} computer use coaching sessions of 40 participants spanning creativity, engineering, and productivity domains. \system{} has \turns{} turns dialogues grounded in \duration{} hours screen recordings, \nevents{} input events, and \nfiles{} file snapshots. To capture both the communicative and pedagogical functions of language use, we annotate utterances with dialogue acts and coaching methods.

We then evaluate 6 state-of-the-art models as computer use coaching agents using \system{}. Results show that model utterances differ distributionally from human utterances and are less lexically and semantically diverse. Compared to human coaches, models predominantly use direct instruction ($>$45\% of turns of all models vs. 30\% for humans). They provide less feedback on the screen state and ask fewer questions.
The best model (Gemini-3.1-Pro) can
generate plausible local guidance (CLAIR = 41.4) similar to human references when knowing the coaching method. Yet, current models all rely more on textual history than on the learner's screen state as models perform similarly without visual context, but their scores drop substantially without dialogue context. 

In real-time interaction, learners coached by models completed fewer milestones and retained fewer skills. We show that models still struggle to provide guidance grounded in evolving screen states and rarely offer reusable knowledge that helps learners accomplish similar tasks independently in the future. 
These findings suggest that effective computer use coaching agents require adaptive combinations of instruction and feedback based on the learner's evolving progress and opportunities for transferable skill development.

In summary, we contribute: 
\begin{itemize}[itemsep=0em, topsep=0.2em, parsep=0em, partopsep=0em]
    \item \system{}, a \duration{} hour human expert-novice GUI-grounded dialogue dataset with fine-grained annotations of dialogue acts and coaching methods. 
    \item Quantitative and qualitative characterizations distinguishing human and model coaching. 
    \item Expert rating and interactive human evaluations revealing communication and grounding gaps in real-time model coaching.
\end{itemize}
\section{Related Work}

\begin{table*}[t]
\centering
\small
\setlength{\tabcolsep}{4.2pt}
\renewcommand{\arraystretch}{1.12}
\resizebox{\linewidth}{!}{%
\begin{tabular}{p{3cm}p{1.8cm}c c c c cccccc}
\toprule
\multirow{2}{*}{\textbf{Dataset}} 
& \multirow{2}{*}{\textbf{Domain}} 
& \multirow{2}{*}{\textbf{Source}} 
& \multicolumn{2}{c}{\textbf{Setup}} 
& \multirow{2}{*}{\textbf{Duration}} 
& \multirow{2}{*}{\textbf{\# Turns}} 
& \multicolumn{5}{c}{\textbf{Data}} \\
\cmidrule(lr){4-5} \cmidrule(lr){8-12}
& & 
& \textbf{\# Ppl.} & \textbf{Relationship} 
& & 
& \textbf{Text} & \textbf{Audio} & \textbf{Video} & \textbf{Action} & \textbf{File} \\
\midrule
Portal Dialogue Corpus 
& Game 
& Real
& 2 & Collaborator
& 11.5h & 24.5k
& \cmark & \cmark & \cmark & \xmark & \cmark \\

MathDial
& Education 
& Semi-Synthetic
& 1 & Expert-Novice 
& -- & 28.3k
& \cmark & \xmark & \xmark & \xmark & \xmark \\

VideoCAD 
& Professional 
& Synthetic
& 0 & -- 
& 2.1kh & --
& \xmark & \xmark & \cmark & \cmark & \xmark \\

AssistGUI 
& Professional 
& Real
& 1 & -- 
& $<$8.3h & --
& \xmark & \xmark & \cmark & \cmark & \xmark \\

RealWebAssist 
& Everyday 
& Real
& 1 & -- 
& $<$6h & --
& \cmark & \cmark & \cmark & \xmark & \xmark \\

GUIDE 
& Professional 
& Real
& 1 & -- 
& 67.5h & --
& \cmark & \cmark & \cmark & \xmark & \xmark \\

MixAssist 
& Professional 
& Real
& 2 & Expert-Novice
& 7h & 431
& \cmark & \cmark & \xmark & \xmark & \xmark \\

\midrule
\system{} 
& Professional 
& Real
& 2 & Expert-Novice
& \duration{}h & 22.7k
& \cmark & \cmark & \cmark & \cmark & \cmark \\
\bottomrule
\end{tabular}%
}
\caption{Compared to related datasets~\cite{tomlin_characterizing_2025, macina_mathdial_2023, man_videocad_2025, gao_assistgui_2024, ye_realwebassist_2025, yang_guide_2026, clemens_mixassist_2025}, \system{} captures real human--human computer use coaching with much richer multimodal data.}
\label{tab:comparison_digitalcoach}
\end{table*}
\paragraph{Language Use in Situated Collaboration.}
Prior work has studied how agents ground and coordinate through language under partial observability and differing task knowledge~\cite{allen_trains_1995, allen2002human, lin_decision-oriented_2024, shaikh2024grounding, suhr_executing_2019, bara_mindcraft_2021, udagawa_annotated_2020}. 
More recent work moves toward richer settings, such as video games, household tasks, and collaborative drawing~\cite{zhang_proactive_2025, tomlin_characterizing_2025, padmakumar2022teach, howtodiv_2025, bhattacharyya2026can, kim2019codraw}.
However, no existing collaborative dialogue dataset is grounded in GUI workflows where people seek and provide help via multimodal communication (\emph{e.g.,} spatial referencing). Our dataset fills this gap by capturing expert-novice multimodal communication during GUI tasks.

\paragraph{Pedagogical and Instructional Dialogue.}
Unlike transactional dialogue systems that focus on task completion and intent tracking, such as making reservations~\cite{gasic_pomdp-based_2013} or tool use~\cite{yao_tau-bench_2024}, pedagogical dialogue emphasizes on how systems support learning through feedback, questioning, and scaffolding~\cite{sawyer_cognitive_2022}. 
Prior datasets such as MathDial~\cite{macina_mathdial_2023}, MathTutorBench~\cite{macina_mathtutorbench_2025}, and ConvoLearn~\cite{sharma_convolearn_2026} evaluate tutoring quality in domains including mathematics and geography, while MixAssist captures expert--amateur collaboration in music production workflows~\cite{clemens_mixassist_2025}. However, these datasets are either text-based or grounded in static content, and none capture instruction grounded in an evolving visual environment. 
\system{} fills this gap with multimodal expert-novice interactions grounded in real-time GUI state and user actions.

\paragraph{Collaborative GUI Agents.}
GUI agent research primarily studies whether agents can interpret interface state and execute actions to complete user goals. Benchmarks such as WebArena and OSWorld evaluate agents on web and desktop tasks using screenshots, natural language goals, and interface actions~\cite{zhou_webarena_2023, xie_osworld_2024}. More recent work extends to screen recordings that capture computer interactions over time~\cite{man_videocad_2025, lin_videogui_nodate, jang_videowebarena_2025, lin_showui_2025,zhang_showui-aloha_2026}.
However, existing GUI agent benchmarks primarily focus on autonomous task completion rather than collaboration with users during analytic and creative workflows where user agency matters~\cite{shen_completion_2025}. Recent work explores more proactive assistants that pause at decision points or reason about user intent~\cite{peng_morae_2025, huq_cowpilot_2025, yang_guide_2026}, but the language and dynamics of collaborative interaction remain underexplored. \system{} addresses this gap by characterizing human and model coaching.
\section{\system{} Dataset}

\noindent
We study the collaborative task of computer use coaching, where an expert and learner collaborate to teach the learner key skills in complex software applications, so that the learner can leave the interaction with expanded creative agency in the software.
In designing our study, we follow two principal design considerations: we aim to capture and characterize (a) \textit{multimodal dialogue}, with language use grounded in states (file snapshots), observation (screen recordings), and action (input events), and (b) \textit{learning outcomes}, using pre/post tasks measuring if learners retain learned skills.


\system{} contains \turns{} utterances across 72 human  expert-learner coaching sessions (Table~\ref{tab:comparison_digitalcoach}). Beyond dialogue, \system{} also captures rich multimodal activity traces including \duration{} hours of screen recordings, \nevents{} input events, and \nfiles{} file snapshots to support research on language grounding in action and perception. Appendix~\ref{sec:dataset-detail} contains additional details.

\begin{figure}[t]
    \centering
    \includegraphics[width=\columnwidth]{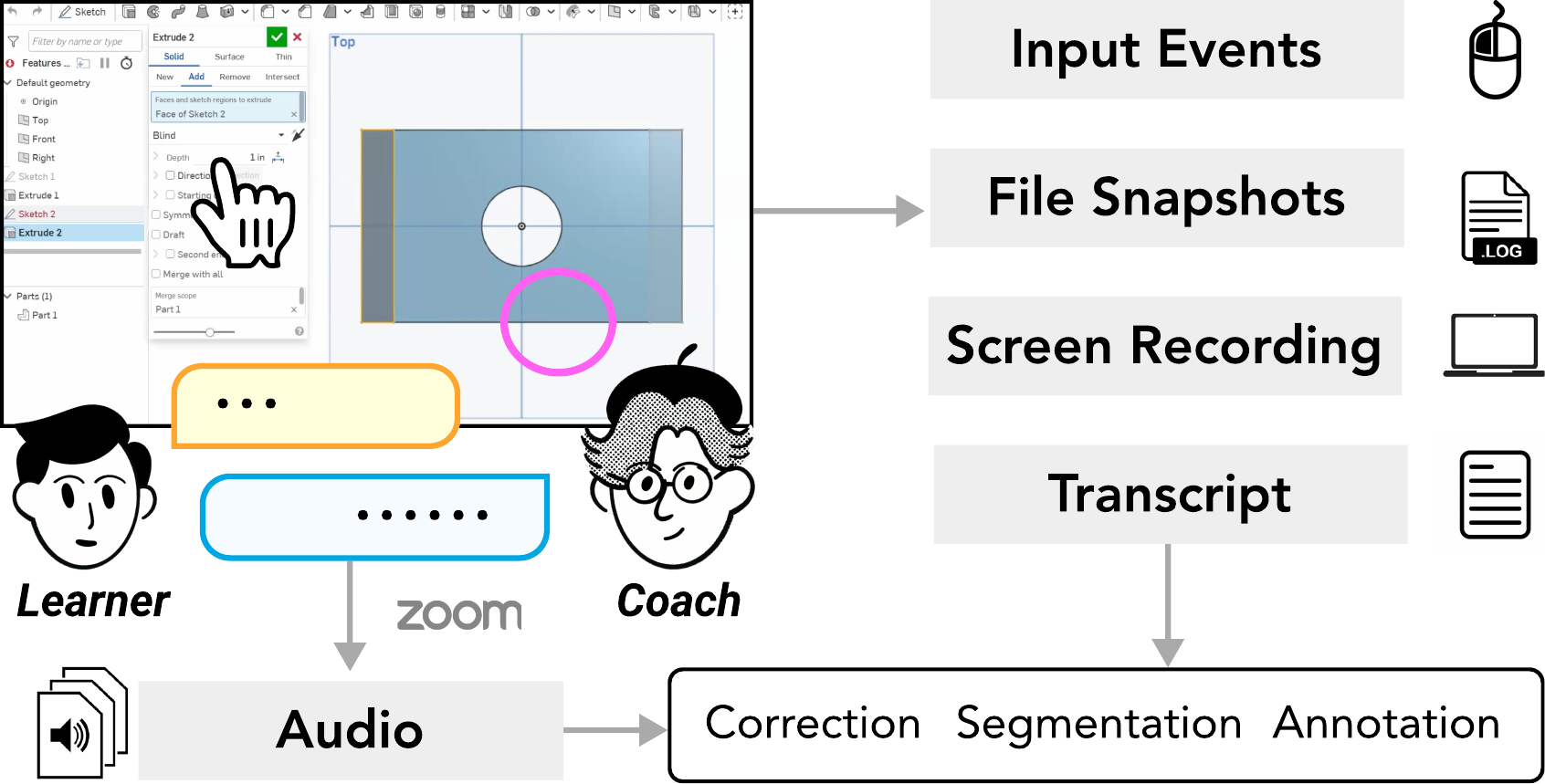}
    \caption{
     Collection setup, illustrated in the CAD software Onshape. A novice learner operates the software while an expert coach provides real-time verbal instructions and screen annotations.
    }
    \label{fig:study_setup}
\end{figure}

\subsection{Data Collection and Construction}

\paragraph{Tasks.} We selected 5 software applications spanning 3 domains: productivity (Excel), creativity (FL Studio, Blender, Figma), and engineering (Onshape). We sourced 18 tasks (\emph{e.g.,} Figure~\ref{fig:study_setup} shows making a saddle bracket in OnShape) from official and popular online tutorials, covering a range of lengths, difficulties, and modalities (text, audio, image, and 3D model, see Appendix~\ref{sec:tasks}).

\paragraph{Participants.}
We recruited a total of 20 English-speaking software-specific experts and 20 English-speaking novices through professional networks and Upwork,\footnote{\url{https://www.upwork.com/}} comprising 20 total pairs  (4 pairs $\times$ 5 applications). Novices had on average under one year ($\mu$ = 0.25, $\sigma$ = 0.44) of experience with and a strong interest in the target software; experts had at least 5 years of software experience ($\mu$ = 9.25, $\sigma$ = 5.25) and 6 months of coaching experience ($\mu$ =4.65, $\sigma$ = 3.31).
\footnote{Data collection and human evaluation studies in section~\ref{sec:human-eval} and~\ref{sec:interactive-eval} are approved by our institution’s IRB.} \newtext{See Appendix~\ref{sec:participants} for detailed information about participants' demographics and prior experiences.}

\paragraph{Collection Setup.} Each session paired one expert with one novice for approximately 3 hours, covering 3-4 tasks in a single software application. Sessions were conducted over Zoom, with learners sharing their screens while experts coached via verbal instruction, screen annotation, or remote control (Figure~\ref{fig:study_setup}). Before and after each tutorial task, learners also accomplished matched pre- and post-tasks targeting the same software skills as the tutorial but with variations (\emph{e.g.,} making a donut in Blender in tutorial v.s. making a caramel apple in pre/post task). Learners completed these independently (up to 15 min per task, Appendix~\ref{sec:protocol} describes the full study protocol).
\section{Dialogue Acts and Coaching Methods}
\paragraph{Dialogue Acts.}
To characterize communicative functions of language use in computer use coaching, we developed a dialogue act schema based on Dialog Act Markup in Several Layers~\cite[DAMSL;][]{core_coding_1997, stolcke_dialogue_2000}. We adapted DAMSL by merging similar categories (\emph{e.g.,} \textit{Command} and \textit{Suggestion}) and removing less relevant acts (\emph{e.g.,} \textit{Exclamation}). The final schema consists of information-seeking acts (\textit{Info Request}), information-providing acts (\textit{Answer, Inform, Opinion}), action-oriented acts (\textit{Action Directive}), and grounding acts (\textit{Backchannel}) that are common in a task-oriented multimodal coaching dialogue.

\paragraph{Coaching Methods.}
We further annotate each coach utterance to capture what types of guidance are delivered, based on prior work on cognitive apprenticeship theory~\cite{sawyer_cognitive_2022, ahn_answer_2026}, and video tutorial instruction~\cite{yang_beyond_2023}. Our schema separates procedural guidance (\textit{Direct Instruction, Plan}), knowledge (\textit{Explanation, Tip}), feedback (\textit{Confirmation, Diagnosis}), and learner-centered elicitation (\textit{Clarification, Reflection, Articulation, Exploration}). 

\subsection{Annotation and Classification} \noindent Three trained annotators independently labeled dialogue acts for a subset of utterances (200 total utterances; 40 randomly-sampled continuous turns per task), and coaching methods for 133 utterances from coach from the same sampled subset. The three annotators achieve substantial agreement \cite[Fleiss $\kappa$ = 0.79 for dialogue acts, $\kappa$ = 0.76 for coaching methods; ][]{fleiss_equivalence_1973}. We use an LLM classifier (GPT-5.4) to annotate dialogue acts and coaching methods across the entire dataset~\cite[mean Cohen's $\kappa$ = 0.69 for dialogue acts and 0.66 for coaching methods;][]{cohen1960coefficient}. We evaluate classifier performance against human annotations, achieving an F1 of 0.85 for dialogue acts and 0.83 for coaching methods. See Appendix~\ref{sec:annotation} for further details.

\subsection{Dataset Analysis}
\noindent Computer use coaching dialogue acts are task-oriented and are asymmetrically distributed across participants (Coach: 65.8\% vs. Learner: 34.2\%). Coaches mostly give \textit{Action Directives} (37\%) and \textit{Inform} (29\%), while learners primarily provide \textit{Backchannels} (50\%) or ask \textit{Info Requests} (13\%). Human coaches rely heavily on \textit{Direct Instruction} (31\%), but also use confirmation (17\%), explanation (20\%), planning (7\%), diagnosis (7\%), and learner-centered elicitation (7\%). 
Coaching method bigram sequences further show that \textit{Direct Instruction} often appear after \textit{Confirmation} ($N = 1026$, 5.71\%), \textit{Explanation} ($N = 1020$, 5.67\%), \textit{Plan} ($N = 537$, 2.99\%), and \textit{Diagnosis} ($N = 502$, 2.79\%).
\newtext{Following prior work~\cite{poh_show_2026}, we used frame-by-frame pixel extraction to detect the screen annotation color (\emph{i.e.}, pink). 4067 out of 14973 (27.2\%) coach utterances co-occur with a screen annotation.}
Results highlight that experts often first orient the learner to a goal, concept, or problem before giving an actionable next step. \newtext{They use screen annotation for UI reference and visual design illustration.}\footnote{See annotation result details in Appendix~\ref{sec:additional-annotation}.}

\section{Do Models Coach Like Humans?}
\label{sec:coach-eval}
\noindent We first analyze the distinctions between coaching behaviors in model and human coaches.
\begin{figure}[t]
    \centering
    \includegraphics[width=\linewidth]{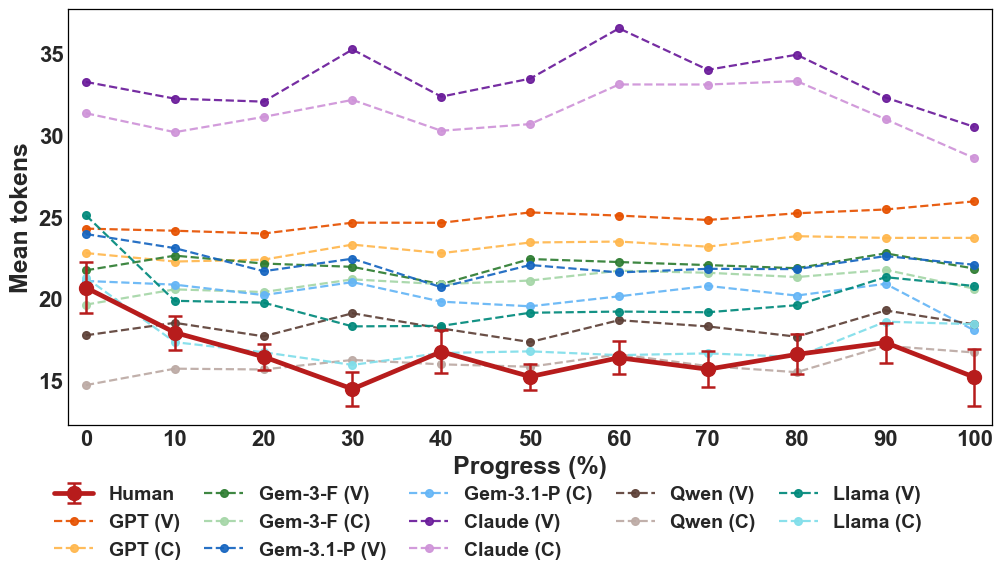}
    \caption{ 
    Mean token length across dialogue progress of sampled 18 sessions for human coaches and six models under Vanilla (V) and Coach (C) prompt conditions.
    }
    \label{fig:human-model-mean-token}
\end{figure}

\paragraph{Task Definition.} Formally, we cast the coach's task as given the preceding context $C_i$ consisting of the dialogue history and sampled screen frames within a time window $\Delta$, and a task description $T$ specifying the learning objectives, intermediate goals, and expected outcome, the model must generate a coaching utterance $\hat{y}_i$ for each human coach turn $y_i$ at time $t_i$. We hypothesize that an effective coaching agent should generate $\hat{y}_i$ that closely matches the human reference $y_i$.

\subsection{Setup}
\paragraph{Models.}
We selected 4 closed and 2 open models with multimodal input: GPT-5.4~\cite{openai_introducing_2026}, Gemini-3.1-Pro and Gemini-3-Flash~\cite{noauthor_gemini_nodate}, Claude-Sonnet-4.6~\cite{Claude4.6}, Qwen-3-VL-8B-Instruct~\cite{bai_qwen3-vl_2025}, and Llama-4-Scout-17B~\cite{noauthor_llama_nodate}. All models use default settings, generating one sentence per input given conversation and visual context at 1 fps (up to 30s).

\paragraph{Data.} \newtext{To preserve the temporal dependencies in coaching interactions,} we evaluate models on complete session trajectories. We randomly sampled one complete expert--learner session per task (18 tasks $\times$ 1 session) and ran model generation over each full trajectory, yielding 18 sessions as our evaluation set.

\paragraph{Conditions.} We compare two prompting conditions: \emph{(i)}~a \textbf{vanilla prompt} containing the task description and dialogue history, and \emph{(ii)}~a \textbf{coach prompt} that additionally provides our coaching method taxonomy and examples and asks it to choose the most appropriate ones (see Appendix~\ref{sec:prompts}). 

\subsection{Results}
\noindent Figure~\ref{fig:human-model-mean-token} shows that models generate longer utterances and do not shorten over the course of the dialogue as human coaches do~\cite{hua_talk_2024}. We therefore studied whether this added verbosity reflects diverse and richer coaching behavior or not.
\begin{figure}[t]
    \centering
\includegraphics[width=\columnwidth]{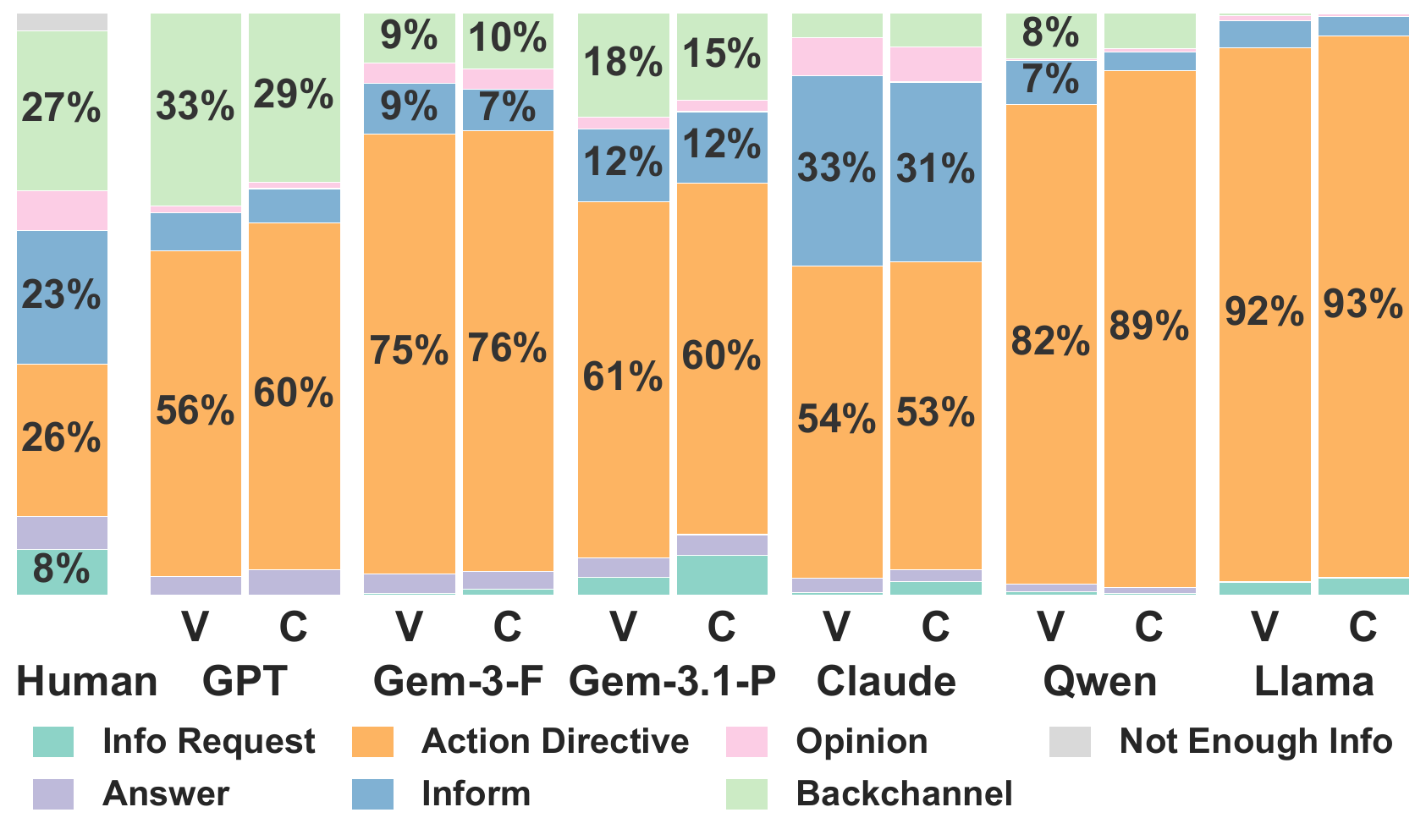}
    \caption{
Dialogue act distributions of sampled 18 sessions for human coaches and six models under Vanilla (V) and Coach (C) prompt conditions.}
    \label{fig:human-model-dialogue-acts}
\end{figure}

\begin{figure}[t]
    \centering
    \includegraphics[width=\columnwidth]{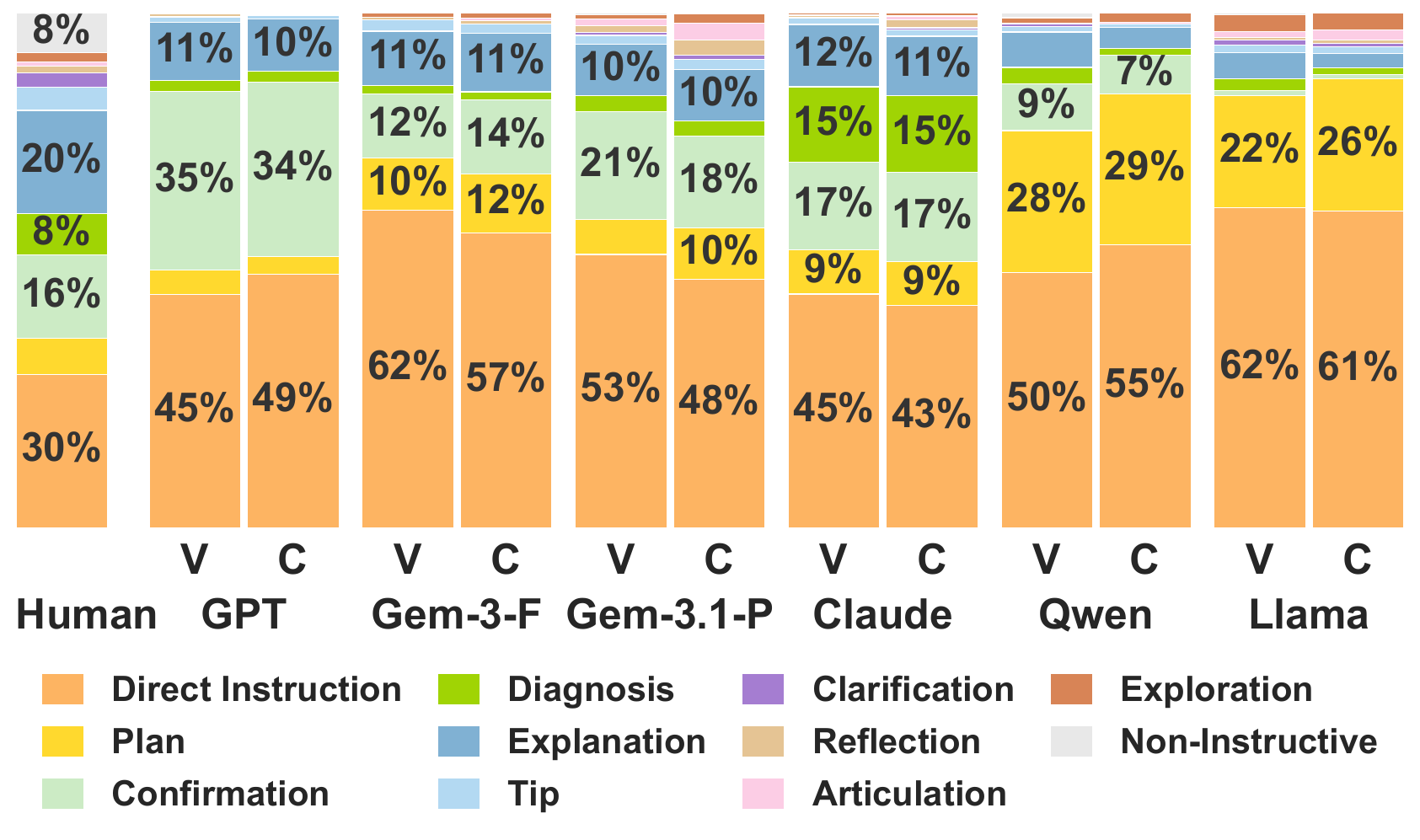}
    \caption{
Coaching method distributions of sampled utterances for human coaches and six models under Vanilla (V) and Coach (C) prompt conditions.}
    \label{fig:human-model-coaching}
\end{figure}

\begin{figure}[t]
    \centering
    \includegraphics[width=\linewidth]{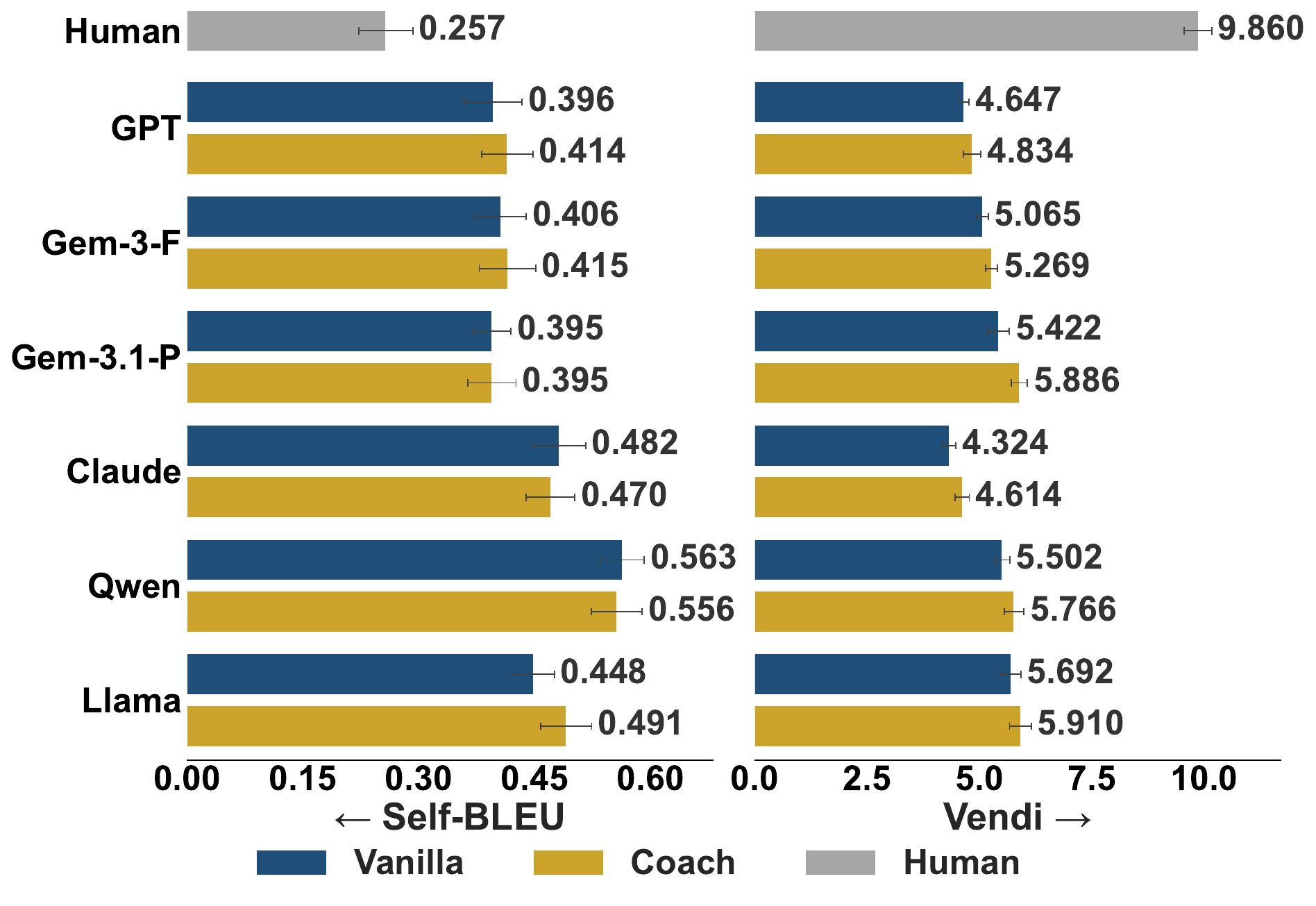}
    \caption{ 
    Lexical diversity (Self-BLEU, $\downarrow$) and semantic diversity (Vendi Score, $\uparrow$) of coaching utterances for human coaches and six models under vanilla (blue) and coach (gold) prompt conditions.
    }
    \label{fig:pragmatic_diversity}
\end{figure}

\paragraph{Models predominantly give direct instruction.} Figure~\ref{fig:human-model-dialogue-acts} shows that, compared to humans, all models predominantly use \textit{Action Directives} (>53\% vs. 26\%) and rarely request information from learners except Gemini-3.1-Pro (Coach) (<6\% vs. 8\%). 
In terms of coaching methods, models predominantly use direct instruction (>45\% vs. 30\%) but rarely provide conceptual guidance such as \textit{Explanation}(<12\% vs. 20\%) and \textit{Tip} (<1\% vs. 6\%) and learner-centered methods (Figure~\ref{fig:pragmatic_diversity}). 
Llama-4-Scout and Qwen-3-VL-8B-Instruct are instruction-heavy, with over 80\% of their dialogue acts as \textit{Action Directives} and nearly all of their coaching methods as \textit{Direct Instruction} and \textit{Plan}. Gemini and Claude are more human, using relatively more \textit{Inform} acts and more \textit{Explanation} and \textit{Confirmation} methods. 
GPT-5.4 is the least diverse model, using only \textit{Direct Instruction}, \textit{Confirmation}, and \textit{Explanation} with no learner-centric elicitation.

\begin{table*}[t]
\centering
\small
\setlength{\tabcolsep}{4pt}
\renewcommand{\arraystretch}{1.08}
\resizebox{\textwidth}{!}{%
\begin{tabular}{lcccccccc}
\toprule
& & \multicolumn{2}{c}{\textbf{Modality}} 
& \multicolumn{3}{c}{\textbf{Context}} 
& \multicolumn{2}{c}{\textbf{Prompt}} \\
\cmidrule(lr){3-4}\cmidrule(lr){5-7}\cmidrule(lr){8-9}
\textbf{Model} 
& \textbf{Default} 
& \textbf{Text-Only} 
& \textbf{Image-Only} 
& \textbf{1s} 
& \textbf{10s} 
& \textbf{60s} 
& \textbf{Vanilla} 
& \textbf{Coach} \\
\midrule
\textbf{GPT-5.4}      
& 32.4$_{\pm \text{1.3}}$          
& 31.8$_{\pm \text{1.3}}$          
& 25.3$_{\pm \text{1.2}}$          
& 25.4$_{\pm \text{1.2}}$          
& 31.2$_{\pm \text{1.3}}$          
& 32.7$_{\pm \text{1.3}}$          
& 24.0$_{\pm \text{1.2}}$          
& 23.8$_{\pm \text{1.2}}$ \\

\textbf{Gemini-3-Flash}     
& \underline{33.9$_{\pm \text{1.4}}$}          
& 32.7$_{\pm \text{1.4}}$          
& \underline{27.2$_{\pm \text{1.3}}$}          
& \underline{28.4$_{\pm \text{1.3}}$}          
& \underline{32.9$_{\pm \text{1.4}}$}          
& \underline{34.6$_{\pm \text{1.4}}$}          
& \underline{25.2$_{\pm \text{1.3}}$}          
& \underline{25.0$_{\pm \text{1.3}}$} \\

\textbf{Gemini-3.1-Pro}     
& \textbf{\boldmath 41.4$_{\pm \text{1.5}}$} 
& \textbf{\boldmath 41.9$_{\pm \text{1.5}}$} 
& \textbf{\boldmath 30.0$_{\pm \text{1.4}}$} 
& \textbf{\boldmath 33.7$_{\pm \text{1.4}}$} 
& \textbf{\boldmath 39.3$_{\pm \text{1.5}}$} 
& \textbf{\boldmath 41.6$_{\pm \text{1.5}}$} 
& \textbf{\boldmath 30.6$_{\pm \text{1.4}}$} 
& \textbf{\boldmath 31.0$_{\pm \text{1.4}}$} \\

\textbf{Claude-Sonnet-4.6}  
& 32.3$_{\pm \text{1.3}}$          
& \underline{35.9$_{\pm \text{1.4}}$}          
& 23.7$_{\pm \text{1.2}}$          
& 25.6$_{\pm \text{1.2}}$          
& 31.4$_{\pm \text{1.3}}$          
& 32.3$_{\pm \text{1.4}}$          
& 22.4$_{\pm \text{1.2}}$          
& 24.1$_{\pm \text{1.2}}$ \\

\textbf{Qwen-3-VL-Instruct} 
& 27.0$_{\pm \text{1.2}}$          
& 28.3$_{\pm \text{1.2}}$          
& 20.2$_{\pm \text{1.0}}$          
& 23.0$_{\pm \text{1.1}}$          
& 27.0$_{\pm \text{1.2}}$          
& 26.7$_{\pm \text{1.2}}$          
& 18.8$_{\pm \text{1.1}}$          
& 19.5$_{\pm \text{1.1}}$ \\

\textbf{Llama-4-Scout}      
& 19.5$_{\pm \text{1.0}}$          
& 24.4$_{\pm \text{1.2}}$          
& 16.3$_{\pm \text{0.8}}$          
& 18.6$_{\pm \text{1.0}}$          
& 20.2$_{\pm \text{1.0}}$          
& 19.7$_{\pm \text{1.0}}$          
& 14.8$_{\pm \text{0.8}}$          
& 14.2$_{\pm \text{0.9}}$ \\
\bottomrule
\end{tabular}%
}
\caption{CLAIR scores ($[1,100]$, $\uparrow$; mean $\pm$ 95\% CI)  for next coach utterance generation. The default setting uses text-visual input, 30s context, and oracle prompt. Other columns report ablations over input modality, context window, and prompt condition. CLAIR scores are judged by GPT-4.1 over 1,427 sampled pairs per run.}
\label{tab:clair_ablation}
\end{table*}

\paragraph{Model utterances are less diverse than human coaching.}
To measure whether models repeat the same responses or adapt across turns, we report lexical diversity ~\cite[Self-BLEU, $\downarrow$;][]{montahaei_jointly_2019} and semantic diversity~\cite[Vendi Score, $\uparrow$;][]{friedman_vendi_2023}. Figure~\ref{fig:pragmatic_diversity} shows that human coaches produce the most lexically and semantically diverse outputs, while all models fall substantially short on both metrics. 
This gap is consistent with the more diverse dialogue acts of human coaches, whereas models tend to generate more scripted procedural guidance, a pattern consistent with prior findings that language models often underrepresent the diversity of human communication~\cite{guo_benchmarking_2024,chan2022s}.

\paragraph{Prompting with coaching methods does not shift model behavior.} Model method distributions in coach prompt remain largely unchanged relative to the vanilla condition across both dialogue acts (Figure~\ref{fig:human-model-dialogue-acts}) and coaching methods (Figure~\ref{fig:human-model-coaching}). The gap between model and human coaching is not one of awareness but of capability: models default to instructional behavior even when explicitly guided otherwise. Closing this gap likely requires training interventions rather than prompting alone.

\begin{table}[t]
\centering
\small
\setlength{\tabcolsep}{4pt}
\renewcommand{\arraystretch}{1.08}
\begin{tabularx}{\linewidth}{Xcc}
\toprule
\textbf{Model} & \multicolumn{2}{c}{\textbf{MAUVE} (95\% CI)} \\
\cmidrule(lr){2-3}
& \textbf{Vanilla} & \textbf{Coach} \\
\midrule
\textbf{GPT-5.4} & 0.017$_{\pm \text{0.008}}$ & 0.018$_{\pm \text{0.009}}$ \\
\textbf{Gemini-3-Flash} & 0.042$_{\pm \text{0.016}}$ & 0.039$_{\pm \text{0.019}}$ \\
\textbf{Gemini-3.1-Pro} & \textbf{0.075$_{\pm \text{0.037}}$} & \textbf{0.118$_{\pm \text{0.049}}$} \\
\textbf{Claude-Sonnet-4.6} & 0.034$_{\pm \text{0.018}}$ & 0.036$_{\pm \text{0.016}}$ \\
\textbf{Qwen-3-VL-Instruct} & 0.038$_{\pm \text{0.018}}$ & 0.031$_{\pm \text{0.018}}$ \\
\textbf{Llama-4-Scout} & \underline{0.053$_{\pm \text{0.025}}$} & \underline{0.045$_{\pm \text{0.027}}$} \\
\bottomrule
\end{tabularx}
\caption{MAUVE scores ($[0,1]$, $\uparrow$; mean $\pm$ 95\% CI) between model and human distributions.}
\label{tab:mauve}
\end{table}
\section{Do Models Speak Like Humans?}
\label{sec:response-qual}

\noindent At the corpus level, model coaches do not speak like human coaches. Table~\ref{tab:mauve} compares each model's utterance distribution with the human coach distribution using MAUVE~\cite{pillutla_mauve_2021}. All model scores are low, even the best Gemini-3.1-Pro with the coach prompt (0.118 $_{\pm\text{0.049}}$). We further evaluate whether individual model utterances match the content and quality of human coach responses given the same context.

\subsection{Automatic Evaluation}
\label{sec:auto-eval}

\paragraph{Data.} \newtext{In Section~\ref{sec:coach-eval}, we found that models default to direct instruction, while coaching methods like explanation were more common in human utterances but rare in model ones. To evaluate models' ability to generate each coaching method when explicitly specified,} we randomly sampled 1{,}427 coach utterances across 5 software domains (300 per domain), balancing across coaching methods within each domain (targeting 30 per method). When a method had fewer than 30 utterances, all available examples were included. We excluded utterances tagged as \textit{Not Enough Information} or \textit{Non-Instructive}.

\paragraph{Conditions.}
By default, models receive both dialogue history and visual context sampled at 1 fps from the previous 30 seconds, and use an \textbf{oracle prompt} that reveals the coaching method used in the reference human coach utterance (see Appendix~\ref{sec:prompts}). We ablate three factors to compare model performance: \textbf{(1) Input modality}: We compare text-only context, visual-only context, and combined text-visual context, \textbf{(2) Context length}: We vary the context window across 1s, 10s, 30s, and 60s and \textbf{(3) Prompt}: We compare a vanilla prompt, a coach prompt, and the oracle prompt.

\paragraph{Metrics.} 
We compute content similarity between model utterances and human reference ones using CLAIR~\cite{chan_clair_2023}. We also report standard reference-based generation metrics, including BLEU, METEOR, ROUGE-L, and BERTScore, as well as CLAIR scores split by software application and coaching method in Appendix~\ref{sec:additional-eval}.

\paragraph{Results.}
Table~\ref{tab:clair_ablation} reports CLAIR scores of all models and conditions.
Gemini-3.1-Pro achieves the highest overall scores, while Llama-4-Scout lags considerably behind across all conditions.

\paragraph{Visual context is underutilized.}
Dropping visual input and using text alone causes only a modest score change for most models (e.g., Gemini-3.1-Pro: 41.4$_{\pm\text{1.5}}$ $\to$ 41.9$_{\pm\text{1.5}}$), yet providing visual input only causes a sharp drop across all models (e.g., Gemini-3.1-Pro: 41.4$_{\pm\text{1.5}}$ $\to$ 30.0$_{\pm\text{1.4}}$), suggesting models rely primarily on text.

\paragraph{More context improves utterance quality.}
Shortening the context window to 1s substantially degrades performance
across all models (\emph{e.g.}, Gemini-3.1-Pro: 41.4$_{\pm\text{1.5}}$ $\to$ 33.7$_{\pm\text{1.4}}$; Qwen-3-VL-Instruct: 27.0$_{\pm\text{1.2}}$ $\to$ 23.0$_{\pm\text{1.1}}$),  while extending to 60s yields marginal further gains.
Models benefit from observing a recent screen and conversations between learner and coach, but returns diminish beyond 10 seconds.


\subsection{Human Expert Evaluation}
\label{sec:human-eval}
\noindent \newtext{MAUVE and CLAIR measure similarity between model and human coaching utterances. Yet,} model-generated coaching utterances may differ from the human reference but still be valid and relevant. We therefore conduct a human expert evaluation to assess the quality of model-generated utterances.

\paragraph{Method.}
We randomly sampled a subset (N=100, 20 utterances per software tool) from the dataset in Section~\ref{sec:auto-eval} and recruited 15 expert evaluators (3 per software tool). Each evaluator judges 20 instances (2 per coaching method), and each instance was independently annotated by all 3 evaluators assigned to that software. For each instance, evaluators saw the task description, prior dialogue, and screen context. They ranked 3 candidate coaching utterances (human coach, Gemini-3.1-Pro Vanilla, and Gemini-3.1-Pro Oracle) and then rated each utterance along 3 dimensions using a three-point scale (1 = not, 2 = partially, 3 = very), drawn from prior VQA and proactive assistant work~\cite{zhang_proactive_2025, levinboim_quality_2021}: (1) \textbf{Relevance:} whether the utterance was grounded in visual and conversational content, (2) \textbf{Correctness:} how correct the utterance was and (3) \textbf{Naturalness:} whether the utterance sounded natural.

\begin{figure}[t]
    \centering
    \includegraphics[width=\linewidth]{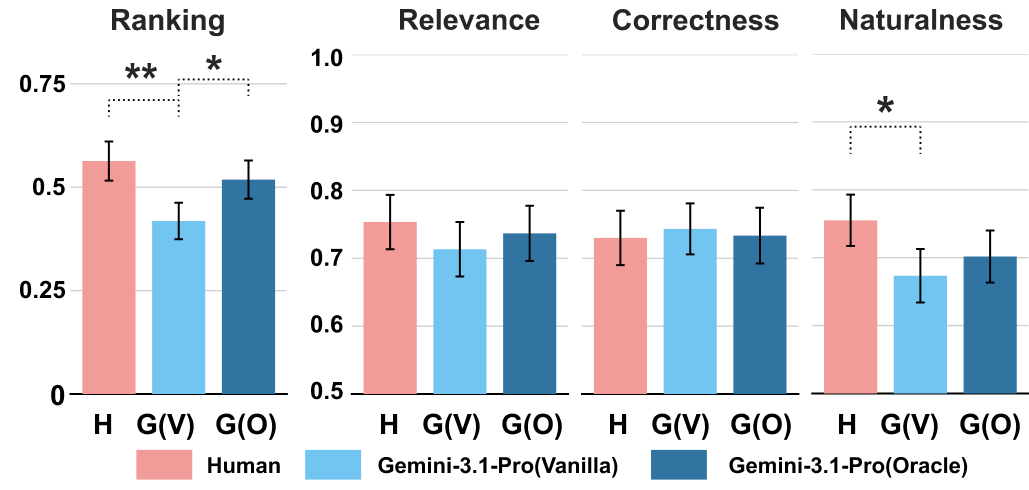}
    \caption{ 
    Mean normalized ranking  ($\uparrow$) and ratings ($\uparrow$) for human (H), Gemini-3.1-Pro Vanilla prompt (G(V)), and Oracle prompt (G(O)) coaching utterances (* = $p < 0.05$, ** = $p < 0.01$, Friedman test followed by pairwise Wilcoxon signed-rank tests, Bonferroni correction).}
    \label{fig:expert-eval}
\end{figure}

\paragraph{Results.} Figure~\ref{fig:expert-eval} shows human evaluators ranked human and Oracle conditions significantly higher than Vanilla ($p<.01$ and $p<.05$ respectively), suggesting that knowing the target coaching methods improves preference. Models scored comparably to humans on Relevance and Correctness, while human utterances were rated significantly more natural than Vanilla ($p<.05$), implying models generate locally plausible but unnatural utterances. Expert evaluators noted that effective coaching should be concise, accurate, and grounded in the learner's current screen state. 
Coaches should give direct instruction when learners are stuck, while encouraging independent thinking/using prior knowledge when possible.
\section{How Do Models Interact with Humans? }
\label{sec:interactive-eval}
\noindent We have shown that models speak and coach differently from human coaches. \textit{But what are interaction dynamics and effectiveness when models coach novices to use software in real time?}

\paragraph{Method.} We conducted an interactive evaluation with 10 participants from professional networks (P1--P10, 2 per software application) with the same setup as our data collection, yielding 36 model--human coaching sessions. 
To balance inference time and quality, learners completed tasks with guidance from Gemini-3-Flash with the Vanilla prompt and a 10s context window at 1 fps.
\newtext{Gemini-3-Flash achieved the second-highest CLAIR score (Table~\ref{tab:clair_ablation}) while providing substantially lower response latency than Gemini-3.1-Pro. As multimodal models demonstrated poor performance in generating coordinates~\cite{park_r-vlm_2025}, we did not enable models to provide annotations.}
Learners can interact with models via asking questions or receiving the model's proactive instructions every 20 seconds (See Appendix~\ref{sec:ix-eval-setup} for system setup \newtext{and model selection rationale}).

\subsection{Learning Outcome}
\noindent Both human ($N=72$) and model ($N=36$) coaching produced significant improvement in matched pre/post tasks ($p<.001$, Wilcoxon signed-rank test), but human coaching was substantially more effective (Table~\ref{tab:learning-progress-human-model}). Human coaching yielded a mean gain of 54.75\% (33.49\%~$\to$~88.24\%), with 81.9\% of sessions improving and none declining. Model coaching produced a smaller mean gain of 31.67\% (13.33\%~$\to$~45.00\%), with only 58.3\% of sessions improving, 25\% showing no progress, and one declining.
Models were far less effective than humans in teaching learners to use software independently.

\newtext{We further compute Pearson correlations between the proportion of each coaching method and pre/post task learning gain, noting that these analyses are exploratory given the small sample sizes and that correlations do not imply causation. When pooling human and model sessions ($N$ = 30), \textit{Diagnosis} was positively correlated with learning gain ($r=$ 0.69, $p<$ .001), whereas \textit{Direct Instruction} was negatively correlated ($r=$ -0.48, $p=$ .007). State-checking coaching methods may be particularly beneficial for building effective coaching agents.}
Appendix~\ref{sec:add-learn-outcome-results} reports subjective ratings, pre/post task outcomes, qualitative feedback, and \newtext{learning outcome details for each participant}.

\begin{table}[t]
\centering
\small
\setlength{\tabcolsep}{3pt}
\renewcommand{\arraystretch}{1.08}
\begin{tabularx}{\linewidth}{Xcrrr}
\toprule
\textbf{Progress} 
& \textbf{Setup} 
& \textbf{Tutorial} 
& \textbf{Pre-Task} 
& \textbf{Post-Task} \\
\midrule
\multirow{2}{*}{\textbf{100\%}}
& H & 72 (100.0\%) & 13 (18.1\%) & 49 (68.1\%) \\
& M & 9 (25.0\%)   & 2 (5.6\%)   & 9 (25.0\%) \\
\midrule
\multirow{2}{*}{\textbf{50--100\%}}
& H & 0 (0.0\%)    & 7 (9.7\%)   & 15 (20.8\%) \\
& M & 7 (19.4\%)   & 2 (5.6\%)   & 6 (16.7\%) \\
\midrule
\multirow{2}{*}{\textbf{0--50\%}}
& H & 0 (0.0\%)    & 22 (30.6\%) & 8 (11.1\%) \\
& M & 15 (41.7\%)  & 8 (22.2\%)  & 12 (33.3\%) \\
\midrule
\multirow{2}{*}{\textbf{0\%}}
& H & 0 (0.0\%)    & 30 (41.7\%) & 0 (0.0\%) \\
& M & 5 (13.9\%)   & 24 (66.7\%) & 9 (25.0\%) \\
\toprule
\multirow{2}{*}{\textbf{Total}}
& H & 72 (100.0\%) & 72 (100.0\%) & 72 (100.0\%) \\
& M & 36 (100.0\%) & 36 (100.0\%) & 36 (100.0\%) \\
\bottomrule
\end{tabularx}
\caption{Progress outcomes in human (H; $N=72$) and model (M; $N=36$) coaching sessions.}
\label{tab:learning-progress-human-model}
\end{table}

\subsection{Language Use}
\begin{figure}[t]
    \centering
    \includegraphics[width=\linewidth]{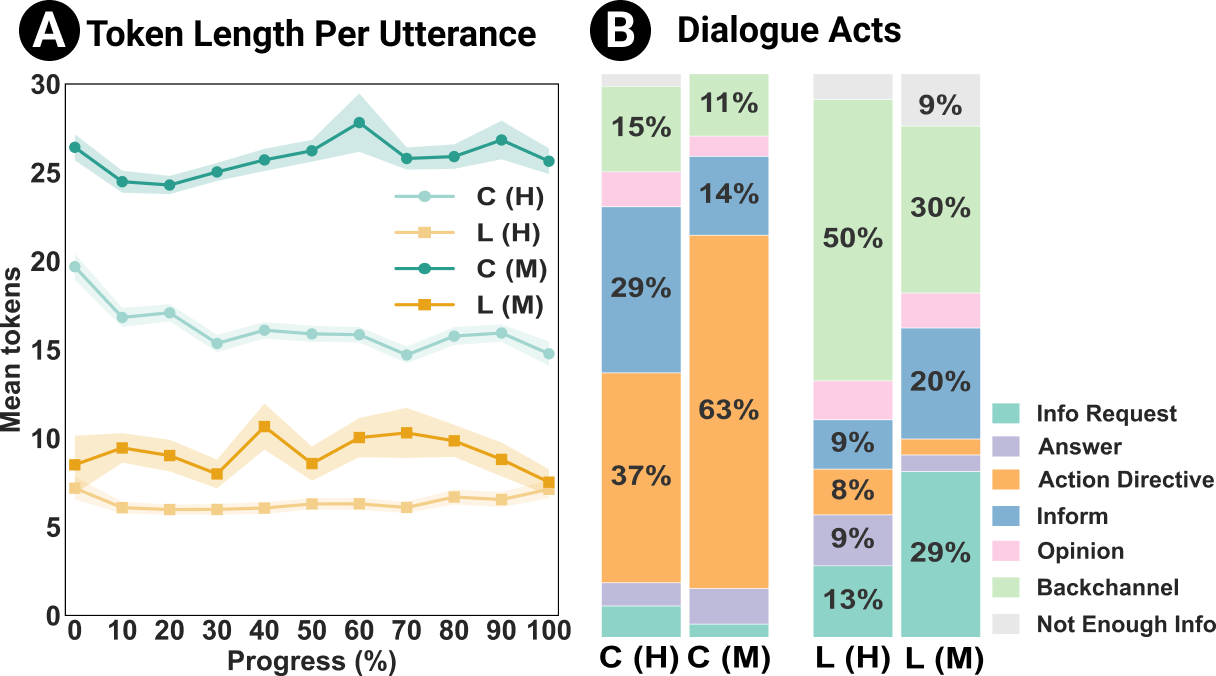}
    \caption{ 
    (A) Token length per utterance and (B) distribution of dialogue acts of coaches (C) and learners (L) in human (H; $N=72$) and model (M; $N=36$) coaching sessions.
    }
    \label{fig:interactive-info}
\end{figure}

\noindent The 36 model-human coaching sessions comprise 3,241 turns. Compared to human coaching sessions, models use more tokens per utterance, extensively use \textit{Action Directives} (63\% vs. 37\%), and rarely use \textit{Inform} (14\% vs. 29\%) that build learner understanding and \textit{Info Request} (2\% vs. 7\%) to check-in learning state. Human learners, in turn, ask more questions (29\% vs. 13\%) and provide more information (20\% vs. 9\%), primarily due to the lack of proactivity of models (Figure~\ref{fig:interactive-info}).

\subsection{Failure Case Analysis}
\begin{figure}[t]
    \centering
    \includegraphics[width=\linewidth]{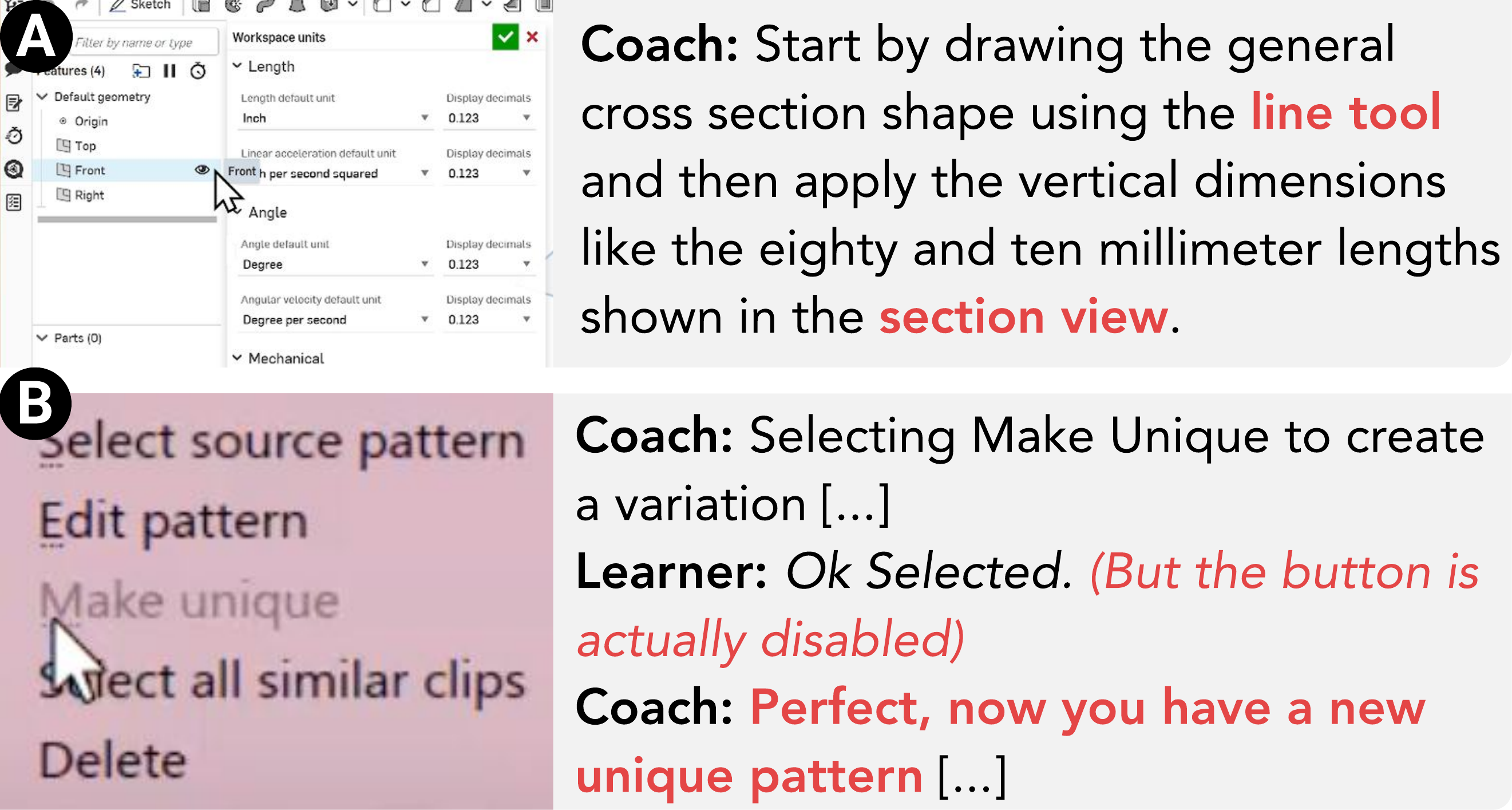}
    \caption{ 
    Representative (A) communication and (B) grounding failures in interactive evaluation.
    }
    \label{fig:model-outcome}
\end{figure}

\noindent Figure~\ref{fig:model-outcome} shows representative failure cases in model coaching sessions. 

\paragraph{Communication Gap.}
Models by default generated step-by-step instructions. While learners could replicate operations, they did not understand how or why (P4, P7, P9). Learners were still confused about the high-level plan and how to think about completing the task even after successful tutorials, leading to worse outcomes. 
Six participants found models too verbose and terminology-heavy, and novices found it hard to follow multiple steps and locate the relevant tools hidden in layers of menus (Figure~\ref{fig:model-outcome} A). P3 observed that they can follow simple steps at the beginning, but guidance became increasingly confusing as the task state became more complex. Thus, they wanted slower and more concise guidance on where and how to find the target tool. P8 also suggested that, like a human coach, models could point or describe nearby icons to help learners find the right tool.

\paragraph{Grounding Gap.}
Models fell short in tracking learners' current screen state, which required learners to explicitly describe progress (P1, P2, P5, P6) or ask what to do next (P3, P9, P10). P1 reported that they have to say \textit{``I'm done''} to make the model move on to the next step. This limitation of real-time multimodal understanding prevented the model from detecting and recovering when learners were stuck or had gone off track. For example, P2 was told to click a button that did not exist, yet the model kept repeating the same instruction.
Similarly, P5 could not find where to create a pivot sheet, but the model repeated the same long instruction rather than trying a different approach.
Models also did not point out mistakes unless explicitly asked. Even when they did respond, models sometimes hallucinated guidance based on misleading utterances rather than the screen state (Figure~\ref{fig:model-outcome} B).

\paragraph{Quantitative Characterization.}
\newtext{Our qualitative findings pointed to three grounding breakdowns that are specific to model coaches:
(1) \textbf{Insufficient Guidance (IG)} for locating the relevant UI elements;
(2) \textbf{Wrong Interpretation (WI)} of the current screen state;
(3) \textbf{Repeating Instructions (RI)} instead of diagnosing the issue or suggesting an alternative.
To quantify these failures, 2 researchers independently coded all 3,241 turns across the 10 human–model sessions and resolved ambiguities through discussion.
In total, we identified 294 breakdown episodes, including 79 IG (26.9\%), 122 WI (41.5\%), and 93 RI (31.6\%).
Screen annotation can potentially fix IG by highlighting the locations of UI elements. WI and RI
(73.1\%), by contrast, reflect failures of visual understanding and task error repair, which requires improvements in the model's visual reasoning capabilities.}
\section{Discussion and Conclusion}

\noindent We introduce \system{}, the first multimodal dataset of human expert-novice computer use coaching sessions. Our analysis reveals that current state-of-the-art multimodal models exhibit significant communication and grounding gaps compared to human coaches: they more frequently use direct instruction, and fail to adapt to the learner's evolving screen state. In an interactive evaluation, these gaps directly result in worse learning outcomes, with model-coached learners completing fewer milestones and retaining fewer skills.

\paragraph{Agency-Preserving Language Technology.}
Effective coaching balances direct instruction with learner agency. Human experts combine procedural guidance with explanation to build independence rather than replacing learner agency. For surface-level friction such as locating a menu item, direct instruction reduces unnecessary cognitive load, while for conceptual tasks such as design, agents should instead encourage reflection.

\paragraph{Proactive and Grounded Multimodal Agents.}
Our findings suggest that effective coaching agents must go beyond reactive question answering. Agents should proactively monitor learner progress and select pedagogical interventions accordingly. This requires tighter multimodal grounding, including real-time screen understanding and spatially situated feedback such as screen annotations to support the coaching that humans provide naturally.

\section*{Acknowledgments}
\noindent This research was supported by an Amazon Research Award, a Google gift, a Google ML and Systems Junior Faculty Award, a Technical AI Safety Research award from Coefficient Giving, and an NVIDIA Academic Grant Program award. \newtext{We thank Siddharth Chilamkur and Elaine Zhang for their help in the early stage of data annotation}. We also thank our study participants for their time and valuable contribution to this work.

\section*{Limitations}
\paragraph{Scale and Diversity of Data.} \system{} captures \duration{} hours of coaching sessions across 5 software domains and 40 participants. All sessions were collected on a Windows laptop to standardize the environment and reflect industry standards (\emph{e.g.,} Blender is more commonly used on Windows). \newtext{Our learners are students from an institution of higher education and likely grew up using technology so they might have more digital familiarity than the general population.} 
Future work could expand the dataset with more sessions, operating systems and software applications, and participants to capture richer interaction behaviors. 

\paragraph{Interactive Evaluation.} Evaluating computer use coaching agents remains an open challenge. \newtext{Our interactive evaluation tested one model with 10 participants. The model did not have access to the annotation tools available to human coaches. Future work can evaluate additional models (\emph{e.g.}, duplex models) and participant pools under matched interaction settings. In addition,} task completion alone does not fully capture learning. While our pre/post tasks provide an effective measure of learning, they cannot fully capture skill retention over time. Future benchmarks can therefore evaluate coaching along multiple dimensions (\emph{e.g.,} agency, task milestones, interaction traces, and pre/post tasks) and multiple domains (\emph{e.g.,} physical tasks). \system{} opens directions for building gym-style benchmarks for coaching agents.

\section*{Ethical Considerations}
\label{sec:ethics}
\paragraph{Human Data Collection and Evaluation} Our human evaluation study was approved by our institution’s Institutional Review Board (IRB). In our study, we ensured that all participants were compensated fairly for their time and contributions. The payment was determined based on the average market rate for such studies, participant expertise, and the complexity and duration of the tasks (\$20 / hour for learners and human evaluators; \$20 -- 60 / hour for coaches, depending on their experiences). \newtext{We release the de-identified dataset on Hugging Face and evaluation code on Github.}

\paragraph{Hallucinations in Language Models} Our work also uses LLMs to annotate dialogue acts and coaching methods. While we have shown that these approaches better align with humans, LLMs are not free from potential hallucinations and can lead to inaccurate results.

\paragraph{Automation vs. Augmentation.}
While LLM agents are increasingly framed as replacements for human labor, our work points to a different role for agents: not automating tasks, but supporting humans in developing new skills. However, this framing also requires caution. For example, coaching agents could be deployed as substitutes for human experts and thus reduce access to human mentorship and weaken the social and emotional support from human coaches. We therefore view coaching agents as a complement to, rather than a replacement for, human expertise.

\paragraph{Trust in Coaching Agents.}
Learners may place high trust in computer use coaching agents, especially when they lack the expertise to judge if its guidance is correct. Incorrect or overconfident instructions can mislead learners and reinforce misconceptions. We are aware that future design of computer use coaching agents should communicate uncertainty, support verification, and make it easy for learners to question or override their guidance.

\bibliography{custom}

\appendix
\clearpage

\section*{Appendix}
\label{sec:appendix}

\renewcommand{\theequation}{\thesection.\arabic{equation}}
\renewcommand{\thefigure}{\thesection.\arabic{figure}}
\renewcommand{\thetable}{\thesection.\arabic{table}}

\makeatletter
\@addtoreset{equation}{section}
\@addtoreset{figure}{section}
\@addtoreset{table}{section}
\makeatother

Our appendix is organized as follows: 
\begin{itemize}
    \item \autoref{sec:dataset-detail} provides detailed statistics and linguistic examples of the dataset.
    \item \autoref{sec:protocol} outlines the experimental platform setup and facilitation scripts.
    \item \autoref{sec:tasks} catalogs the specific software tasks across the five tested domains.
    \item \autoref{sec:participants} provides participant demographics and recruitment details.
    \item \autoref{sec:transcription} details the human-in-the-loop transcript processing workflow.
    \item \autoref{sec:annotation} defines the dialogue act and coaching methods codebooks.
    \item \autoref{sec:additional-eval} details model evaluation baselines and metrics.
    \item \autoref{sec:prompts} provides the full text prompts utilized for LLM pipelines.
    \item \autoref{sec:add-learn-outcome-results} provides quantitative and qualitative human learning outcomes.
    \item \autoref{app:artifact-license-use} discusses the license of produced artifacts.
    \item \autoref{app:package-settings} provides information on settings for used code packages.
    \item \autoref{app:consent} provides further details on the human subjects, consent, and data protection protocols.
    \item \autoref{app:ai} states our AI Use Disclosure.
    
\end{itemize}

\section{Dataset Details}
\label{sec:dataset-detail}
\noindent To support further research, the \system{} dataset, including the screen recording, audio transcripts, mouse and keyboard events, and the file snapshots, will be made publicly available upon publication. The \system{} dataset will be accessible on Hugging Face and the evaluation code on GitHub.

\begin{figure}[t]
    \centering
    \includegraphics[width=\columnwidth]{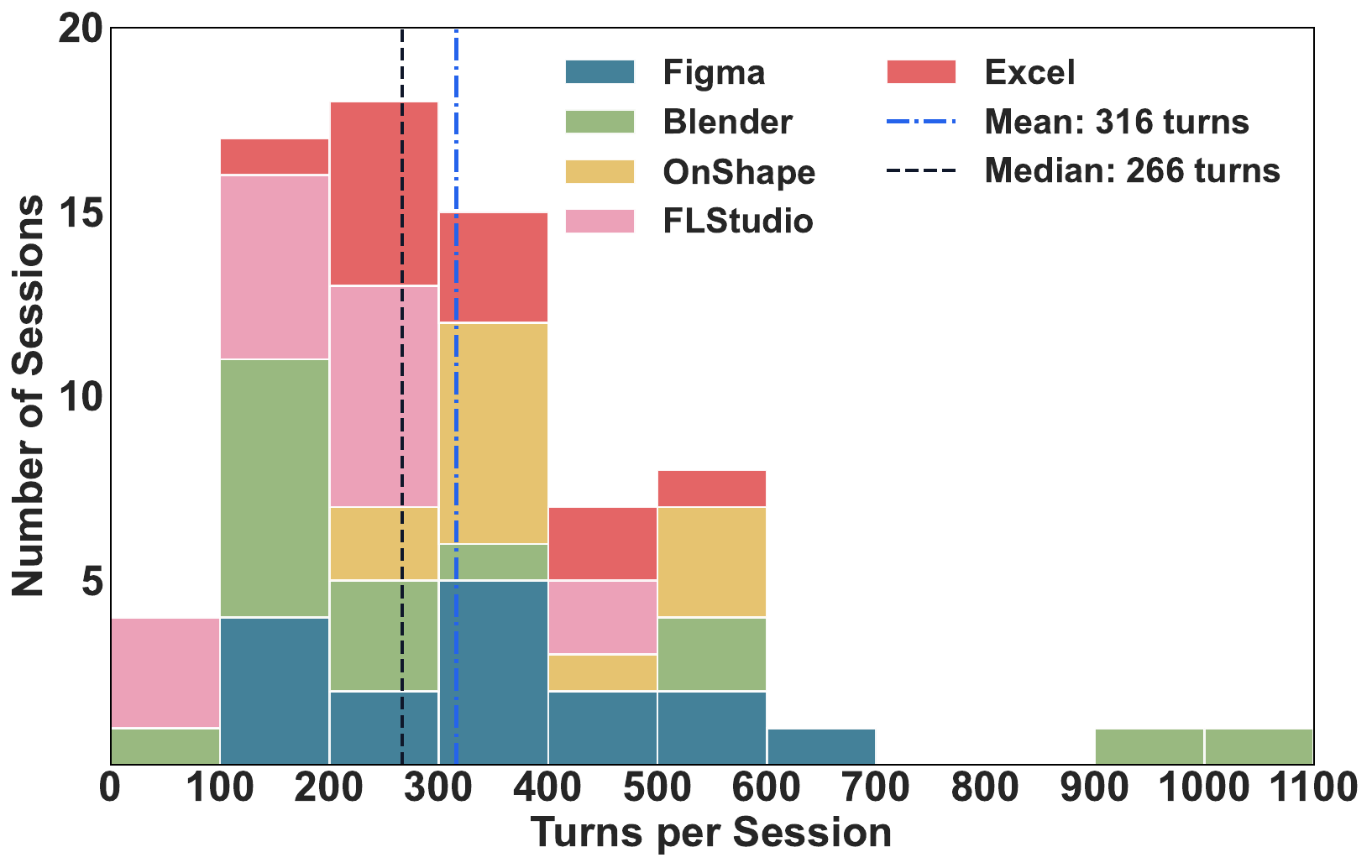}
    \caption{
    Distribution of number of utterances across \sessions{} sessions. Each bar shows the number of recorded coaching sessions within a duration range.
    }
    \label{fig:turn_count}
\end{figure}

\begin{figure}[t]
    \centering
    \includegraphics[width=\columnwidth]{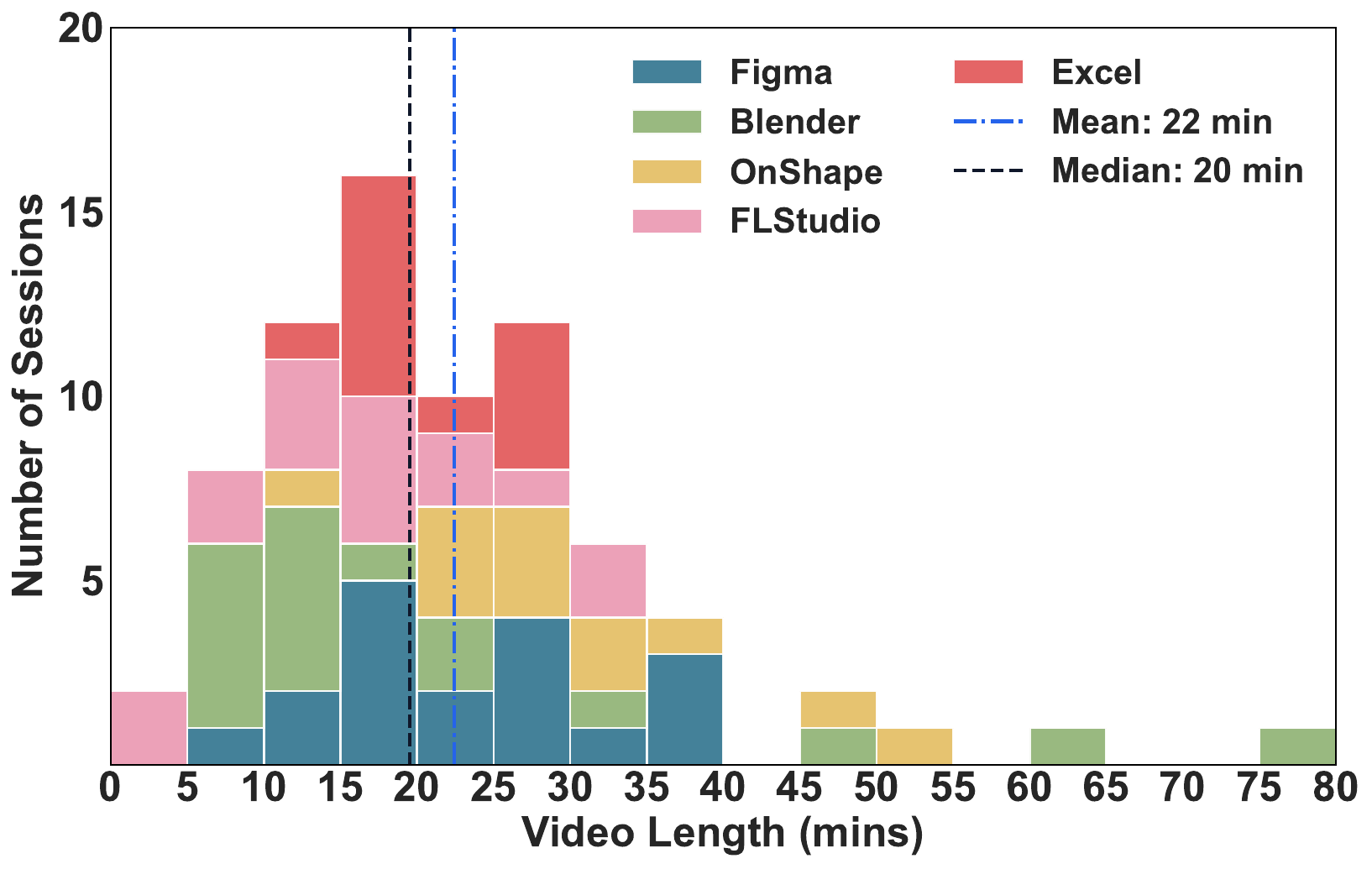}
    \caption{
    Distribution of video durations across \sessions{} sessions. Each bar shows the number of recorded coaching sessions within a duration range.
    }
    \label{fig:video_length}
\end{figure}

\begin{table}[t]
\centering
\small
\begin{tabularx}{\linewidth}{Xr}
\toprule
\textbf{Statistic} & \textbf{Value} \\
\midrule
\# Software Applications & 5 \\
\# Participants & 40 \\
\# Sessions & 72 \\
\midrule
Video Duration & \duration{}h \\
Avg. Duration & \avgduration{} \\
Min Duration & \minduration{} \\
Max Duration & \maxduration{} \\
\midrule
\# Total Turns & \turns{} \\
\% Coach Turns & 65.81\% \\
\% Learner Turns & 34.19\% \\
Avg. Turns per Session & 303.36 \\
Avg. Turn Tokens & 9.69 \\
\midrule
\# Input Events & \nevents{} \\
\# File Snapshots & \nfiles{} \\
\bottomrule
\end{tabularx}
\caption{Summary statistics of \system{}.}
\label{tab:dataset_overview}
\end{table}

In this section, we provide detailed information about \system{} and several examples of linguistic phenomena.
Table~\ref{tab:dataset_overview} shows detailed statistics of \system{}.
Figure~\ref{fig:turn_count} shows the distribution of number of utterances across all sessions.
Figure~\ref{fig:video_length} shows the distribution of duration across all sessions.
Figure~\ref{fig:coaching_method} shows the distribution of coaching methods across the conversation progress of all 72 sessions.

\subsection{Additional Language Statistics}
\begin{figure}[t]
    \centering
    \includegraphics[width=\columnwidth]{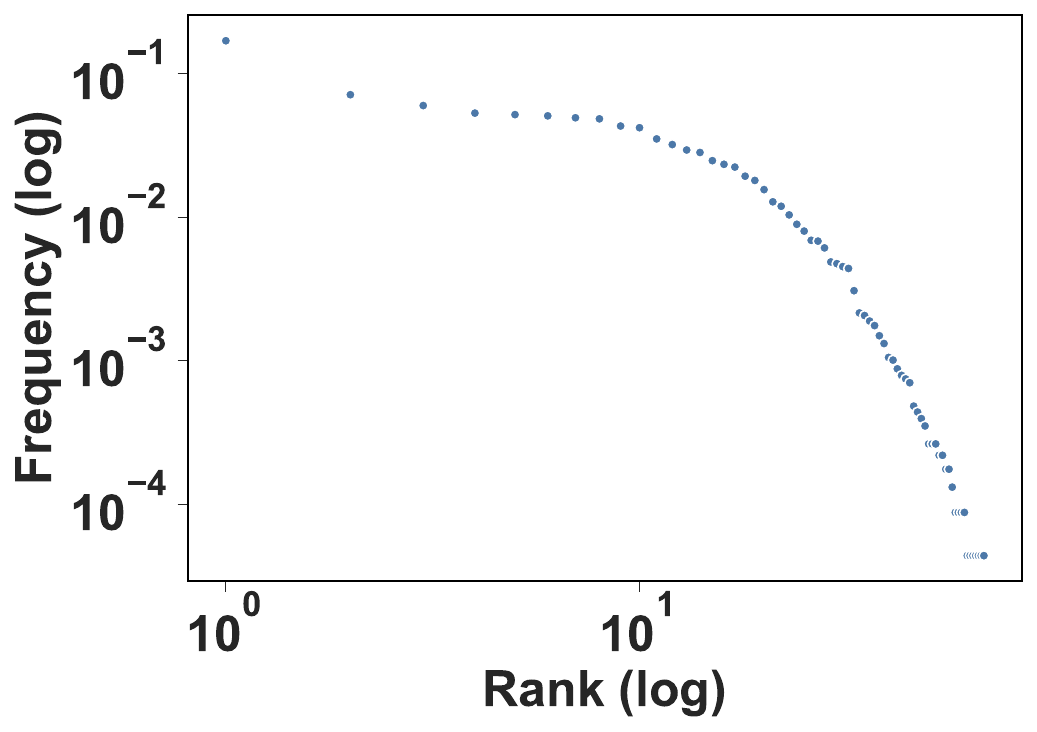}
    \caption{ 
    Token frequencies in \system{} follow a Zipfian distribution.
    }
    \label{fig:token_freq}
\end{figure}

\begin{figure}[t]
    \centering
    \includegraphics[width=\columnwidth]{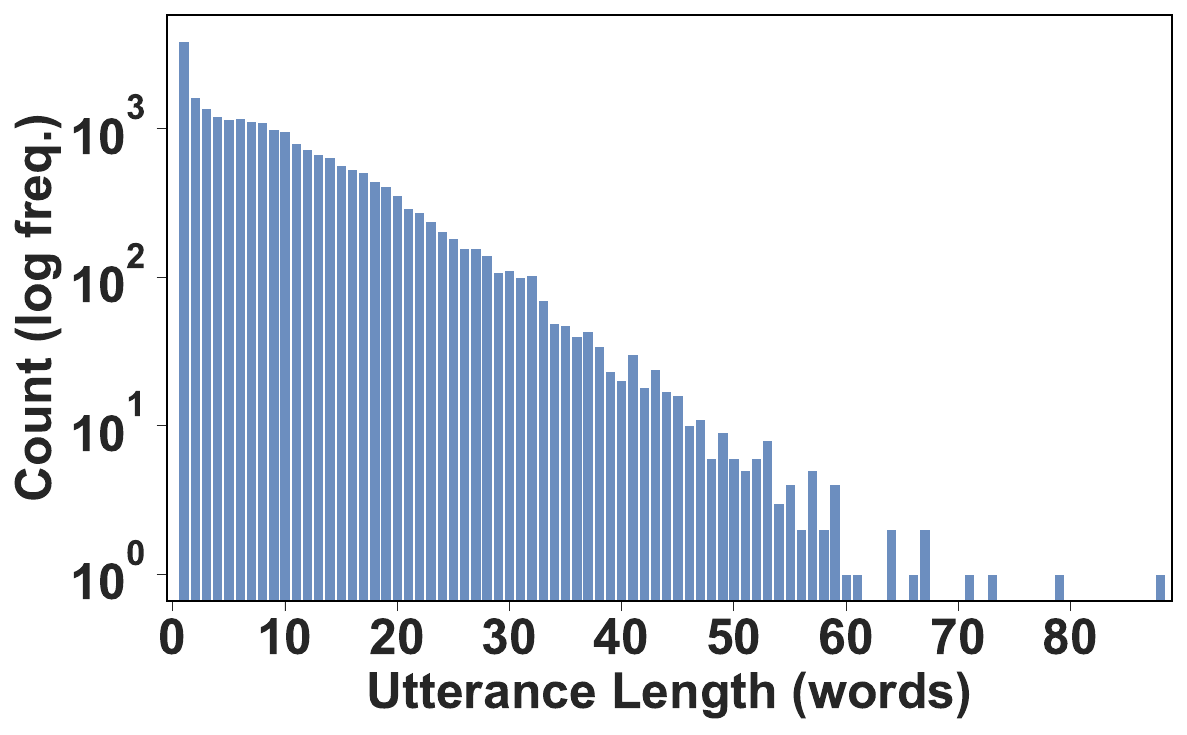}
    \caption{
    Token length (number of words per utterance) follows an exponential distribution.
    }
    \label{fig:token_len}
\end{figure}

\begin{figure}[t]
    \centering
    \includegraphics[width=\columnwidth]{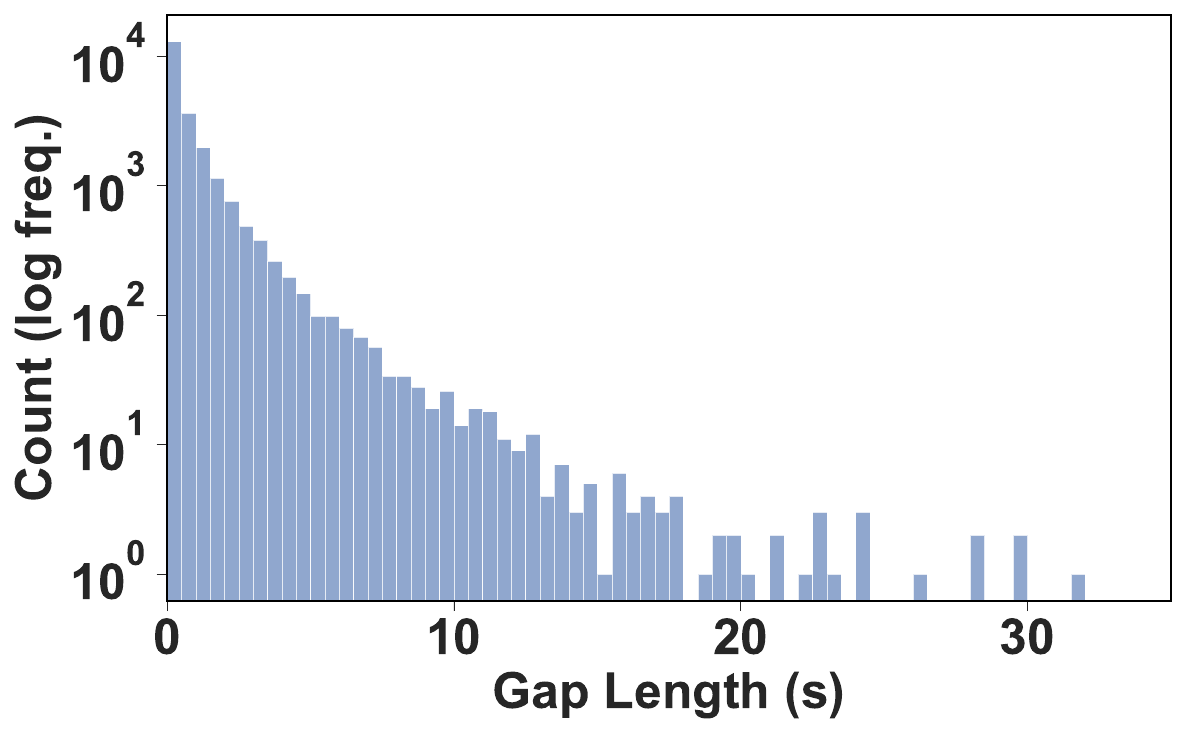}
    \caption{
     Distribution of gap length follows an exponential distribution. We define the gap as the duration of silence between the end of one speaker and the beginning of the next speaker~\cite{heldner_pauses_2010}.
    }
    \label{fig:gap_len}
\end{figure}

\noindent We observed linguistic phenomena in \system{}. 
Token frequency (Figure~\ref{fig:token_freq}) follows a Zipfian distribution. Token length (Figure~\ref{fig:token_len}) and gap length((Figure~\ref{fig:gap_len}) follow exponential distributions, with fewer tokens and smaller gaps being most common.

\subsection{Example Linguistic Phenomena}
\noindent In this section we provide examples of the linguistic phenomena observed in our coaching sessions.

\paragraph{Convention Formation.}
0104/C2L2:
\begin{itemize}
    \item Turn 386 (Coach): ``Alright, now we want to rename this to\ldots \emph{Card}.''
    \item Turn 387 (Learner): ``\emph{Card}.''
    \item Turn 388 (Coach): ``Mhm. And let's make \emph{card} have a corner radius of 8.''
\end{itemize}

\paragraph{Spatial Referencing.}
0104/C2L2:
\begin{itemize}
    \item Turn 60 (Coach): ``Yep, and then just rotate it, um, by dragging on \emph{this corner right here}\ldots''
    \item Turn 62 (Coach): ``Just hover over \emph{this part}.''
    \item Turn 63 (Learner): ``Hover \emph{in}?''
    \item Turn 64 (Coach): ``Hover over, like, \emph{slightly outside of it}, like, \emph{where my pink is}.''
    \item Turn 65 (Learner): ``Ah, slightly--- okay.''
\end{itemize}

\paragraph{Repair.}
0104/C2L2:
\begin{itemize}
    \item Turn 84 (Coach): ``Yep, now we have a, um, like, \emph{component} for this play button.''
    \item Turn 85 (Coach): ``Alright, so now we want to create, like, our album art frame, so\ldots''
    \item Turn 86 (Learner): ``\emph{Wait}, well, component means\ldots uh, \emph{what again, exactly}? It's interactable with\ldots''
    \item Turns 87--90 (Coach): ``So, components are, like, the parent, um, designs\ldots\ [extended explanation] \ldots That's a variant of this parent component.''
    \item Turn 91 (Learner): ``\emph{Okay, okay, got it, got it}.''
\end{itemize}

\section{Study Protocol}
\label{sec:protocol}
\noindent Each study session involved one expert coach, one novice learner, and one researcher facilitator. Sessions lasted approximately 3 hours and consisted of consent and setup, followed by 3--4 task blocks. Each task block included a pre-task, a tutorial task, and a post-task.

\subsection{Device Setup}
\noindent The coach and learner communicated remotely via Zoom throughout the session. Learners conducted all software activities on a study laptop (i.e., Lenovo E14 Gen 6) and use research accounts rather than their personal account to ensure privacy.

We used Zoom as the communication platform and the source of synchronized audio and screen recordings of the session. We recorded at 25 frames per second with 1920×1128 resolution. The audio was recorded from the study laptop microphone in mono channel with a 48K sampling rate.

We used Workflow Induction Toolkit~\cite{wang_how_2025} to log mouse and keyboard actions on the study laptop. In addition, we implemented software-specific file loggers to save file snapshots periodically (i.e., every 10s), allowing us to reconstruct intermediate work states over time.

\subsection{Study Script}
\noindent One facilitator was present at each session. They were provided with the following script to facilitate the session.

\subsubsection{Before Study}
\begin{itemize}
    \item \action{Set up the Zoom meeting}
    \item \action{Rename the participants as Coach, Learner, and Facilitator}
    \item \action{Share the study laptop screen with only the software window visible}
    \item \action{Hide the Zoom video pane and toolbar (Ctrl + Shift + Alt + H)}
\end{itemize}

\subsubsection{Introduction and Consent}
Thank you both for joining today. I am \placeholder{facilitator name}, and I will be the session facilitator.

In this study, you will work together to complete a set of activities using \placeholder{software name}. Today, \placeholder{coach name} will act as the coach, and \placeholder{learner name} will act as the learner. The goal of the study is to understand how experts provide novices with instructions to learn \placeholder{software name}. 

During the session, we will record the learner's screen, both participants' audio, and interaction logs from the learner's computer. I am now sending you the consent form and media release form. By signing these forms, you confirm that you understand the study procedures, that your questions have been answered, and that you agree to participate. \action{Send consent forms.}

We also have this survey we would like you to fill out that asks about your experiences with \placeholder{software name}. Please fill it out and let us
know when you’ve completed it. Let us know if you have any questions about it. \action{Send pre-study survey.}

\placeholder{coach name}, can you see the green pencil icon at the bottom left of your screen? That is the Zoom annotation tool. Set the annotation color to \textbf{pink} and the line thickness to \textbf{thick}. Please use it to briefly draw on the learner's screen. 

Next, \placeholder{learner name}, please grant remote control access to the coach. \placeholder{coach name} may request remote control if needed during the coaching activity. \action{Ask the coach to test remote control by moving the mouse and typing briefly.}

If any technical issues arise during the study, please let me know through Zoom chat. Otherwise, I will remain silent during the task activities and will only intervene if necessary.

\subsubsection{Task Block}
Each task block lasted approximately 50 minutes and consisted of three phases:
\begin{itemize}
    \item Pre-task (up to 15 minutes)
    \item Tutorial task (10--60 minutes)
    \item Post-task (up to 15 minutes)
\end{itemize}

\paragraph{Pre-Task} We will now begin the pre-task. In this task, \placeholder{learner name} will work on the task independently based on the task description. \placeholder{coach name} should remain silent. \placeholder{learner name}, you are encouraged to think aloud while working. You have up to 15 minutes to finish the task but you can stop at any time you want. \action{Share pre-task document and starter file.} \action{Start a 15-minute timer.}

\paragraph{Tutorial Task} We will now begin the tutorial task. There are no restrictions on how the coach and the learner should communicate. \action{Share tutorial task document and starter file.}

\paragraph{Post-Task} We will now begin the post-task. This task is the same as the pre-task. The learner will have 15 minutes to work on the task independently based on the task description. The coach should not provide instruction unless the learner appears to be stuck for 3 minutes, in which case the coach may provide at most one hint. \action{Share post-task document and starter file.} \action{Start a 15-minute timer.}

\paragraph{Actions Before Each Task}
\begin{itemize}
     \item \action{Start file logger.}
     \item \action{Start Workflow Induction Toolkit recorder.}
     \item \action{Start Zoom recording.}
\end{itemize}

\paragraph{Actions After Each Task}
\begin{itemize}
    \item \action{Stop Zoom recording.}
    \item \action{Stop Workflow Induction Toolkit recorder.}
    \item \action{Stop file logger.}
    \item \action{Check that data have been saved correctly.}
\end{itemize}

\section{Tasks}
\label{sec:tasks}
\noindent Table~\ref{tab:software_learning_tasks} lists all tasks used in \system{}.
\begin{table*}[t]
\centering
\small
\setlength{\tabcolsep}{4pt}
\renewcommand{\arraystretch}{1.15}
\begin{tabularx}{\textwidth}{
    >{\RaggedRight\arraybackslash}p{0.13\textwidth}
    >{\RaggedRight\arraybackslash}p{0.17\textwidth}
    >{\RaggedRight\arraybackslash}X
    >{\RaggedRight\arraybackslash}p{0.17\textwidth}
}
\toprule
\textbf{Software} & \textbf{Task} & \textbf{Description} & \textbf{Source} \\
\midrule
\multirow{4}{=}{01 \textbf{Figma}} 
& 01 Potion Bottle
& Combine basic geometric shapes and manipulate vector paths to build a magic potion bottle organized in frames.
& \href{https://help.figma.com/hc/en-us/articles/13543867954711-Create-an-illustration-in-Figma-Design}{Official Tutorial} \\
& 02 Loading Animation 
& Create a looping loading animation using frames, ellipses, Smart Animate, and component variants for use in a mobile app.
& \href{https://help.figma.com/hc/en-us/articles/18892560291351-Create-a-loading-animation-in-Figma}{Official Tutorial}  \\
& 03 Responsive Card 
& Build a responsive podcast episode card using Auto Layout, constraints, image fills, and components that adapt to different sizes.
& \href{https://help.figma.com/hc/en-us/articles/18894664907287-Create-a-responsive-card-with-auto-layout-and-constraints}{Official Tutorial} \\
& 04 Flower Vase 
& Illustrate a transparent glass vase with flowers by designing repeated petal shapes and placing curved stems inside the vase.
& \href{https://help.figma.com/hc/en-us/articles/38456797147415-Illustrate-a-flower-vase-using-shapes-transforms-and-the-glass-effect}{Official Tutorial} \\
\midrule
\multirow{4}{=}{02 \textbf{Blender}} 
& 01 Donut Modeling 
& Create a donut mesh with icing using viewport navigation, Edit Mode, modifiers, and Sculpt Mode.
& \href{https://www.youtube.com/watch?v=tBpnKTAc5Eo}{Online Tutorial} \\
& 02 Shading 
& Assign and preview separate materials for a donut and icing using Blender's Shading view.
& \href{https://www.youtube.com/watch?v=fsLO1F5x7yM}{Online Tutorial} \\
& 03 Dropping Animation 
& Animate an existing donut model to fall downward along the Z axis using keyframes.
& \href{https://www.youtube.com/watch?v=o19U-yPGdyY}{Online Tutorial} \\
& 04 Sprinkles 
& Add procedurally generated sprinkles to an existing donut's icing in Blender.
& \href{https://www.youtube.com/watch?v=EWTOy5-e4Ns}{Online Tutorial} \\
\midrule
\multirow{3}{=}{03 \textbf{OnShape}} 
& 01 Saddle Bracket
& Model a saddle bracket using sketching, extrusion, and semicircular cuts to match provided engineering dimensions.
& \href{https://www.youtube.com/watch?v=Bkvw6dNT-T4}{Online Tutorial} \\
& 02 Geneva Cam
& Model a Geneva cam using sketch geometry, circular repetition, extrusion, and feature patterning from an engineering drawing.
& \href{https://www.youtube.com/watch?v=c50bmEeOIbI}{Online Tutorial} \\
& 03 Hub 
& Build a hub by revolving a base profile and adding repeated curved slots to match specified diameters and angular spacing.
& \href{https://www.youtube.com/watch?v=UK1mblfM-94}{Online Tutorial} \\
\midrule
\multirow{4}{=}{04 \textbf{FL Studio}} 
& 01 Beat 
& Build a drum pattern with bass line using the Channel Rack and arrange patterns in the Playlist.
& \href{https://www.image-line.com/learn/lesson/making-a-beat-in-channel-rack}{Official Tutorial} \\
& 02 Melody 
& Create and edit a melody in the Piano Roll, shaping note timing and dynamics before placing it in the Playlist.
& \href{https://www.image-line.com/learn/lesson/creating-melodies-in-piano-roll}{Official Tutorial} \\
& 03 Song Track 
& Place, duplicate, and organize beat and melody patterns in the Playlist to build a fuller track structure.
& \href{https://www.image-line.com/learn/lesson/arranging-a-track-in-the-playlist}{Official Tutorial} \\
& 04 Mixer 
& Route instruments to Mixer inserts and apply effects such as reverb and delay using FL Studio's Mixer.
& \href{https://www.image-line.com/learn/lesson/adding-effects-in-the-mixer}{Official Tutorial} \\
\midrule
\multirow{3}{=}{05 \textbf{Excel}} 
& 01 SUMIFS
& Use SUMIFS to compute total revenue filtered by multiple conditions across a raw sales dataset.
& \href{https://www.youtube.com/watch?v=jPajIyavh_0}{Online Tutorial} \\
& 02 VLOOKUP 
& Link two tables with VLOOKUP to auto-fetch prices, categories, and tax rates for revenue calculations.
& \href{https://www.youtube.com/watch?v=d3BYVQ6xIE4}{Online Tutorial} \\
& 03 Pivot Table 
& Summarize a dataset into multiple pivot views and charts to answer distinct business questions.
& \href{https://www.youtube.com/watch?v=PdJzy956wo4}{Online Tutorial} \\
\bottomrule
\end{tabularx}
\caption{Overview of tasks used in the study.}
\label{tab:software_learning_tasks}
\end{table*}

\section{Participants}
\label{sec:participants}
\noindent Table~\ref{tab:coach-participants} details the demographics and prior experiences of coach participants. Table~\ref{tab:learner-participants} details the demographics and prior experiences of learner participants.

\begin{table*}[t]
\centering
\small
\setlength{\tabcolsep}{3.5pt}

\begin{tabular*}{\textwidth}{
    @{\extracolsep{\fill}}
    l l c c l c c c
    @{}
}
\toprule
\textbf{ID} &
\textbf{Software} &
\textbf{Age} &
\textbf{Gender} &
\textbf{Profession} &
\textbf{SW Exp. (yrs)} &
\textbf{Teaching Exp. (yrs)} &
\textbf{Teaching Type} \\
\midrule
C1  & Figma     & 29 & F & UI Designer                     & 9  & 4   & M \\
C2  & Figma     & 21 & F & Student-Run Courses Instructor & 5  & 2   & C \\
C3  & Blender   & 21 & F & Student-Run Courses Instructor & 5  & 2   & C \\
C4  & Figma     & 34 & F & UI Designer                     & 6  & 3   & T \\
C5  & Blender   & 23 & M & 3D Visualization Specialist    & 7  & 4   & T \\
C6  & Blender   & 43 & M & Technical Artist               & 10 & 6   & T \\
C7  & Blender   & 27 & M & Freelance 3D Artist            & 5  & 1   & T \\
C8  & OnShape   & 21 & M & Student-Run Courses Instructor & 5  & 1.5 & T \\
C9  & OnShape   & 38 & M & Mechanical Engineer            & 14 & 8   & T \\
C10 & OnShape   & 40 & M & CAD Instructor                 & 8  & 8   & C, T \\
C11 & OnShape   & 42 & M & Mechanical Engineer            & 8  & 3   & M \\
C12 & FL Studio & 40 & M & Music Team Manager             & 10 & 2   & T \\
C13 & Figma     & 26 & F & UX Designer                    & 5  & 1   & M \\
C14 & FL Studio & 38 & M & Audio Engineer                 & 15 & 7   & T \\
C15 & FL Studio & 24 & M & Audio Engineer                 & 5  & 1.5 & M \\
C16 & FL Studio & 32 & M & Music Producer                 & 15 & 9   & T \\
C17 & Excel     & 27 & M & Data Analyst                   & 5  & 3   & M \\
C18 & Excel     & 46 & M & Data Scientist                 & 12 & 12  & C, M \\
C19 & Excel     & 46 & M & Business Manager               & 26 & 10  & M \\
C20 & Excel     & 58 & F & Technical Program Manager      & 10 & 5   & M \\
\bottomrule
\end{tabular*}

\caption{Demographic information and prior experience of expert participants
($N=20$). Gender: F = female, M = male. Teaching type: M = mentoring,
T = tutoring, and C = course instruction.}
\label{tab:coach-participants}
\end{table*}

\begin{table*}[t]
\centering
\small

\begin{tabular}{@{}llcclc@{}}
\toprule
\textbf{ID} &
\textbf{Software} &
\textbf{Age} &
\textbf{Gender} &
\textbf{Profession} &
\textbf{Software Exp. (yrs)} \\
\midrule
L1  & Figma     & 23 & Female & Undergrad Student & 0 \\
L2  & Figma     & 33 & Male   & Undergrad Student & 0 \\
L3  & Blender   & 18 & Male   & Undergrad Student & 0 \\
L4  & Figma     & 25 & Female & Undergrad Student & 0--1 \\
L5  & Blender   & 19 & Male   & Undergrad Student & 0 \\
L6  & Blender   & 25 & Male   & Graduate Student  & 0 \\
L7  & Blender   & 22 & Female & Undergrad Student & 0 \\
L8  & OnShape   & 26 & Male   & Graduate Student  & 0 \\
L9  & OnShape   & 31 & Male   & Graduate Student  & 0 \\
L10 & OnShape   & 25 & Female & Graduate Student  & 0 \\
L11 & OnShape   & 26 & Female & Graduate Student  & 0 \\
L12 & FL Studio & 26 & Female & Graduate Student  & 0 \\
L13 & Figma     & 25 & Male   & Graduate Student  & 0 \\
L14 & FL Studio & 25 & Female & Graduate Student  & 0 \\
L15 & FL Studio & 24 & Female & Graduate Student  & 0 \\
L16 & FL Studio & 23 & Male   & Graduate Student  & 0 \\
L17 & Excel     & 26 & Male   & Graduate Student  & 0--1 \\
L18 & Excel     & 22 & Female & Undergrad Student & 0--1 \\
L19 & Excel     & 23 & Female & Graduate Student  & 0--1 \\
L20 & Excel     & 21 & Male   & Undergrad Student & 0 \\
\bottomrule
\end{tabular}

\caption{Demographic information and prior software experience of learner participants ($N=20$).}
\label{tab:learner-participants}
\end{table*}

\section{Transcription}
\label{sec:transcription}

\noindent We use the automatically generated closed caption (i.e., \texttt{.vtt} file) from Zoom as a starting point as it includes accurate timestamps, speaker labels (coach vs.\ learner), and an initial segmentation of the dialogue. 

\subsection{Transcript Correction}
\noindent We use an LLM (i.e., Gemini 3.1 Pro) to correct wrong words, homophone errors, dropped or extra words, or other mismatches with what is actually said. Prompt used for transcript correction can be found in Figure~\ref{fig:autocorrect-transcription-prompt}. One author then manually validate the transcript to correct the speaker label and utterances. We use following notation to indicate pauses and interruptions: 
\begin{itemize}
    \item Use \texttt{comma} (\texttt{,}) to indicate a short speech pause or hesitation.
    \item Use \texttt{ellipsis} (\texttt{...}) to indicate a longer pause or hesitation.
    \item Use \texttt{double dash} (\texttt{--}) to indicate trailing off or interruption by another speaker.
    \item Use \texttt{parentheses} (\texttt{( )}) to indicate an uncertain best-guess interpretation, e.g., \texttt{(move)}. If no reasonable guess is available, leave the parentheses empty.
     \item Use \texttt{parentheses} (\texttt{( )}) to replace any personal information, e.g., \texttt{(facilitator)}.
\end{itemize}
In general, we will not annotate intonational features and non-spoken noises.

\subsection{Utterance Segmentation}
\noindent Three authors split and merged the corrected utterances according to the following guideline:
\begin{enumerate}
    \item Split the utterance if there is a pause of 2 seconds or longer.
    \item Use changes in intonation and what can be inferred from the semantics to decide whether to split for shorter pauses.
    \begin{enumerate}
        \item Descending pitch can be a hint that an utterance is ending.
        \item If it is difficult to assign different functional meanings to two adjacent spans, they should likely be merged into one utterance.
    \end{enumerate}
    \item Context from the other speaker can also help (e.g., if one half of the utterance is responding directly to the other speaker, it may make sense to split).
    \item Use the multimodal context to decide whether to split.
    \begin{enumerate}
        \item If a short pause coincides with a visible action in the screen state, this can be evidence for a new utterance.
        \item If the screen state remains unchanged and there is only a pause shorter than 2 seconds, the utterances of the same speaker should generally be merged.
    \end{enumerate}
\end{enumerate}

\subsection{Transcript Correction Interface}
\begin{figure}[t]
    \centering
    \includegraphics[width=\columnwidth]{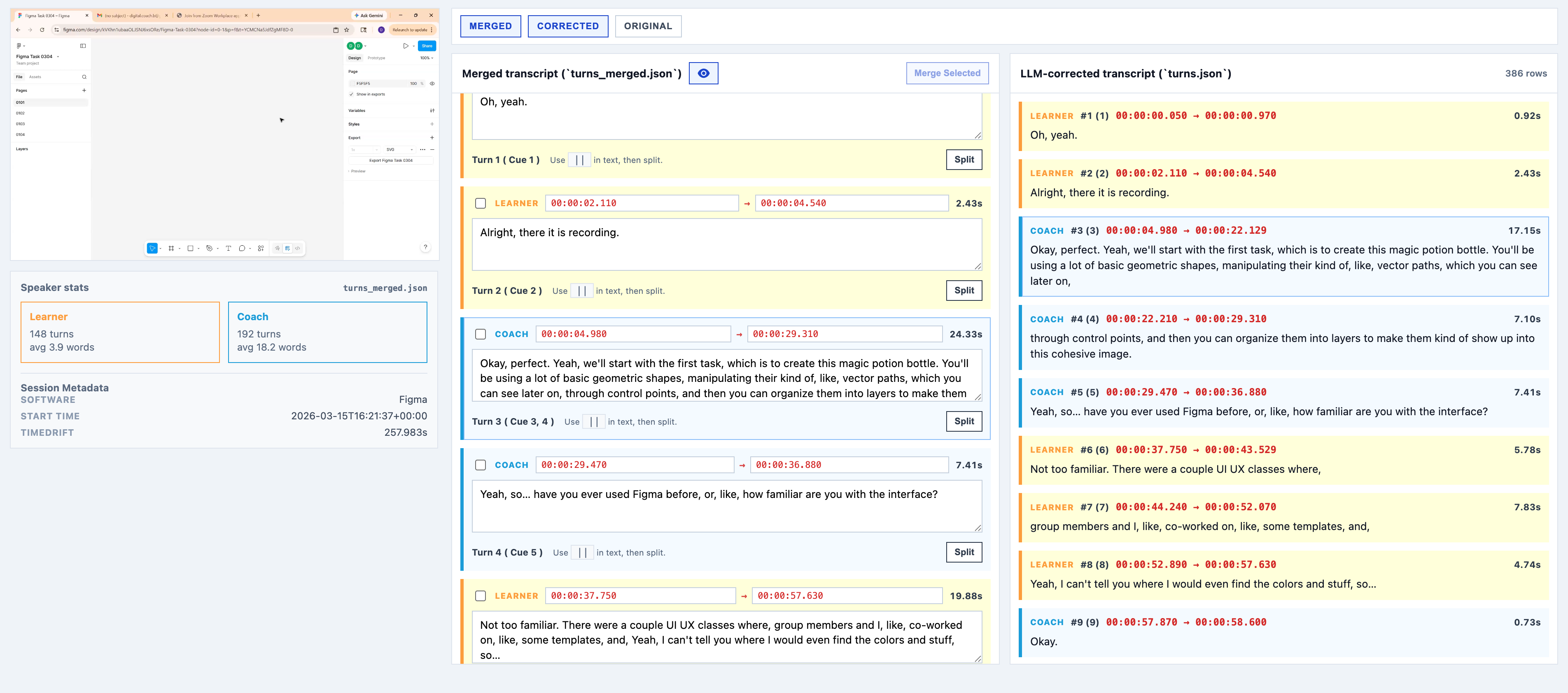}
    \caption{
    Transcription correction interface. Editor can see the dialogue transcript alongside screen recording. They can use the interface to edit, split, merge each utterance and trim its start and end time.
    }
    \label{fig:transcript_correction_interface}
\end{figure}

\noindent The transcript correction interface screenshot is shown in Figure~\ref{fig:transcript_correction_interface}.

\section{Annotation}
\label{sec:annotation}
\subsection{Human Annotation Method}
\noindent To derive the taxonomy, three researchers used open coding on 200 turns to obtain potential coaching methods. Then, they met to merge together similar methods and create a codebook with a name, definition, and example for each coaching method. The researchers iteratively coded samples and revised the codebook to achieve final codes, containing 10 methods. During the coding process, we merged similar codes. Disagreements were resolved through discussion, and the finalized codebook was used for the annotations. Using the codebook with examples, three researchers annotated other 200 utterances and reached a substantial agreement for coaching methods.

\subsection{Dialogue Acts Schema}
\noindent  The full dialogue acts definitions and examples provided to annotators are shown in Table~\ref{tab:dialogue_act_schema}.

\begin{figure}[t]
    \centering
    \includegraphics[width=\columnwidth]{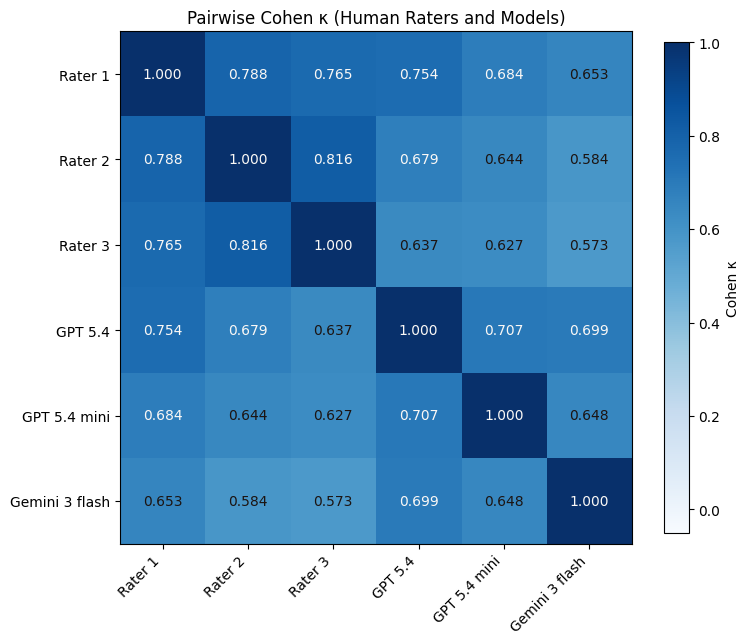}
    \caption{
    Pairwise Cohen's \(\kappa\) agreement among three human raters and three model annotators for dialogue act annotation. Rows and columns denote annotators, and each cell reports the pairwise agreement score. GPT-5.4 aligns with human rater the best.
    }
    \label{fig:dialogue_cohen}
\end{figure}

\begin{table*}[t]
\centering
\small
\setlength{\tabcolsep}{5pt}
\renewcommand{\arraystretch}{1.15}
\begin{tabular}{p{3.0cm} p{7.5cm} p{4.3cm}}
\toprule
\textbf{Tag} & \textbf{Definition} & \textbf{Example} \\
\midrule

\textbf{Info Request}
& Seeks information, clarification, or confirmation from the addressee.
& \textit{``What do you want to do with them?''} \\

\textbf{Action Directive}
& Guides the addressee toward an action.
& \textit{``Click on the top menu.''} \\

\textbf{Answer}
& Provides information requested by a previous utterance.
& \textit{``Because the polygon is not closed.''} \\

\textbf{Inform}
& Provides objective information or describes the current state.
& \textit{``The tempo is set to 120 BPM.''} \\

\textbf{Opinion}
& Provides a subjective judgment, preference, or interpretation.
& \textit{``The alignment looks better now.''} \\

\textbf{Backchannel}
& Signals attention, understanding, agreement, or receipt.
& \textit{``Okay.''} \\

\textbf{Not Enough Info}
& Is too incomplete or ambiguous to classify reliably.
& \textit{``If you just---''} \\

\bottomrule
\end{tabular}
\caption{Dialogue act schema used to annotate coach and learner utterances.}
\label{tab:dialogue_act_schema}
\end{table*}

\subsection{Coaching Methods Schema}
\noindent The full coaching methods definitions and examples provided to annotators are shown in Table~\ref{tab:coaching_method_schema}.
\begin{table*}[t]
\centering
\small
\setlength{\tabcolsep}{5pt}
\renewcommand{\arraystretch}{1.15}
\begin{tabular}{p{3cm} p{8cm} p{4.2cm}}
\toprule
\textbf{Tag} & \textbf{Definition} & \textbf{Example} \\
\midrule

\textbf{Direct Instruction}
& Gives concrete procedural guidance about next action to take.
& \textit{``Click Sketch, and then select that edge.''} \\

\textbf{Plan}
& Proposes a subgoal without specifying the exact action.
& \textit{``Let's first make the basic shape, then add the details.''} \\
\midrule

\textbf{Confirmation}
& Provides affirmation on the learner’s action, result, or understanding.
& \textit{``Yes, that shape looks correct now.''} \\

\textbf{Diagnosis}
& Provides feedback on the learner’s current state or identifies a problem.
& \textit{``The sketch isn't closed, so the software can't generate a solid.''} \\
\midrule

\textbf{Explanation}
& Explains a concept, tool, feature, or reason behind an action or outcome.
& \textit{``We use Extrude here because it turns a 2D profile into a 3D solid.''} \\

\textbf{Tip}
& Provides a transferable shortcut, heuristic, or practical technique.
& \textit{``It's often easier to model one side first and then mirror it.''} \\
\midrule

\textbf{Clarification}
& Asks about the learner's goal, confusion, current state, or intended next step before giving guidance.
& \textit{``Does this make sense to you?''} \\

\textbf{Reflection}
& Prompts the learner to evaluate an outcome, compare alternatives, or consider what happened.
& \textit{``Looking at the result, what do you think went wrong?''} \\

\textbf{Articulation}
& Prompts the learner to verbalize their reasoning, plan, or decision process.
& \textit{``Can you walk me through your reasoning?''} \\

\textbf{Exploration}
& Invites the learner to experiment, choose among alternatives, or practice themselves.
& \textit{``Try a couple of options and see which one you like better.''} \\
\midrule

\textbf{Non-Instructive}
& Talks about task-irrelevant topics or filler.
& \textit{``The internet is not good.''} \\

\bottomrule
\end{tabular}
\caption{Coaching method schema used to annotate expert instructional behavior.}
\label{tab:coaching_method_schema}
\end{table*}

\begin{figure}[t]
    \centering
    \includegraphics[width=\columnwidth]{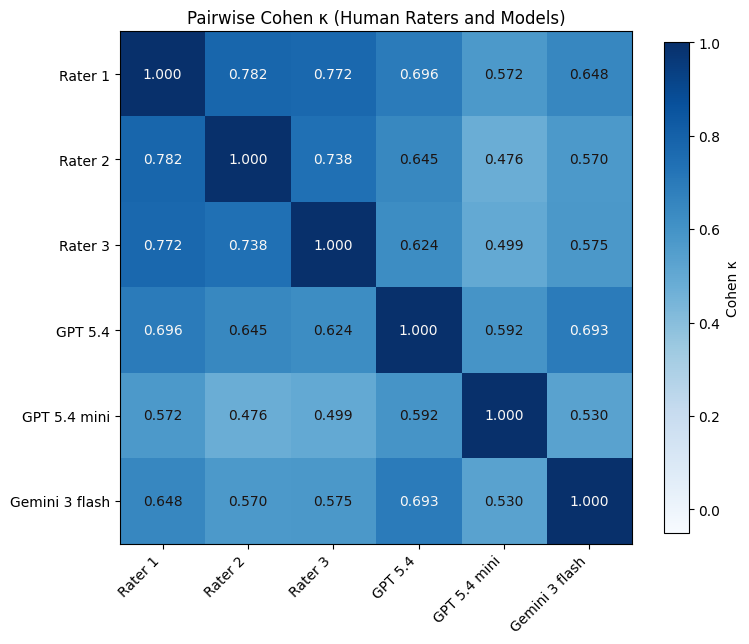}
    \caption{
    Pairwise Cohen's \(\kappa\) agreement among three human raters and three model annotators for coaching method annotation. Rows and columns denote annotators, and each cell reports the pairwise agreement score. GPT-5.4 aligns with human rater the best.
    }
    \label{fig:coaching_cohen}
\end{figure}

\subsection{Additional Annotation Results}
\label{sec:additional-annotation}
\begin{figure}[t]
    \centering
    \includegraphics[width=\columnwidth]{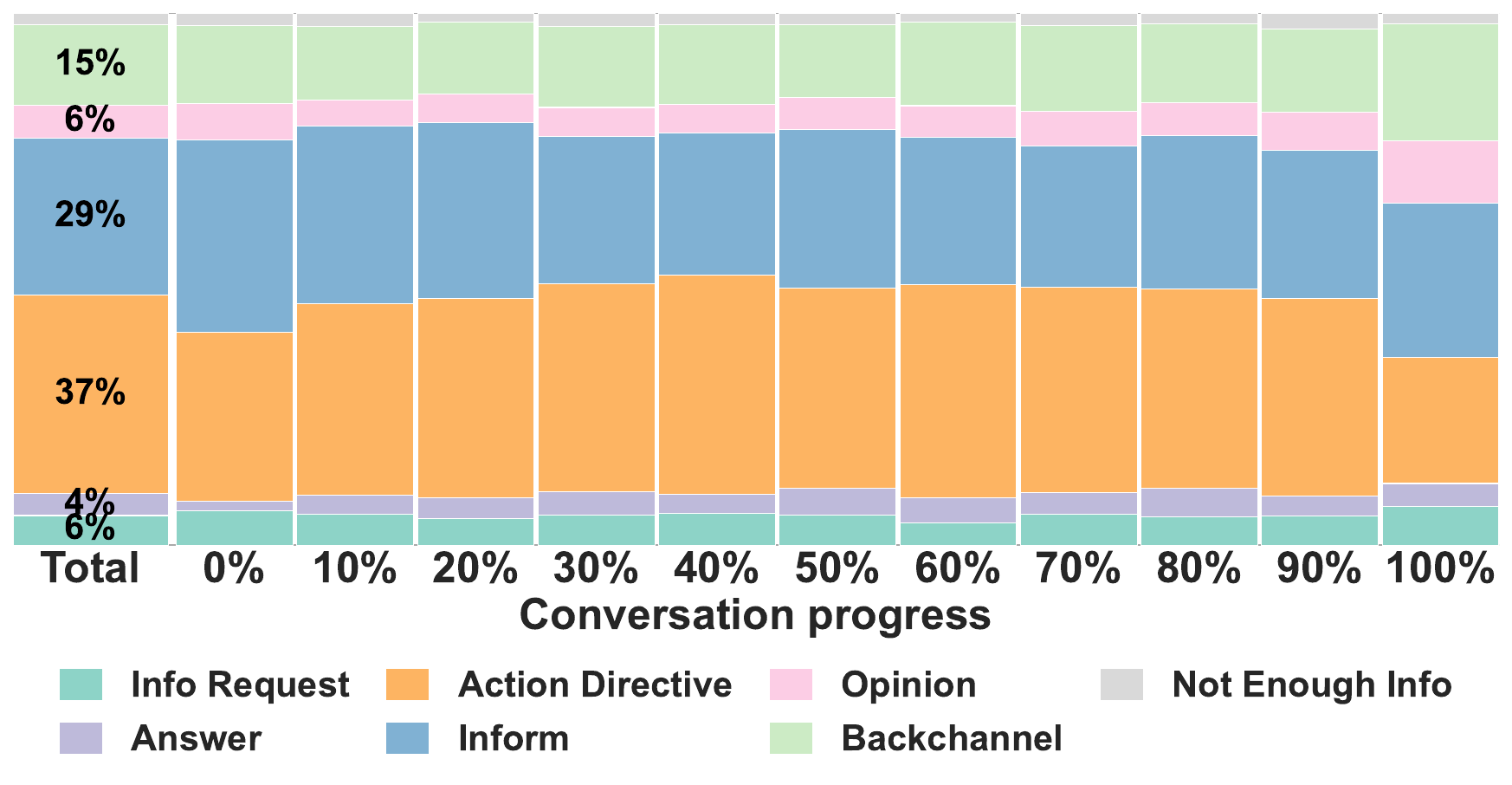}
    \caption{
     Distribution of dialogue acts of coaches across conversation progress.
     Method use remains relatively stable as the conversation progresses, except for the beginning and the end 10\%.
    }
    \label{fig:dialogue_acts_coach}
\end{figure}

\begin{figure}[t]
    \centering
    \includegraphics[width=\columnwidth]{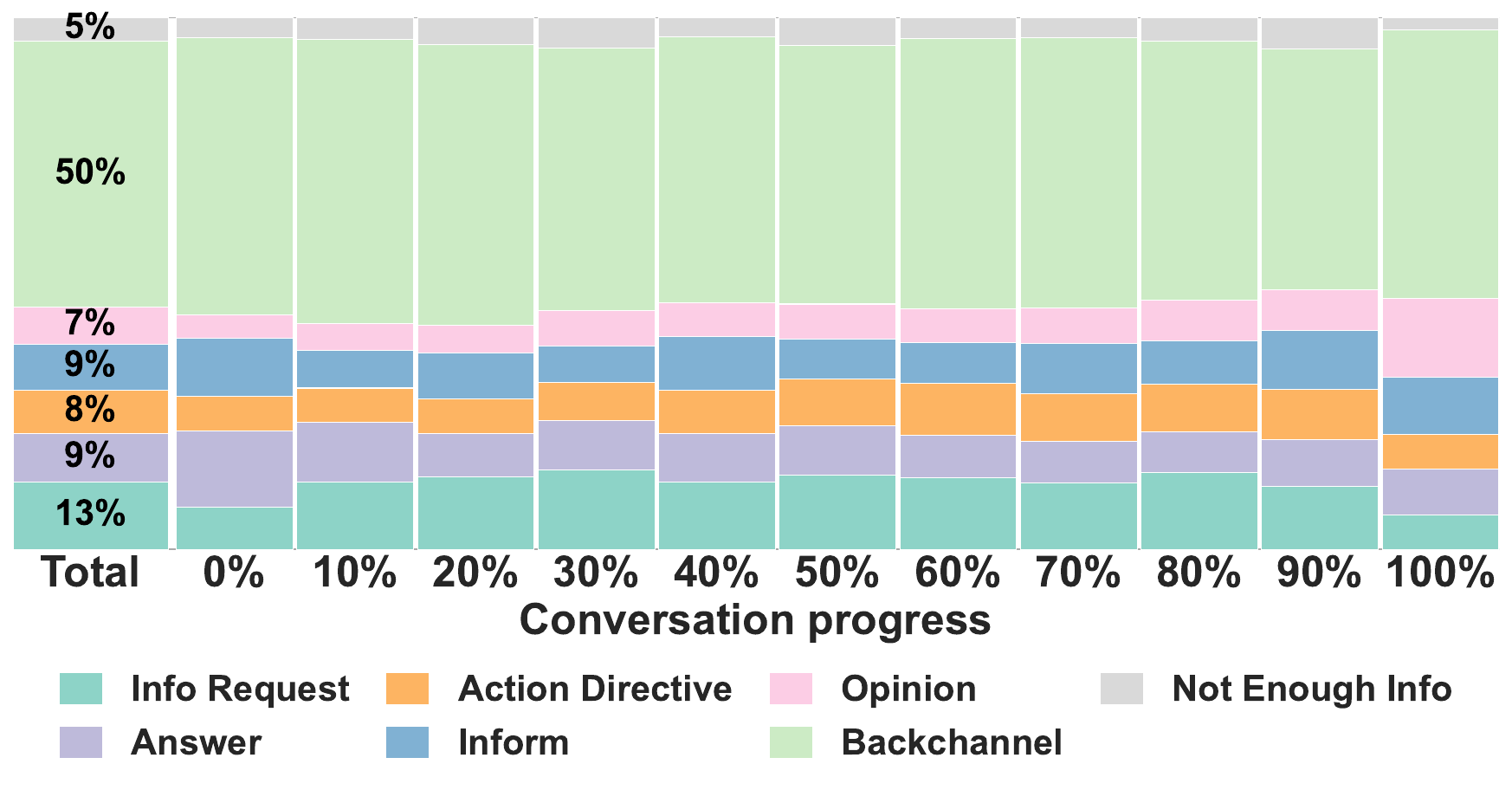}
    \caption{
     Distribution of dialogue acts of learners across conversation progress.
     Dialogue acts remain relatively stable as the conversation progresses, except for the beginning and the end 10\%.
    }
    \label{fig:dialogue_acts_learner}
\end{figure}

\begin{figure}[t]
    \centering
    \includegraphics[width=\columnwidth]{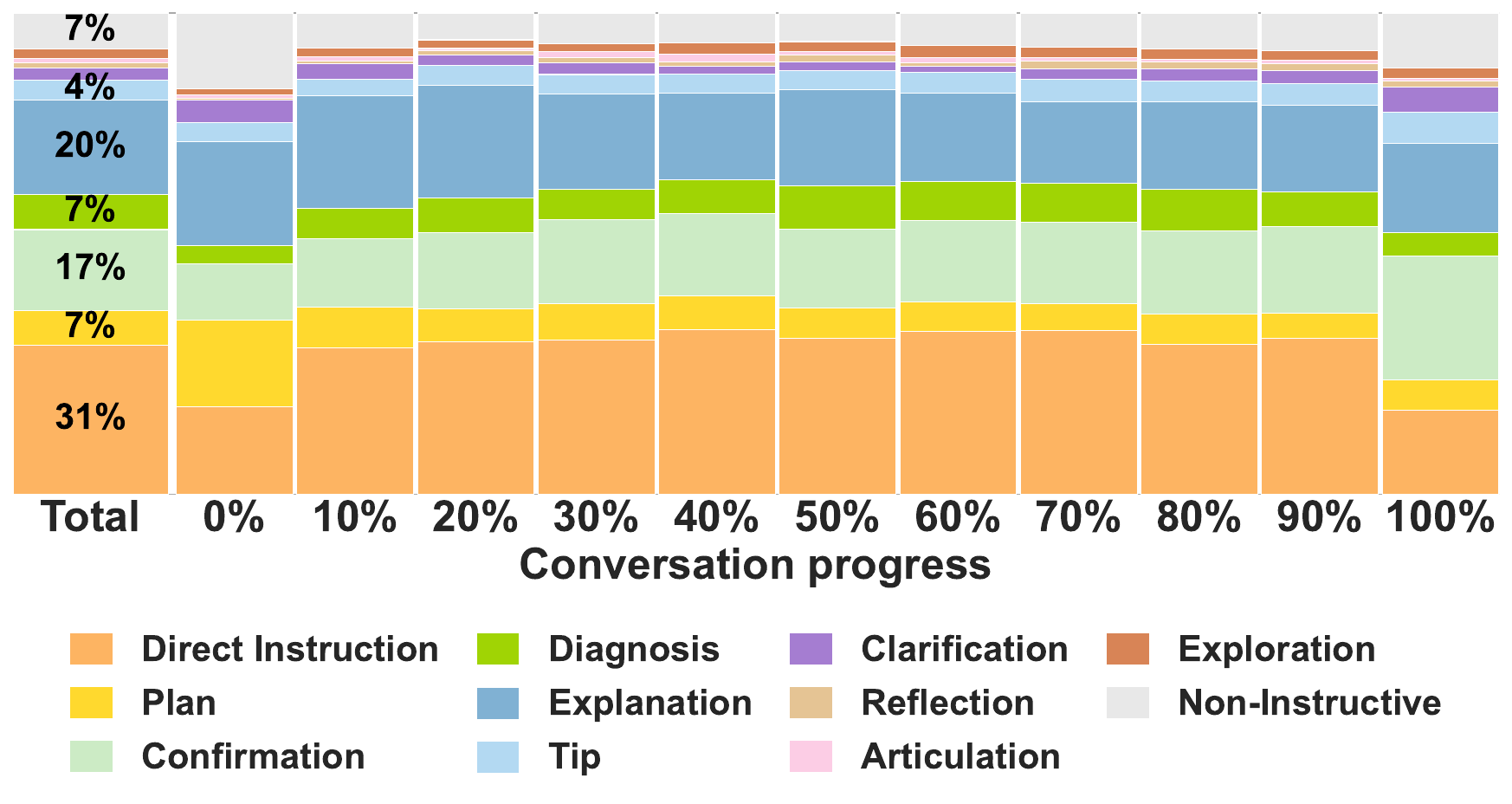}
    \caption{
     Distribution of coaching methods across conversation progress.
     Method use remains relatively stable as the conversation progresses, except for the beginning and the end 10\%.
    }
    \label{fig:coaching_method}
\end{figure}

\begin{figure}[t]
    \centering
\includegraphics[width=\columnwidth]{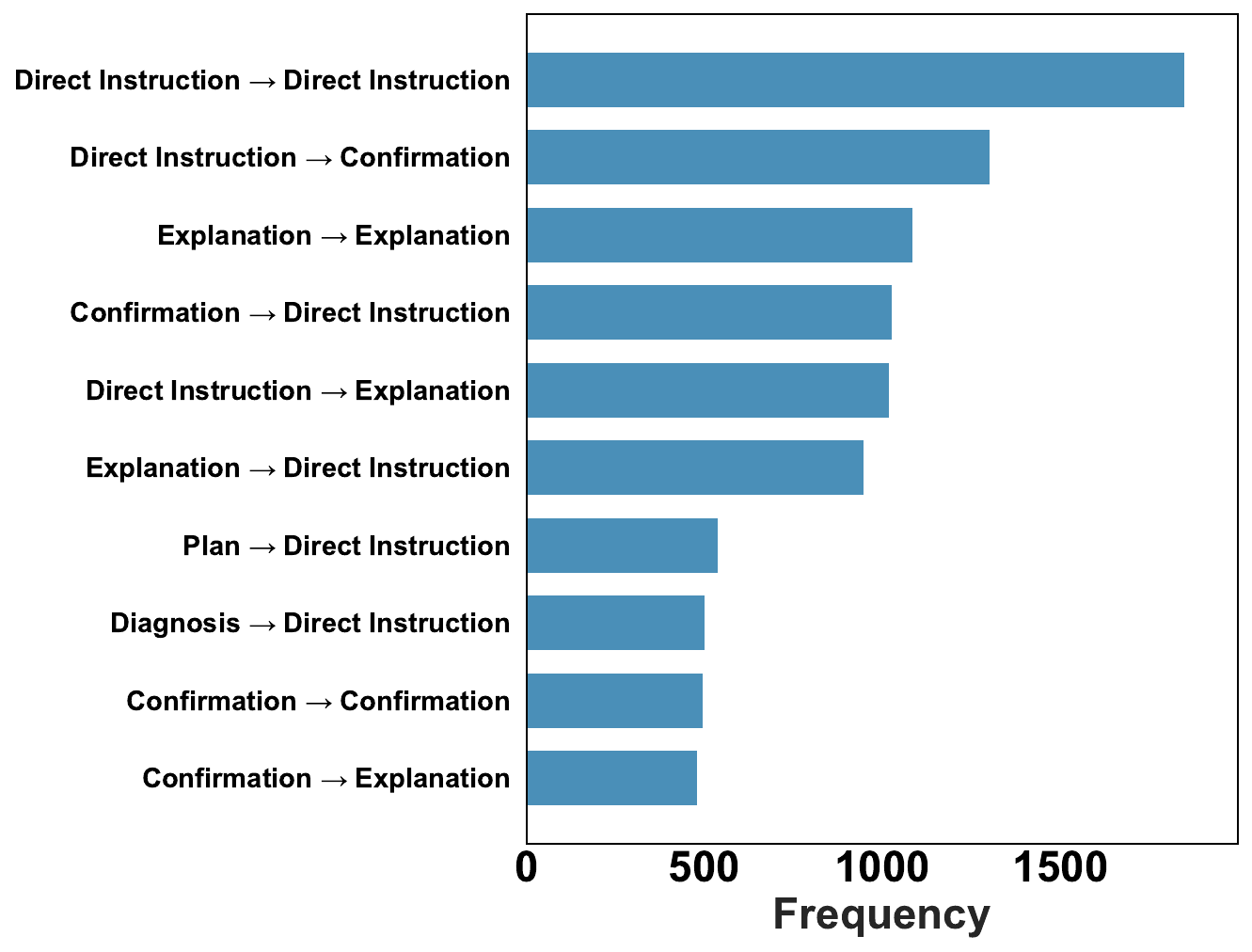}
    \caption{Top-10 most frequent coaching method bigrams.}
    \label{fig:bigram}
\end{figure}

\noindent 
Figure~\ref{fig:dialogue_acts_coach} shows the distribution of dialogue acts of coaches across conversation progress. 
Figure~\ref{fig:dialogue_acts_learner} shows the distribution of dialogue acts of learners across conversation progress. Figure~\ref{fig:coaching_method} shows the distribution of coaching methods across conversation progress. Figure~\ref{fig:bigram} shows top-10 most frequent coaching method
bigrams. 

\newtext{By software, screen annotations co-occur with 50.8\% of coach utterances in FL Studio, 37.5\% in Figma, 30.3\% in Blender, 16.1\% in Onshape, and 0\% in Excel. Eight of 20 coaches were heavy annotation users ($\geq$ 40\%), 4 were moderate users (20--40\%), and 8 used annotations rarely or not at all (<10\%).}

\subsection{Annotation Interface}
\begin{figure}[t]
    \centering
    \includegraphics[width=\columnwidth]{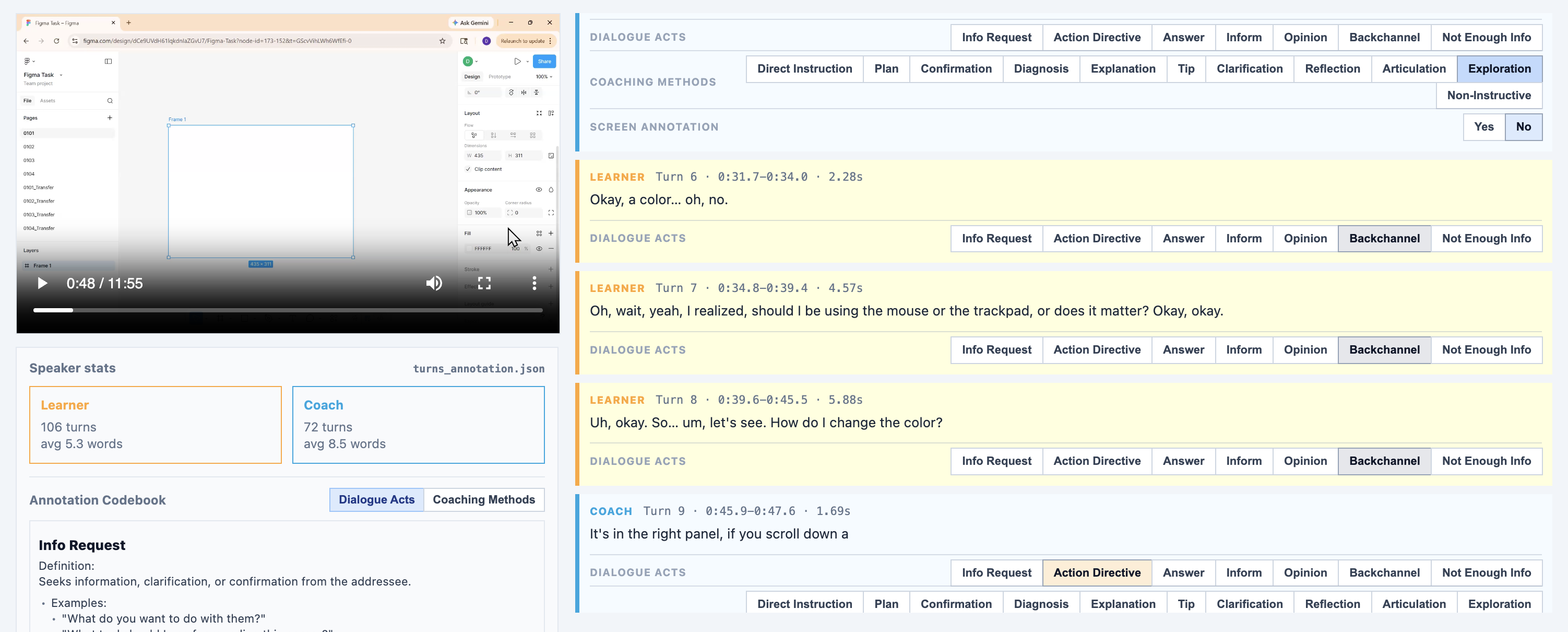}
    \caption{
    Annotation interface for utterance labeling. Annotators can see the dialogue transcript alongside screen recording and assign tags to each utterance.
    }
    \label{fig:annotation_interface}
\end{figure}

\noindent The annotation interface screenshot is shown in Figure~\ref{fig:annotation_interface}.

\section{Additional Evaluation Results}
\label{sec:additional-eval}

\noindent Table~\ref{tab:ngram-jsd} shows Jensen-Shannon divergence (JSD) for dialogue-act and coaching-method n-gram distributions. Table~\ref{tab:bleu_ablation} shows BLEU results for next coach utterance generation. Table~\ref{tab:meteor_ablation} shows METEOR results for next coach utterance generation. Table~\ref{tab:rouge_l_ablation} shows ROUGE-L results for next coach utterance generation. Table~\ref{tab:bertscore_ablation} shows BERTScore results for next coach utterance generation. Table~\ref{tab:domain_clair_base} shows CLAIR scores (Default) split by software application. Table~\ref{tab:method_clair_base} shows CLAIR scores (Default) split by coaching method.

\begin{table*}[t]
\centering
\small
\setlength{\tabcolsep}{3.8pt}
\renewcommand{\arraystretch}{1.08}
\begin{tabular}{llcccccc}
\toprule
\multirow{2}{*}{\textbf{Label Type}} 
& \multirow{2}{*}{\textbf{Model}} 
& \multicolumn{2}{c}{\textbf{1-Gram}} 
& \multicolumn{2}{c}{\textbf{2-Gram}} 
& \multicolumn{2}{c}{\textbf{3-Gram}} \\
\cmidrule(lr){3-4}\cmidrule(lr){5-6}\cmidrule(lr){7-8}
& & \textbf{V} & \textbf{C} & \textbf{V} & \textbf{C} & \textbf{V} & \textbf{C} \\
\midrule

\multirow{6}{*}{\textbf{Dialogue Acts}}
& \textbf{GPT-5.4}
& 0.513
& 0.472
& 0.709
& 0.681
& 0.839
& 0.824 \\

& \textbf{Gemini-3-Flash}
& \underline{0.257}
& 0.251
& \underline{0.454}
& 0.443
& 0.644
& 0.641 \\

& \textbf{Gemini-3.1-Pro}
& \textbf{0.235}
& \textbf{0.204}
& \textbf{0.427}
& \textbf{0.386}
& \underline{0.641}
& \textbf{0.605} \\

& \textbf{Claude-Sonnet-4.6}
& \underline{0.257}
& \underline{0.207}
& 0.456
& \underline{0.396}
& 0.657
& \underline{0.612} \\

& \textbf{Qwen-3-VL-Instruct}
& 0.283
& 0.346
& 0.473
& 0.553
& \textbf{0.639}
& 0.708 \\

& \textbf{Llama-4-Scout}
& 0.336
& 0.371
& 0.541
& 0.589
& 0.693
& 0.738 \\

\midrule

\multirow{6}{*}{\textbf{Coaching Methods}}
& \textbf{GPT-5.4}
& 0.635
& 0.587
& 0.827
& 0.790
& 0.931
& 0.910 \\

& \textbf{Gemini-3-Flash}
& \underline{0.246}
& \underline{0.240}
& \underline{0.461}
& \underline{0.459}
& \underline{0.694}
& \textbf{0.692} \\

& \textbf{Gemini-3.1-Pro}
& 0.259
& \textbf{0.212}
& 0.481
& \textbf{0.438}
& 0.729
& \underline{0.700} \\

& \textbf{Claude-Sonnet-4.6}
& 0.301
& 0.255
& 0.538
& 0.491
& 0.781
& 0.753 \\

& \textbf{Qwen-3-VL-Instruct}
& \textbf{0.225}
& 0.279
& \textbf{0.439}
& 0.503
& \textbf{0.656}
& 0.707 \\

& \textbf{Llama-4-Scout}
& 0.306
& 0.329
& 0.547
& 0.578
& 0.744
& 0.768 \\

\bottomrule
\end{tabular}
\caption{Jensen-Shannon divergence (JSD) for dialogue act and coaching method $n$-gram distributions under Vanilla (V) and Coach (C) conditions. JSD is bounded in $[0, 1]$; lower values indicate closer alignment with human coach turns, with $0$ corresponding to identical distributions. Bold scores indicate the lowest value and underlined scores indicate the second-lowest value within each label type and column.}
\label{tab:ngram-jsd}
\end{table*}

\begin{table*}[t]
\centering
\small
\setlength{\tabcolsep}{4pt}
\renewcommand{\arraystretch}{1.08}
\resizebox{\textwidth}{!}{%
\begin{tabular}{lcccccccc}
\toprule
& & \multicolumn{2}{c}{\textbf{Modality}} 
& \multicolumn{3}{c}{\textbf{Context}} 
& \multicolumn{2}{c}{\textbf{Prompt}} \\
\cmidrule(lr){3-4}\cmidrule(lr){5-7}\cmidrule(lr){8-9}
\textbf{Model} 
& \textbf{Default} 
& \textbf{Text-Only} 
& \textbf{Image-Only} 
& \textbf{1s} 
& \textbf{10s} 
& \textbf{60s} 
& \textbf{Vanilla} 
& \textbf{Coach} \\
\midrule

\textbf{GPT-5.4}
& 0.0150$_{\pm \text{0.0010}}$
& 0.0150$_{\pm \text{0.0010}}$
& 0.0130$_{\pm \text{0.0010}}$
& 0.0130$_{\pm \text{0.0010}}$
& 0.0150$_{\pm \text{0.0010}}$
& 0.0150$_{\pm \text{0.0010}}$
& 0.0120$_{\pm \text{0.0010}}$
& 0.0120$_{\pm \text{0.0010}}$ \\

\textbf{Gemini-3-Flash}
& \underline{0.0180$_{\pm \text{0.0010}}$}
& 0.0180$_{\pm \text{0.0010}}$
& \underline{0.0150$_{\pm \text{0.0010}}$}
& \underline{0.0160$_{\pm \text{0.0010}}$}
& \underline{0.0170$_{\pm \text{0.0010}}$}
& \underline{0.0180$_{\pm \text{0.0010}}$}
& \underline{0.0140$_{\pm \text{0.0010}}$}
& \underline{0.0140$_{\pm \text{0.0010}}$} \\

\textbf{Gemini-3.1-Pro}
& \textbf{\boldmath 0.0250$_{\pm \text{0.0020}}$}
& \textbf{\boldmath 0.0280$_{\pm \text{0.0030}}$}
& \textbf{\boldmath 0.0170$_{\pm \text{0.0010}}$}
& \textbf{\boldmath 0.0210$_{\pm \text{0.0020}}$}
& \textbf{\boldmath 0.0230$_{\pm \text{0.0020}}$}
& \textbf{\boldmath 0.0270$_{\pm \text{0.0030}}$}
& \textbf{\boldmath 0.0180$_{\pm \text{0.0020}}$}
& \textbf{\boldmath 0.0190$_{\pm \text{0.0020}}$} \\

\textbf{Claude-Sonnet-4.6}
& 0.0170$_{\pm \text{0.0010}}$
& \underline{0.0220$_{\pm \text{0.0020}}$}
& 0.0140$_{\pm \text{0.0010}}$
& 0.0150$_{\pm \text{0.0010}}$
& \underline{0.0170$_{\pm \text{0.0020}}$}
& 0.0170$_{\pm \text{0.0010}}$
& 0.0130$_{\pm \text{0.0010}}$
& \underline{0.0140$_{\pm \text{0.0010}}$} \\

\textbf{Qwen-3-VL-Instruct}
& 0.0160$_{\pm \text{0.0010}}$
& 0.0150$_{\pm \text{0.0010}}$
& 0.0130$_{\pm \text{0.0010}}$
& 0.0140$_{\pm \text{0.0010}}$
& 0.0160$_{\pm \text{0.0010}}$
& 0.0160$_{\pm \text{0.0010}}$
& 0.0130$_{\pm \text{0.0010}}$
& 0.0130$_{\pm \text{0.0010}}$ \\

\textbf{Llama-4-Scout}
& 0.0160$_{\pm \text{0.0010}}$
& 0.0170$_{\pm \text{0.0010}}$
& 0.0130$_{\pm \text{0.0010}}$
& 0.0140$_{\pm \text{0.0010}}$
& 0.0150$_{\pm \text{0.0010}}$
& 0.0160$_{\pm \text{0.0010}}$
& 0.0110$_{\pm \text{0.0010}}$
& 0.0110$_{\pm \text{0.0010}}$ \\
\bottomrule
\end{tabular}%
}
\caption{BLEU scores (mean $\pm$ 95\% CI) for next coach utterance generation. The default setting uses text-image input, 30s context, and oracle prompt. Other columns report ablations over modality, context window, and prompt condition. Bold values indicate the best score per column, and underlined values indicate the second-highest score.}
\label{tab:bleu_ablation}
\end{table*}

\begin{table*}[t]
\centering
\small
\setlength{\tabcolsep}{4pt}
\renewcommand{\arraystretch}{1.08}
\resizebox{\textwidth}{!}{%
\begin{tabular}{lcccccccc}
\toprule
& & \multicolumn{2}{c}{\textbf{Modality}} 
& \multicolumn{3}{c}{\textbf{Context}} 
& \multicolumn{2}{c}{\textbf{Prompt}} \\
\cmidrule(lr){3-4}\cmidrule(lr){5-7}\cmidrule(lr){8-9}
\textbf{Model} 
& \textbf{Default} 
& \textbf{Text-Only} 
& \textbf{Image-Only} 
& \textbf{1s} 
& \textbf{10s} 
& \textbf{60s} 
& \textbf{Vanilla} 
& \textbf{Coach} \\
\midrule

\textbf{GPT-5.4}
& 0.136$_{\pm \text{0.004}}$
& 0.134$_{\pm \text{0.004}}$
& 0.120$_{\pm \text{0.004}}$
& 0.121$_{\pm \text{0.004}}$
& 0.133$_{\pm \text{0.004}}$
& 0.134$_{\pm \text{0.004}}$
& 0.105$_{\pm \text{0.004}}$
& 0.108$_{\pm \text{0.004}}$ \\

\textbf{Gemini-3-Flash}
& 0.140$_{\pm \text{0.005}}$
& 0.139$_{\pm \text{0.005}}$
& \underline{0.124$_{\pm \text{0.004}}$}
& 0.128$_{\pm \text{0.004}}$
& 0.138$_{\pm \text{0.005}}$
& 0.142$_{\pm \text{0.005}}$
& 0.115$_{\pm \text{0.004}}$
& 0.111$_{\pm \text{0.004}}$ \\

\textbf{Gemini-3.1-Pro}
& \textbf{\boldmath 0.165$_{\pm \text{0.006}}$}
& \textbf{\boldmath 0.171$_{\pm \text{0.006}}$}
& \textbf{\boldmath 0.126$_{\pm \text{0.005}}$}
& \textbf{\boldmath 0.143$_{\pm \text{0.005}}$}
& \textbf{\boldmath 0.158$_{\pm \text{0.006}}$}
& \textbf{\boldmath 0.169$_{\pm \text{0.006}}$}
& \textbf{\boldmath 0.136$_{\pm \text{0.005}}$}
& \textbf{\boldmath 0.136$_{\pm \text{0.005}}$} \\

\textbf{Claude-Sonnet-4.6}
& \underline{0.149$_{\pm \text{0.004}}$}
& \underline{0.164$_{\pm \text{0.005}}$}
& \textbf{\boldmath 0.126$_{\pm \text{0.004}}$}
& \underline{0.134$_{\pm \text{0.004}}$}
& \underline{0.147$_{\pm \text{0.005}}$}
& \underline{0.152$_{\pm \text{0.004}}$}
& \underline{0.126$_{\pm \text{0.004}}$}
& \underline{0.127$_{\pm \text{0.004}}$} \\

\textbf{Qwen-3-VL-Instruct}
& 0.129$_{\pm \text{0.004}}$
& 0.126$_{\pm \text{0.004}}$
& 0.112$_{\pm \text{0.004}}$
& 0.117$_{\pm \text{0.004}}$
& 0.128$_{\pm \text{0.004}}$
& 0.129$_{\pm \text{0.004}}$
& 0.105$_{\pm \text{0.004}}$
& 0.097$_{\pm \text{0.004}}$ \\

\textbf{Llama-4-Scout}
& 0.125$_{\pm \text{0.004}}$
& 0.134$_{\pm \text{0.004}}$
& 0.115$_{\pm \text{0.004}}$
& 0.120$_{\pm \text{0.004}}$
& 0.123$_{\pm \text{0.005}}$
& 0.121$_{\pm \text{0.005}}$
& 0.095$_{\pm \text{0.004}}$
& 0.089$_{\pm \text{0.004}}$ \\

\bottomrule
\end{tabular}%
}
\caption{ROUGE\mbox{-}L scores (mean $\pm$ 95\% CI) for next coach utterance generation. The default setting uses text-image input, 30s context, and oracle prompt. Other columns report ablations over modality, context window, and prompt condition. Bold values indicate the best score per column, and underlined values indicate the second-highest score.}
\label{tab:rouge_l_ablation}
\end{table*}

\begin{table*}[t]
\centering
\small
\setlength{\tabcolsep}{4pt}
\renewcommand{\arraystretch}{1.08}
\resizebox{\textwidth}{!}{%
\begin{tabular}{lcccccccc}
\toprule
& & \multicolumn{2}{c}{\textbf{Modality}} 
& \multicolumn{3}{c}{\textbf{Context}} 
& \multicolumn{2}{c}{\textbf{Prompt}} \\
\cmidrule(lr){3-4}\cmidrule(lr){5-7}\cmidrule(lr){8-9}
\textbf{Model} 
& \textbf{Default} 
& \textbf{Text-Only} 
& \textbf{Image-Only} 
& \textbf{1s} 
& \textbf{10s} 
& \textbf{60s} 
& \textbf{Vanilla} 
& \textbf{Coach} \\
\midrule

\textbf{GPT-5.4}
& 0.143$_{\pm \text{0.005}}$
& 0.141$_{\pm \text{0.005}}$
& \underline{0.127$_{\pm \text{0.005}}$}
& 0.127$_{\pm \text{0.005}}$
& 0.141$_{\pm \text{0.005}}$
& 0.144$_{\pm \text{0.005}}$
& 0.107$_{\pm \text{0.004}}$
& 0.106$_{\pm \text{0.004}}$ \\

\textbf{Gemini-3-Flash}
& 0.144$_{\pm \text{0.006}}$
& 0.146$_{\pm \text{0.005}}$
& 0.125$_{\pm \text{0.005}}$
& 0.130$_{\pm \text{0.005}}$
& 0.142$_{\pm \text{0.006}}$
& 0.147$_{\pm \text{0.006}}$
& 0.116$_{\pm \text{0.005}}$
& 0.109$_{\pm \text{0.005}}$ \\

\textbf{Gemini-3.1-Pro}
& \underline{0.164$_{\pm \text{0.007}}$}
& \underline{0.166$_{\pm \text{0.007}}$}
& 0.126$_{\pm \text{0.005}}$
& \underline{0.140$_{\pm \text{0.006}}$}
& \underline{0.157$_{\pm \text{0.007}}$}
& \underline{0.173$_{\pm \text{0.007}}$}
& \underline{0.138$_{\pm \text{0.006}}$}
& \underline{0.136$_{\pm \text{0.006}}$} \\

\textbf{Claude-Sonnet-4.6}
& \textbf{\boldmath 0.178$_{\pm \text{0.006}}$}
& \textbf{\boldmath 0.179$_{\pm \text{0.007}}$}
& \textbf{\boldmath 0.149$_{\pm \text{0.005}}$}
& \textbf{\boldmath 0.148$_{\pm \text{0.005}}$}
& \textbf{\boldmath 0.173$_{\pm \text{0.006}}$}
& \textbf{\boldmath 0.182$_{\pm \text{0.006}}$}
& \textbf{\boldmath 0.147$_{\pm \text{0.005}}$}
& \textbf{\boldmath 0.147$_{\pm \text{0.005}}$} \\

\textbf{Qwen-3-VL-Instruct}
& 0.125$_{\pm \text{0.005}}$
& 0.123$_{\pm \text{0.005}}$
& 0.108$_{\pm \text{0.004}}$
& 0.109$_{\pm \text{0.004}}$
& 0.121$_{\pm \text{0.005}}$
& 0.123$_{\pm \text{0.005}}$
& 0.098$_{\pm \text{0.005}}$
& 0.086$_{\pm \text{0.004}}$ \\

\textbf{Llama-4-Scout}
& 0.116$_{\pm \text{0.005}}$
& 0.142$_{\pm \text{0.005}}$
& 0.114$_{\pm \text{0.005}}$
& 0.127$_{\pm \text{0.005}}$
& 0.117$_{\pm \text{0.005}}$
& 0.113$_{\pm \text{0.005}}$
& 0.091$_{\pm \text{0.004}}$
& 0.079$_{\pm \text{0.004}}$ \\

\bottomrule
\end{tabular}%
}
\caption{METEOR scores (mean $\pm$ 95\% CI) for next coach utterance generation. The default setting uses text-image input, 30s context, and oracle prompt. Other columns report ablations over modality, context window, and prompt condition. Bold values indicate the best score per column, and underlined values indicate the second-highest score.}
\label{tab:meteor_ablation}
\end{table*}

\begin{table*}[t]
\centering
\small
\setlength{\tabcolsep}{4pt}
\renewcommand{\arraystretch}{1.08}
\resizebox{\textwidth}{!}{%
\begin{tabular}{lcccccccc}
\toprule
& & \multicolumn{2}{c}{\textbf{Modality}} 
& \multicolumn{3}{c}{\textbf{Context}} 
& \multicolumn{2}{c}{\textbf{Prompt}} \\
\cmidrule(lr){3-4}\cmidrule(lr){5-7}\cmidrule(lr){8-9}
\textbf{Model} 
& \textbf{Default} 
& \textbf{Text-Only} 
& \textbf{Image-Only} 
& \textbf{1s} 
& \textbf{10s} 
& \textbf{60s} 
& \textbf{Vanilla} 
& \textbf{Coach} \\
\midrule

\textbf{GPT-5.4}
& 0.840$_{\pm \text{0.001}}$
& 0.839$_{\pm \text{0.001}}$
& 0.836$_{\pm \text{0.001}}$
& 0.836$_{\pm \text{0.001}}$
& 0.840$_{\pm \text{0.001}}$
& 0.840$_{\pm \text{0.001}}$
& 0.834$_{\pm \text{0.001}}$
& 0.835$_{\pm \text{0.001}}$ \\

\textbf{Gemini-3-Flash}
& \underline{0.846$_{\pm \text{0.001}}$}
& \underline{0.845$_{\pm \text{0.001}}$}
& \underline{0.843$_{\pm \text{0.001}}$}
& \underline{0.844$_{\pm \text{0.001}}$}
& \underline{0.845$_{\pm \text{0.001}}$}
& \underline{0.846$_{\pm \text{0.001}}$}
& \underline{0.839$_{\pm \text{0.001}}$}
& \underline{0.839$_{\pm \text{0.001}}$} \\

\textbf{Gemini-3.1-Pro}
& \textbf{\boldmath 0.850$_{\pm \text{0.001}}$}
& \textbf{\boldmath 0.852$_{\pm \text{0.001}}$}
& \textbf{\boldmath 0.844$_{\pm \text{0.001}}$}
& \textbf{\boldmath 0.847$_{\pm \text{0.001}}$}
& \textbf{\boldmath 0.849$_{\pm \text{0.001}}$}
& \textbf{\boldmath 0.851$_{\pm \text{0.001}}$}
& \textbf{\boldmath 0.842$_{\pm \text{0.001}}$}
& \textbf{\boldmath 0.844$_{\pm \text{0.001}}$} \\

\textbf{Claude-Sonnet-4.6}
& 0.841$_{\pm \text{0.001}}$
& \underline{0.845$_{\pm \text{0.001}}$}
& 0.837$_{\pm \text{0.001}}$
& 0.840$_{\pm \text{0.001}}$
& 0.842$_{\pm \text{0.001}}$
& 0.842$_{\pm \text{0.001}}$
& 0.835$_{\pm \text{0.001}}$
& 0.836$_{\pm \text{0.001}}$ \\

\textbf{Qwen-3-VL-Instruct}
& 0.841$_{\pm \text{0.001}}$
& 0.840$_{\pm \text{0.001}}$
& 0.837$_{\pm \text{0.001}}$
& 0.838$_{\pm \text{0.001}}$
& 0.840$_{\pm \text{0.001}}$
& 0.840$_{\pm \text{0.001}}$
& 0.837$_{\pm \text{0.001}}$
& 0.837$_{\pm \text{0.001}}$ \\

\textbf{Llama-4-Scout}
& 0.840$_{\pm \text{0.001}}$
& 0.841$_{\pm \text{0.001}}$
& 0.837$_{\pm \text{0.001}}$
& 0.838$_{\pm \text{0.001}}$
& 0.839$_{\pm \text{0.001}}$
& 0.840$_{\pm \text{0.001}}$
& 0.836$_{\pm \text{0.001}}$
& 0.835$_{\pm \text{0.001}}$ \\

\bottomrule
\end{tabular}%
}
\caption{BERTScore scores (mean $\pm$ 95\% CI) for next coach utterance generation. The default setting uses text-image input, 30s context, and oracle prompt. Other columns report ablations over modality, context window, and prompt condition. Bold values indicate the best score per column, and underlined values indicate the second-highest score.}
\label{tab:bertscore_ablation}
\end{table*}

\begin{table*}[t]
\centering
\small
\setlength{\tabcolsep}{4pt}
\renewcommand{\arraystretch}{1.08}
\resizebox{\textwidth}{!}{%
\begin{tabular}{lcccccc}
\toprule
\textbf{Model}
& \textbf{Overall}
& \textbf{Figma}
& \textbf{Blender}
& \textbf{OnShape}
& \textbf{FL Studio}
& \textbf{Excel} \\
\midrule

\textbf{GPT-5.4}
& 32.4$_{\pm \text{1.3}}$
& 32.4$_{\pm \text{3.0}}$
& 29.8$_{\pm \text{2.9}}$
& 32.7$_{\pm \text{2.8}}$
& \underline{35.0$_{\pm \text{3.2}}$}
& 32.5$_{\pm \text{3.0}}$ \\

\textbf{Gemini-3-Flash}
& \underline{33.9$_{\pm \text{1.4}}$}
& \underline{36.8$_{\pm \text{3.3}}$}
& \underline{30.6$_{\pm \text{3.0}}$}
& \underline{35.1$_{\pm \text{3.1}}$}
& 33.7$_{\pm \text{3.4}}$
& \underline{33.2$_{\pm \text{3.1}}$} \\

\textbf{Gemini-3.1-Pro}
& \textbf{\boldmath 41.4$_{\pm \text{1.5}}$}
& \textbf{\boldmath 40.2$_{\pm \text{3.5}}$}
& \textbf{\boldmath 38.6$_{\pm \text{3.3}}$}
& \textbf{\boldmath 45.1$_{\pm \text{3.3}}$}
& \textbf{\boldmath 41.7$_{\pm \text{3.6}}$}
& \textbf{\boldmath 41.4$_{\pm \text{3.3}}$} \\

\textbf{Claude-Sonnet-4.6}
& 32.3$_{\pm \text{1.3}}$
& 32.4$_{\pm \text{3.0}}$
& 30.4$_{\pm \text{3.0}}$
& 34.1$_{\pm \text{3.0}}$
& 33.3$_{\pm \text{3.1}}$
& 31.5$_{\pm \text{2.9}}$ \\

\textbf{Qwen-3-VL-Instruct}
& 27.0$_{\pm \text{1.2}}$
& 26.2$_{\pm \text{2.7}}$
& 24.4$_{\pm \text{2.5}}$
& 28.0$_{\pm \text{2.5}}$
& 29.1$_{\pm \text{2.9}}$
& 27.7$_{\pm \text{2.8}}$ \\

\textbf{Llama-4-Scout}
& 19.5$_{\pm \text{1.0}}$
& 19.8$_{\pm \text{2.3}}$
& 17.3$_{\pm \text{2.0}}$
& 19.0$_{\pm \text{1.9}}$
& 21.9$_{\pm \text{2.5}}$
& 19.8$_{\pm \text{2.2}}$ \\

\bottomrule
\end{tabular}%
}
\caption{CLAIR scores (mean $\pm$ 95\% CI) by domain under the text+image input, 30s, Oracle condition. Bold values indicate the best score per column, and underlined values indicate the second-highest score.}
\label{tab:domain_clair_base}
\end{table*}

\begin{table*}[t]
\centering
\small
\setlength{\tabcolsep}{4pt}
\renewcommand{\arraystretch}{1.08}
\resizebox{\textwidth}{!}{%
\begin{tabular}{lccccccccccc}
\toprule
\textbf{Model}
& \textbf{Overall}
& \textbf{Artic.}
& \textbf{Clarif.}
& \textbf{Confirm.}
& \textbf{Diagn.}
& \textbf{Direct}
& \textbf{Explan.}
& \textbf{Explor.}
& \textbf{Plan}
& \textbf{Reflect.}
& \textbf{Tip} \\
\midrule

\textbf{GPT-5.4}
& 32.4$_{\pm \text{1.3}}$
& 35.6$_{\pm \text{5.1}}$
& 23.4$_{\pm \text{3.6}}$
& \underline{41.9$_{\pm \text{4.3}}$}
& 23.2$_{\pm \text{3.5}}$
& 32.1$_{\pm \text{4.1}}$
& \underline{38.6$_{\pm \text{4.3}}$}
& 33.6$_{\pm \text{4.1}}$
& 35.6$_{\pm \text{4.1}}$
& 29.1$_{\pm \text{3.9}}$
& 31.7$_{\pm \text{4.1}}$ \\

\textbf{Gemini-3-Flash}
& \underline{33.9$_{\pm \text{1.4}}$}
& \underline{40.6$_{\pm \text{5.4}}$}
& \underline{27.0$_{\pm \text{4.3}}$}
& 37.6$_{\pm \text{4.4}}$
& \underline{30.0$_{\pm \text{4.4}}$}
& 33.2$_{\pm \text{4.5}}$
& 37.4$_{\pm \text{4.5}}$
& \underline{37.2$_{\pm \text{4.2}}$}
& \underline{38.4$_{\pm \text{4.6}}$}
& \underline{30.3$_{\pm \text{4.2}}$}
& 29.0$_{\pm \text{4.3}}$ \\

\textbf{Gemini-3.1-Pro}
& \textbf{\boldmath 41.4$_{\pm \text{1.5}}$}
& \textbf{\boldmath 48.8$_{\pm \text{5.7}}$}
& \textbf{\boldmath 34.0$_{\pm \text{4.7}}$}
& \textbf{\boldmath 51.4$_{\pm \text{4.8}}$}
& \textbf{\boldmath 34.0$_{\pm \text{4.4}}$}
& \textbf{\boldmath 39.9$_{\pm \text{4.8}}$}
& \textbf{\boldmath 42.8$_{\pm \text{4.7}}$}
& \textbf{\boldmath 46.6$_{\pm \text{4.5}}$}
& \textbf{\boldmath 43.7$_{\pm \text{4.6}}$}
& \textbf{\boldmath 37.0$_{\pm \text{4.9}}$}
& \textbf{\boldmath 37.9$_{\pm \text{4.6}}$} \\

\textbf{Claude-Sonnet-4.6}
& 32.3$_{\pm \text{1.3}}$
& 39.3$_{\pm \text{4.9}}$
& 24.4$_{\pm \text{3.8}}$
& 36.1$_{\pm \text{4.3}}$
& 27.7$_{\pm \text{4.2}}$
& \underline{33.7$_{\pm \text{4.3}}$}
& 36.0$_{\pm \text{4.2}}$
& 34.5$_{\pm \text{4.2}}$
& 33.3$_{\pm \text{4.2}}$
& 28.2$_{\pm \text{3.8}}$
& \underline{32.1$_{\pm \text{4.1}}$} \\

\textbf{Qwen-3-VL-Instruct}
& 27.0$_{\pm \text{1.2}}$
& 28.1$_{\pm \text{4.5}}$
& 19.5$_{\pm \text{2.9}}$
& 33.0$_{\pm \text{4.1}}$
& 19.7$_{\pm \text{2.8}}$
& 24.4$_{\pm \text{3.5}}$
& 32.3$_{\pm \text{4.1}}$
& 28.7$_{\pm \text{3.7}}$
& 33.2$_{\pm \text{4.0}}$
& 26.0$_{\pm \text{3.5}}$
& 25.8$_{\pm \text{3.7}}$ \\

\textbf{Llama-4-Scout}
& 19.5$_{\pm \text{1.0}}$
& 24.3$_{\pm \text{4.1}}$
& 15.3$_{\pm \text{2.7}}$
& 17.1$_{\pm \text{2.8}}$
& 15.0$_{\pm \text{2.4}}$
& 17.6$_{\pm \text{2.8}}$
& 20.3$_{\pm \text{3.2}}$
& 25.3$_{\pm \text{3.6}}$
& 24.9$_{\pm \text{3.6}}$
& 17.0$_{\pm \text{2.4}}$
& 19.5$_{\pm \text{2.7}}$ \\

\bottomrule
\end{tabular}%
}
\caption{CLAIR scores (mean $\pm$ 95\% CI) by coaching method under the Text+Image, 30s, Oracle condition. Bold values indicate the best score per column, and underlined values indicate the second-highest score.}
\label{tab:method_clair_base}
\end{table*}

\subsection{Human Expert Evaluation Setup}
\label{sec:expert-eval-setup}

\noindent To ensure a fair comparison, all candidate utterances and prior chat messages are converted to audio via a AI text-to-speech voice.
The human expert evaluation interface screenshot is shown in Figure~\ref{fig:annotation_interface}.
\begin{figure}[t]
    \centering
    \includegraphics[width=\columnwidth]{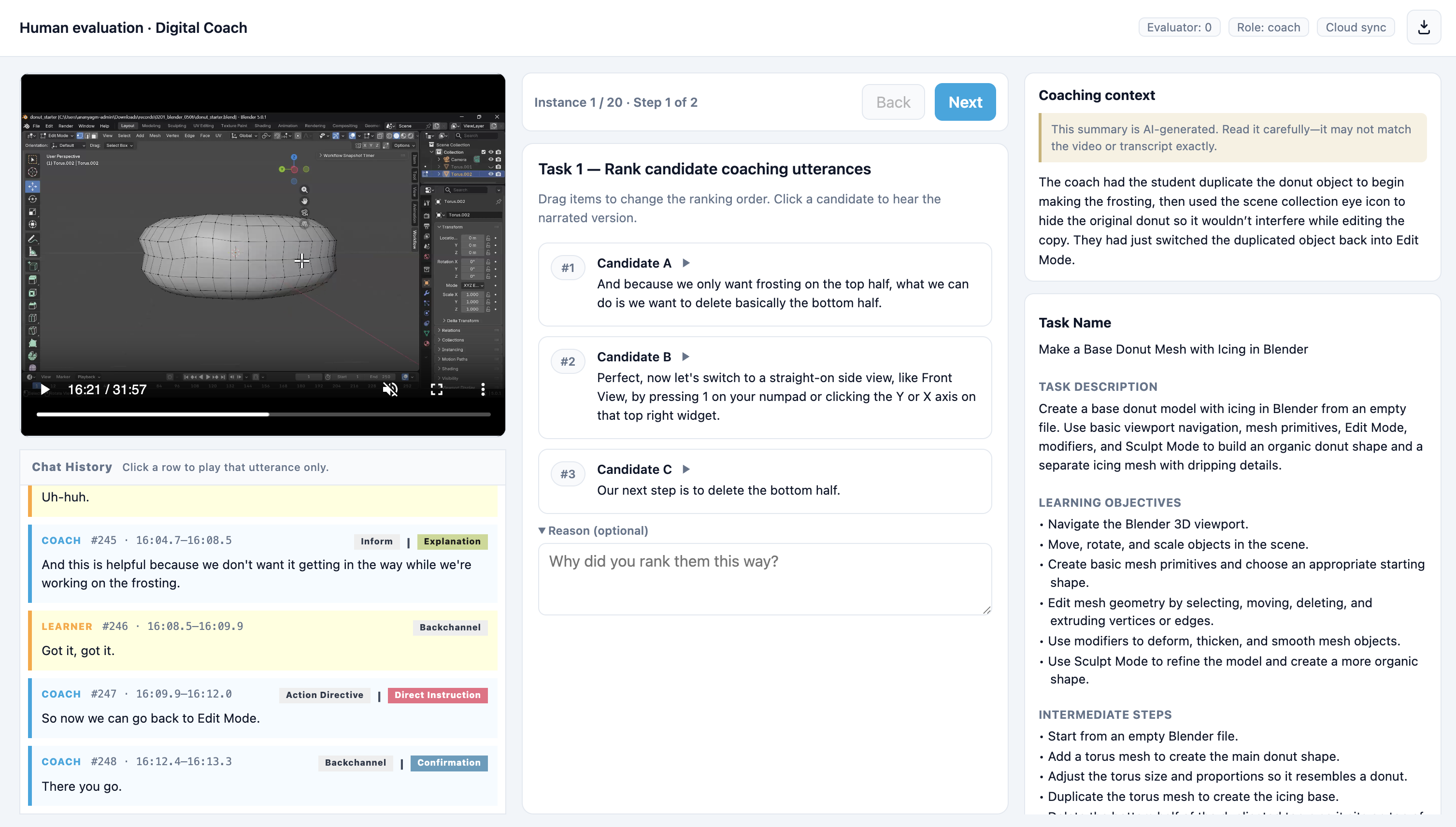}
    \caption{
     Human expert evaluation interface. Raters can see the the previous conversation and screen context and listen to AI-narrated candidate coaching utterances.
    }
    \label{fig:expert-eval-interface}
\end{figure}

\subsection{Interactive Evaluation Setup}
\label{sec:ix-eval-setup}
\begin{figure}[t]
    \centering
    \includegraphics[width=\columnwidth]{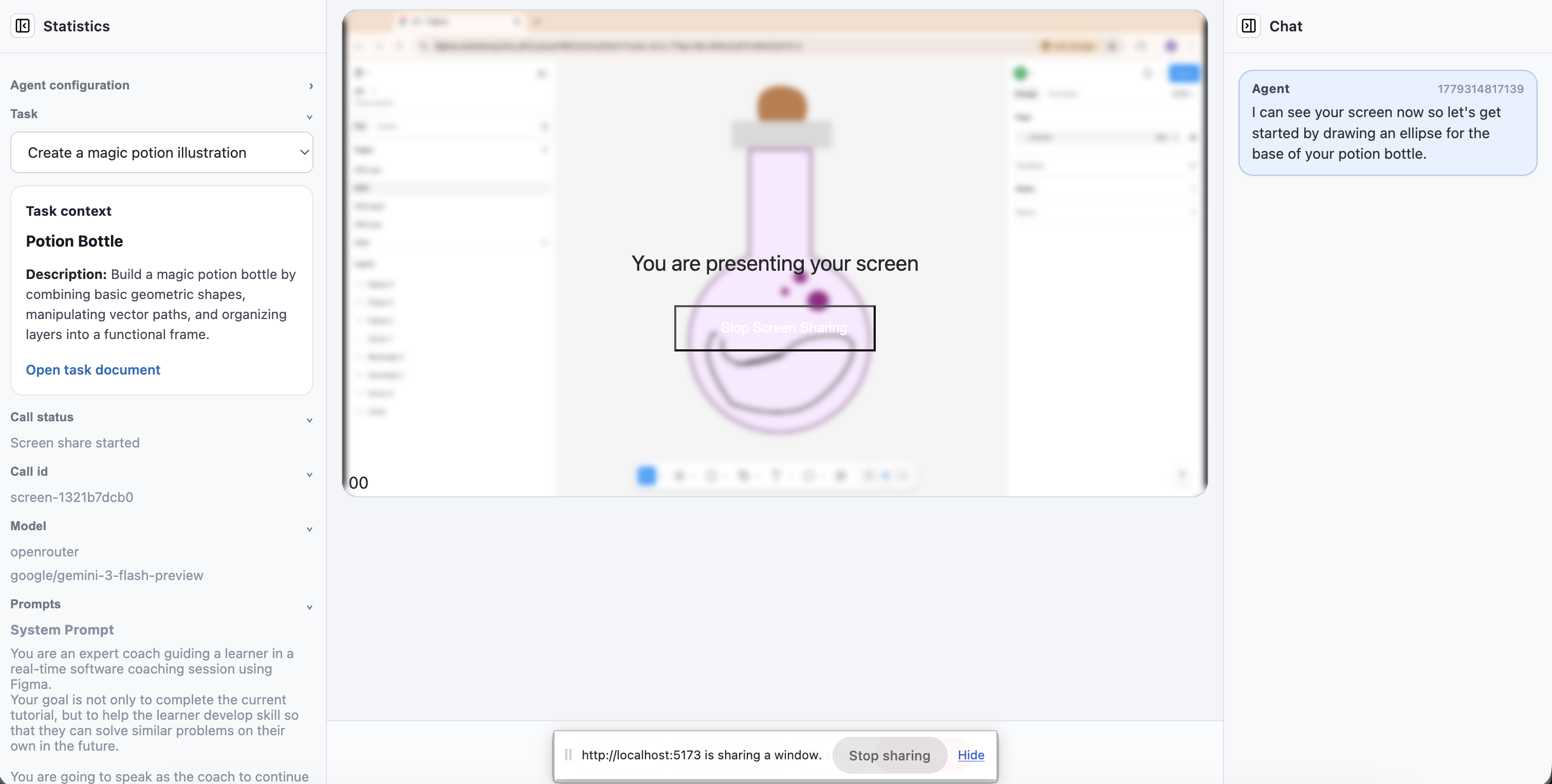}
    \caption{
    Interactive evaluation interface. A novice shares their screen and speaks over a live call. Model coach observes recent frames and dialogue and replies with audio.
    }
    \label{fig:ix-eval-interface}
\end{figure}

\noindent Interactive evaluation system uses a browser client (React, Stream Video WebRTC) paired with a Python backend (Vision agents, FastAPI). A novice shares screen and audio, and the software coach agent joins the same call server-side (Figure~\ref{fig:ix-eval-interface}). By default, we route model inference through OpenRouter, Deepgram speech-to-text, and ElevenLabs text-to-speech. Proactive check-ins fire every 20s. Sessions log model--learner transcripts for offline analysis.

\newtext{We tested Gemini-3.1-Pro with two participants in our pilot study, but its high latency made real-time interaction difficult as instructions and feedback were often out of sync with user interactions. Prior research~\cite{miller_response_1968} confirms that latency > 4s degrades quality of experience for conversational agents.
Using the same system prompt and screen frames from the preceding 10 seconds at 1 fps, we ran each model 10 times. Gemini-3.1-Pro took approximately 270\% as long per turn as Gemini-3-Flash ($\mu$ = 6.5s, $\sigma$ = 0.7 vs. $\mu$ = 2.4s, $\sigma$ = 0.6). As a model coach session has 61 coaching turns on average, using Gemini-3.1-Pro will result in an additional gap of 250s between turns, making real-time coaching difficult.}

\section{Prompts}
\label{sec:prompts}
\noindent Figure~\ref{fig:autocorrect-transcription-prompt} shows the automatic transcript correction prompt. 
Figure~\ref{fig:dialogue-act-annotation-prompt} shows the automatic dialogue act annotation prompt. 
Figure~\ref{fig:coaching-method-annotation-prompt} shows the automatic coaching method annotation prompt. 
Figure~\ref{fig:next-coach-utterance-prompt} shows the next coach utterance generation prompt.

\begin{figure*}[t]
\centering
\fbox{%
\begin{minipage}{0.94\textwidth}
\vspace{0.8em}

{\normalsize\textbf{Auto-Correct Transcription Prompt}}

\vspace{0.4em}
\hrule
\vspace{0.2em}

\small
\ttfamily
\begin{flushleft}
You are aligning a reference transcript to the actual speech in the attached audio.\\

\vspace{0.6em}
The JSON below is the reference transcript for this recording. Each object has timing, speaker labels, and a "text" field.\\
The "text" values may contain wrong words, homophone errors, dropped or extra words, or other mismatches with what is actually said.\\
Your task is to correct the "text" field to match what is actually said contextually.\\

\vspace{0.6em}
Task:\\
1. Listen to the audio carefully.\\
2. For each object \textbf{in array order}, produce the wording that best matches what is spoken\\
\hspace*{1em}in the corresponding time range. Fix mistranscriptions only.\\
3. Do \textbf{NOT} invent new utterances, merge, or split segments. The number of strings you\\
\hspace*{1em}output must equal \{n\} (one corrected line per input object).\\
4. Do \textbf{NOT} change punctuation style unnecessarily; natural English is fine.\\
5. ONLY correct the "text" field; keep the same timing, speaker, and id.\\
6. Use , to indicate short speech pauses or hesitation.\\
7. Use ... to indicate a longer pause or hesitation.\\
8. Use -- to indicate a trail off or being interrupted by another speaker.\\
9. The first letter of the first word in the "text" field should be capitalized.

\vspace{0.6em}
Output requirements:\\
- Return ONLY valid JSON: a single array of \{n\} strings.\\
- String i is the corrected "text" for input object i (same order as the reference array).\\
- No markdown, no code fences, no keys other than the array itself.\\

\vspace{0.6em}
Good Examples:\\
- "bit, there's pill." -> "bit, there's fill."\\
- "Cool. Mom. Gone." -> "Cool. Mmmmm. Done."\\
- "It does sometimes take a second, so baby... huh." ->\\
\hspace*{1em}"It does sometimes take a second, so maybe... huh."\\

\vspace{0.6em}
Reference transcript (do not echo it back; only output the string array):\\

\vspace{0.4em}
\{reference\_transcript\}
\end{flushleft}

\vspace{0.8em}
\end{minipage}%
}
\caption{Prompt used for automatic transcript correction.}
\label{fig:autocorrect-transcription-prompt}
\end{figure*}

\begin{figure*}[t]
\centering
\fbox{%
\begin{minipage}{0.94\textwidth}
\vspace{0.8em}

{\normalsize\textbf{Dialogue Act Annotation Prompt}}

\vspace{0.4em}
\hrule
\vspace{0.2em}

\small
\ttfamily
\begin{flushleft}
You are a linguistics expert annotating utterances from a task-oriented instructional dialogue.\\

\vspace{0.6em}
Your job is to read each utterance from both Coach and Learner in context and decide what dialogue act the speaker is trying to do.\\

\vspace{0.6em}
Choose all applicable labels from the schema below. Use the full transcript for context.\\

\vspace{0.6em}
Annotation Schema:\\
\{\{DIALOGUE\_ACTS\}\}\\

\vspace{0.6em}
Instructions:\\
For each turn, return a JSON array of objects with these fields:\\
- \texttt{"turn\_id"}: the turn number as an integer\\
- \texttt{"text"}: the original text of the turn\\
- \texttt{"dialogue\_acts"}: an array of all applicable labels from the schema above\\

\vspace{0.6em}
Output requirements:\\
- Return ONLY valid JSON as an array of objects.\\
- Do not include markdown fences.\\
- Do not include any extra text.\\

\vspace{0.6em}
Transcript:\\

\vspace{0.4em}
\{\{TRANSCRIPT\}\}
\end{flushleft}

\vspace{0.8em}
\end{minipage}%
}
\caption{Prompt used for automatic dialogue act annotation.}
\label{fig:dialogue-act-annotation-prompt}
\end{figure*}

\begin{figure*}[t]
\centering
\fbox{%
\begin{minipage}{0.94\textwidth}
\vspace{0.8em}

{\normalsize\textbf{Coaching Method Annotation Prompt}}

\vspace{0.4em}
\hrule
\vspace{0.2em}

\small
\ttfamily
\begin{flushleft}
You are an expert annotator annotating coach utterances from a task-oriented instructional dialogue.\\

\vspace{0.6em}
Your job is to read each coach turn in context and decide which coaching method labels are used.\\

\vspace{0.6em}
Choose all applicable labels from the schema below. Use the full transcript for context to understand what the coach is responding to, but only annotate Coach turns.\\

\vspace{0.6em}
Annotation Schema:\\
\{\{COACHING\_METHODS\}\}\\

\vspace{0.6em}
Instructions:\\
For each coach turn, return a JSON array of objects with these fields:\\
- \texttt{"turn\_id"}: the turn number as an integer\\
- \texttt{"text"}: the original text of the turn\\
- \texttt{"coaching\_methods"}: an array of all applicable labels from the annotation classes\\

\vspace{0.6em}
Output requirements:\\
- Return ONLY valid JSON as an array of objects.\\
- Do not annotate Learner turns.\\
- Do not include markdown fences.\\
- Do not include any extra text.\\

\vspace{0.6em}
Transcript:\\

\vspace{0.4em}
\{\{TRANSCRIPT\}\}
\end{flushleft}

\vspace{0.8em}
\end{minipage}%
}
\caption{Prompt used for automatic coaching method annotation.}
\label{fig:coaching-method-annotation-prompt}
\end{figure*}

\begin{figure*}[t]
\centering
\fbox{%
\begin{minipage}{0.94\textwidth}
\vspace{0.8em}

{\normalsize\textbf{Next Coach Utterance Generation Prompt }}

\vspace{0.4em}
\hrule
\vspace{0.2em}

\small
\ttfamily
\noindent\makebox[\linewidth][c]{%
\colorbox{red!8}{%
\begin{minipage}{\dimexpr0.96\linewidth-2\fboxsep\relax}
\vspace{0.6em}
\textbf{System message:}

\vspace{0.5em}
You are an expert coach guiding a learner in a real-time computer use coaching session using \{\{SOFTWARE\_TOOL\}\}.\\
Your goal is not only to complete the current tutorial, but to help the learner develop skill so that they can solve similar problems on their own in the future.\\

\vspace{0.6em}
You are going to speak as the coach to continue the conversation.\\
\textcolor{blue}{You can choose the appropriate coaching method(s) defined in the following section to use in your response.}\\
You are looking at the learner's recent screen and you have your recent conversation with the learner.\\
The cursor is where the learner is currently interacting with the software.\\
If there are any visual pink color screen annotations, those are your previous annotations to the learner's screen. You can use them to guide your response.\\
Your response should sound like something a human coach would actually say aloud in real-time during the session.\\

\vspace{0.6em}
Generation Rules:\\
- Give one concise sentence as your response.\\
- Treat the chat history as a live conversation (user = Learner, assistant = Coach).\\
- Avoid overly verbose and overly formal language.\\
- Avoid any special characters or markdown formatting like lists, bold, italic, etc.\\
- Avoid transcript artifacts such as ellipses, dashes.\\
- Do not mention being an AI, model, assistant, or system.\\

\vspace{0.6em}
Output Rules:\\
- Return ONLY the coach utterance as plain text.\\
- No JSON, no speaker prefix, no markdown.\\

\vspace{0.6em}
\textcolor{blue}{Coaching Method Definitions and Examples:}\\
\textcolor{blue}{\{\{COACHING\_METHODS\_DEFINITION\_AND\_EXAMPLES\}\}}\\

\vspace{0.6em}
Task Description:\\
\{\{TASK\_DESCRIPTION\}\}
\end{minipage}%
}%
}

\vspace{0.3em}

\noindent\makebox[\linewidth][c]{%
\colorbox{orange!12}{%
\begin{minipage}{\dimexpr0.96\linewidth-2\fboxsep\relax}
\vspace{0.6em}
\textbf{User message: Dialogue History}

\vspace{0.5em}
Recent \{\{CONTEXT\_WINDOW\}\} dialogue history (ordered from earliest to most recent): \\

\{\{Learner turns are shown as User messages. Coach turns are shown as Assistant messages.\}\}
\end{minipage}%
}%
}

\vspace{0.3em}

\noindent\makebox[\linewidth][c]{%
\colorbox{yellow!18}{%
\begin{minipage}{\dimexpr0.96\linewidth-2\fboxsep\relax}
\vspace{0.6em}
\textbf{User message: Visual Context}

\vspace{0.5em}
Recent \{\{CONTEXT\_WINDOW\}\} learner's recent screen (ordered from earliest to most recent):\\

\vspace{0.4em}
\{\{RECENT\_SCREEN\_FRAMES\}\}
\end{minipage}%
}%
}

\vspace{0.3em}

\noindent\makebox[\linewidth][c]{%
\colorbox{blue!8}{%
\begin{minipage}{\dimexpr0.96\linewidth-2\fboxsep\relax}
\vspace{0.6em}
\textbf{User message: Final Instruction}

\vspace{0.5em}
You are going to speak as the coach to continue the conversation.\\
\textcolor{orange}{Choose the most appropriate coaching method(s) and give your coaching utterance. ONLY return the coaching utterance, no coaching method tags.}\\
\textcolor{red}{Your next coach utterance should use the following coaching method(s):\\
\{\{COACHING\_METHODS\}\}}
\vspace{0.6em}
\end{minipage}%
}%
}

\end{minipage}%
}
\caption{Prompt structure for next coach utterance generation, including \colorbox{red!8}{system prompt}, \colorbox{orange!12}{recent dialogue}, \colorbox{yellow!18}{recent screen context}, and \colorbox{blue!8}{final instruction}. Colored text marks prompt condition additions: \textcolor{blue}{blue} for both Coach and Oracle prompts, \textcolor{orange}{orange} for Coach only, and \textcolor{red}{red} for Oracle only.}
\label{fig:next-coach-utterance-prompt}
\end{figure*}


\section{Additional Human Coaching Results}
\label{sec:add-learn-outcome-results}
\subsection{Problematic Outcome Tags}
\noindent Table~\ref{tab:learning-tags-human-model} shows problematic outcome tags for human and model coaching sessions.

\begin{table*}[t]
\centering
\small
\setlength{\tabcolsep}{4pt}
\renewcommand{\arraystretch}{1.08}

\begin{tabular}{llccc}
\toprule
\textbf{Tag} 
& \textbf{Setup} 
& \textbf{Tutorial Task} 
& \textbf{Pre-Task} 
& \textbf{Post-Task} \\
\midrule
\multirow{2}{*}{Gave up}
& H & 0 (0.0\%)    & 19 (26.4\%) & 0 (0.0\%) \\
& M & 15 (41.7\%) & 31 (86.1\%) & 17 (47.2\%) \\
\midrule
\multirow{2}{*}{Wrong method}
& H & 0 (0.0\%)   & 17 (23.6\%) & 2 (2.8\%) \\
& M & 3 (8.3\%)  & 7 (19.4\%)  & 2 (5.6\%) \\
\midrule
\multirow{2}{*}{Mistakes}
& H & 0 (0.0\%)   & 3 (4.2\%)  & 5 (6.9\%) \\
& M & 5 (13.9\%) & 0 (0.0\%)  & 7 (19.4\%) \\
\midrule
\multirow{2}{*}{Missed steps}
& H & 0 (0.0\%)   & 2 (2.8\%)  & 8 (11.1\%) \\
& M & 7 (19.4\%) & 0 (0.0\%)  & 9 (25.0\%) \\
\bottomrule
\end{tabular}

\caption{Problematic outcome tags for human coaching sessions (H; $N=72$) and model coaching sessions (M; $N=36$). Tags are not mutually exclusive.}
\label{tab:learning-tags-human-model}
\end{table*}

\subsection{Coaching Methods and Learning Outcome}
\begin{table*}[t]
\centering
\small

\resizebox{\textwidth}{!}{%
\begin{tabular}{@{}lllrrrrrrrrrrrrrrrrr@{}}
\toprule
\textbf{Condition} &
\textbf{Session} &
\textbf{Software} &
\textbf{\# Labels} &
\textbf{DI} &
\textbf{P} &
\textbf{Conf} &
\textbf{Diag} &
\textbf{Expl} &
\textbf{Tip} &
\textbf{Clar} &
\textbf{Refl} &
\textbf{Art} &
\textbf{Explr} &
\textbf{NI} &
\textbf{$\Delta T_1$} &
\textbf{$\Delta T_2$} &
\textbf{$\Delta T_3$} &
\textbf{$\Delta T_4$} &
\textbf{Mean $\Delta$} \\
\midrule
Human--Human & C1L1   & Figma     & 479  & \textbf{29.2} & 9.8  & \underline{15.7} & 7.7  & 12.3 & 5.8 & 4.8 & 1.5 & 0.0 & 3.3 & 9.8 & 100 & 100 & 100 & 80  & 95 \\
Human--Human & C2L2   & Figma     & 1519 & \textbf{37.7} & 7.2  & \underline{20.9} & 6.5  & 13.0 & 2.5 & 1.8 & 0.7 & 0.3 & 3.3 & 6.1 & 100 & 80  & 20  & 40  & 60 \\
Human--Human & C3L3   & Blender   & 856  & \textbf{26.6} & 5.1  & 19.0 & 9.6  & \underline{21.5} & 6.9 & 1.2 & 1.3 & 0.0 & 2.3 & 6.4 & 45  & 100 & 0   & 100 & 61.2 \\
Human--Human & C4L4   & Figma     & 807  & \textbf{25.1} & 7.8  & \underline{19.9} & 7.9  & 17.8 & 4.1 & 4.0 & 0.6 & 0.1 & 4.5 & 8.1 & 0   & 60  & 0   & 100 & 40 \\
Human--Human & C5L5   & Blender   & 1213 & \textbf{35.3} & 5.4  & 11.5 & 7.8  & \underline{19.7} & 5.4 & 1.9 & 0.6 & 0.0 & 3.4 & 9.1 & 35  & 10  & 100 & 25  & 42.5 \\
Human--Human & C6L6   & Blender   & 1217 & \textbf{33.8} & 7.6  & \underline{18.9} & 7.7  & 15.2 & 3.9 & 1.6 & 0.3 & 1.7 & 2.1 & 7.1 & 30  & 0   & 10  & 100 & 35 \\
Human--Human & C7L7   & Blender   & 1321 & \textbf{32.5} & 5.2  & 13.0 & 7.0  & \underline{24.0} & 3.9 & 3.3 & 0.8 & 0.0 & 2.8 & 7.5 & 70  & 90  & 30  & 45  & 58.8 \\
Human--Human & C8L8   & OnShape   & 944  & \underline{27.0} & 9.0  & 13.7 & 8.6  & \textbf{28.1} & 6.7 & 1.1 & 0.6 & 0.1 & 2.3 & 2.9 & 25  & 90  & 80  & --  & 65 \\
Human--Human & C9L9   & OnShape   & 1146 & \textbf{29.2} & 7.0  & \underline{19.6} & 10.7 & 14.8 & 1.9 & 4.1 & 3.5 & 2.5 & 1.5 & 5.2 & 100 & 90  & 30  & --  & 73.3 \\
Human--Human & C10L10 & OnShape   & 900  & \textbf{30.8} & 8.9  & 20.0 & 5.6  & \underline{21.2} & 5.1 & 1.2 & 0.9 & 1.6 & 1.6 & 3.2 & 100 & 50  & 50  & --  & 66.7 \\
Human--Human & C11L11 & OnShape   & 976  & \textbf{34.4} & 5.9  & \underline{25.1} & 8.7  & 13.6 & 2.4 & 3.4 & 1.3 & 1.0 & 0.7 & 3.4 & 60  & 80  & 20  & --  & 53.3 \\
Human--Human & C12L12 & FL Studio & 712  & \textbf{34.1} & 5.9  & 11.4 & 5.8  & \underline{31.5} & 5.9 & 0.3 & 0.1 & 0.0 & 1.1 & 3.9 & 100 & 0   & 0   & 60  & 40 \\
Human--Human & C13L13 & Figma     & 820  & \textbf{38.2} & 7.8  & 14.2 & 6.0  & \underline{16.5} & 3.9 & 2.1 & 1.3 & 0.8 & 3.0 & 6.2 & 0   & 75  & 60  & 100 & 58.8 \\
Human--Human & C14L14 & FL Studio & 588  & \textbf{36.0} & 5.6  & 14.5 & 4.8  & \underline{20.6} & 4.4 & 4.1 & 0.3 & 0.0 & 1.7 & 8.0 & 100 & 0   & 20  & 0   & 30 \\
Human--Human & C15L15 & FL Studio & 644  & \textbf{34.9} & 7.1  & 19.4 & 4.2  & \underline{20.8} & 5.3 & 1.7 & 0.2 & 0.2 & 1.4 & 4.8 & 65  & 0   & 0   & 100 & 41.2 \\
Human--Human & C16L16 & FL Studio & 1026 & \underline{24.3} & 5.8  & 17.5 & 6.3  & \textbf{27.3} & 7.1 & 2.5 & 0.7 & 0.1 & 2.6 & 5.8 & 100 & 0   & 0   & 83  & 45.8 \\
Human--Human & C17L17 & Excel     & 664  & \textbf{32.5} & 8.9  & 17.5 & 9.2  & \underline{19.3} & 4.7 & 1.4 & 0.0 & 0.0 & 0.8 & 5.9 & 34  & 100 & 40  & --  & 58 \\
Human--Human & C18L18 & Excel     & 539  & \textbf{28.9} & 10.2 & 12.8 & 8.5  & \underline{22.4} & 5.0 & 2.8 & 3.1 & 0.7 & 1.5 & 3.9 & 70  & 100 & 5   & --  & 58.3 \\
Human--Human & C19L19 & Excel     & 699  & \underline{19.7} & 10.2 & \textbf{20.2} & 8.7 & 18.7 & 6.9 & 1.4 & 1.3 & 1.1 & 2.7 & 9.0 & 80  & 100 & 10  & --  & 63.3 \\
Human--Human & C20L20 & Excel     & 913  & \textbf{26.9} & 8.8  & \underline{16.6} & 10.4 & 14.3 & 2.6 & 3.4 & 2.3 & 4.7 & 2.2 & 7.7 & 85  & 100 & 10  & --  & 65 \\
\midrule
Human--Model & P1  & Figma     & 406 & \textbf{49.5} & 8.6  & \underline{13.3} & 4.2  & 13.1 & 3.9 & 2.2 & 2.0 & 0.0 & 2.2 & 1.0 & 0   & 0   & 0  & 0   & 0 \\
Human--Model & P2  & Blender   & 380 & \textbf{51.0} & 6.0  & \underline{18.7} & 5.8  & 16.3 & 1.6 & 0.0 & 0.0 & 0.0 & 0.5 & 0.0 & 0   & 100 & 80 & 100 & 70 \\
Human--Model & P3  & OnShape   & 281 & \textbf{56.6} & 12.5 & \underline{13.9} & 5.3  & 8.2  & 1.8 & 0.4 & 1.1 & 0.0 & 0.4 & 0.0 & 10  & 0   & 0  & --  & 3.3 \\
Human--Model & P4  & FL Studio & 343 & \textbf{50.7} & 7.9  & \underline{23.6} & 5.5  & 10.2 & 1.2 & 0.0 & 0.0 & 0.0 & 0.9 & 0.0 & 40  & 50  & 0  & 0   & 22.5 \\
Human--Model & P5  & Excel     & 271 & \textbf{41.0} & 12.2 & 16.2 & 10.0 & \underline{18.8} & 0.7 & 0.7 & 0.0 & 0.0 & 0.4 & 0.0 & 80  & 0   & 60 & --  & 46.7 \\
Human--Model & P6  & Figma     & 367 & \textbf{46.0} & 11.2 & \underline{21.8} & 1.6  & 10.3 & 2.7 & 0.0 & 1.1 & 1.1 & 3.8 & 0.3 & -70 & 30  & 0  & 0   & -10 \\
Human--Model & P7  & Blender   & 252 & \textbf{55.6} & 3.6  & \underline{17.9} & 8.7  & 10.7 & 2.8 & 0.0 & 0.4 & 0.0 & 0.4 & 0.0 & 80  & 100 & 30 & 100 & 77.5 \\
Human--Model & P8  & OnShape   & 487 & \textbf{48.5} & 4.5  & \underline{25.7} & 4.9  & 15.4 & 0.8 & 0.0 & 0.0 & 0.0 & 0.0 & 0.2 & 60  & 20  & 40 & --  & 40 \\
Human--Model & P9  & FL Studio & 507 & \textbf{47.1} & 8.1  & \underline{19.9} & 3.9  & 8.3  & 2.8 & 0.2 & 5.1 & 0.0 & 4.5 & 0.0 & 10  & 30  & 0  & 0   & 10 \\
Human--Model & P10 & Excel     & 304 & \textbf{38.5} & \underline{16.4} & 16.4 & 7.9 & 16.4 & 1.0 & 0.7 & 0.3 & 0.0 & 1.3 & 1.0 & 50 & 100 & 40 & -- & 63.3 \\
\bottomrule
\end{tabular}%
}
\caption{Coaching methods and learning outcome for each Human--Human and Human--Model session. Coaching method values are percentages. $\Delta T_i$ denotes the learner's learning gain on task $T_i$, computed as post-task progress minus pre-task progress in percentage. DI = Direct Instruction, P = Plan, Conf = Confirmation, Diag = Diagnosis, Expl = Explanation, Tip = Tip, Clar = Clarification, Refl = Reflection, Art = Articulation, Explr = Exploration, and NI = Non-Instructive. Bold and underline indicate the most and second-most frequent coaching methods within each session.}
\label{tab:session-coaching-methods}
\end{table*}

\noindent Table~\ref{tab:session-coaching-methods} details coaching methods and learning outcomes for each Human--Human and Human--Model session.

\subsection{Pre/Post Task Outcome Examples}
\noindent Table~\ref{tab:learning_outcomes} shows examples of pre-task and post-task outcomes in visual tasks (Figma, Blender, OnShape).

\begin{table*}[p]
    \centering
    \small
    \renewcommand{\arraystretch}{1.2}
    \setlength{\tabcolsep}{5pt}
    
    \renewcommand{\tabularxcolumn}[1]{m{#1}}

    \begin{tabularx}{\textwidth}{@{}cccCCC@{}}
        \toprule
        \textbf{Software} & \textbf{Task ID} & \textbf{Session ID} & \textbf{Target} & \textbf{Pre-Task} & \textbf{Post-Task} \\
        \midrule
        
        Figma & 0104 & C1L1 & 
        \includegraphics[width=\linewidth,height=4cm,keepaspectratio]{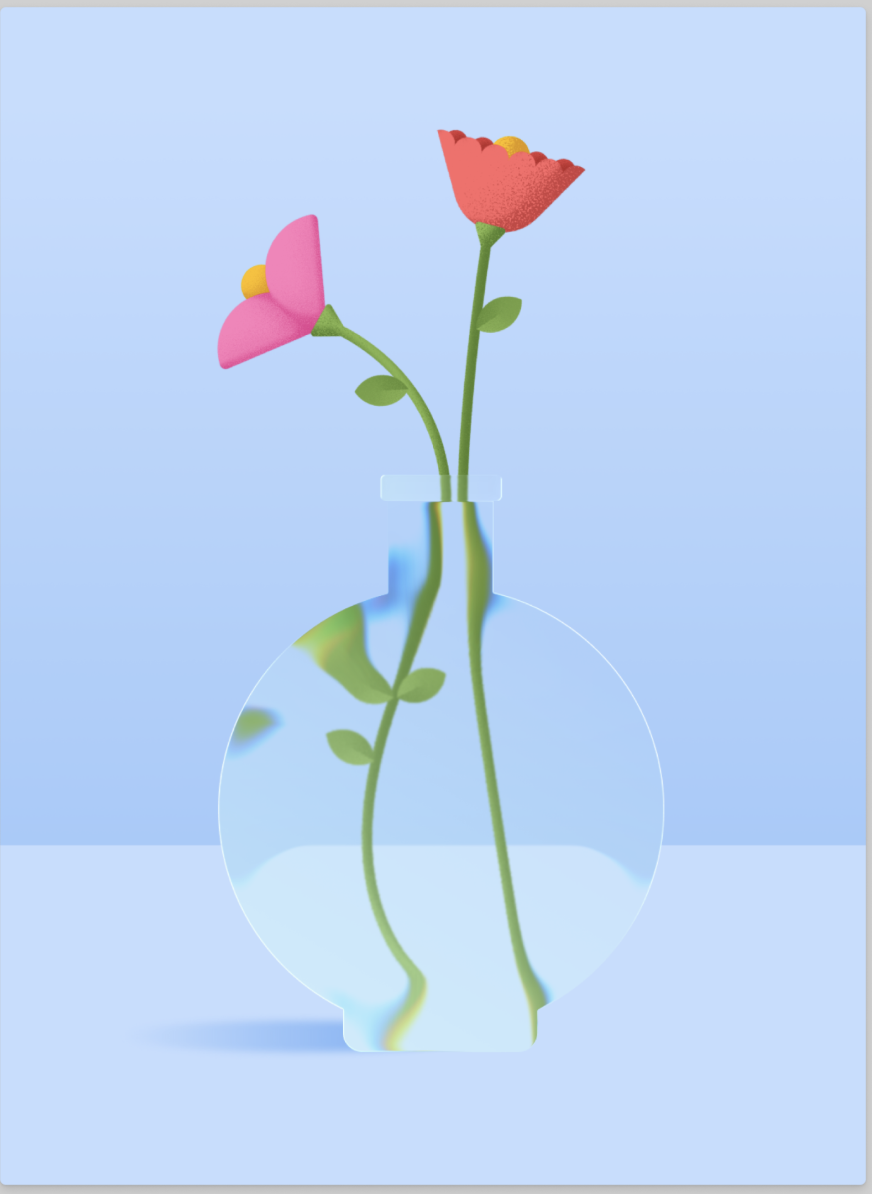} &
        \includegraphics[width=\linewidth,height=4cm,keepaspectratio]{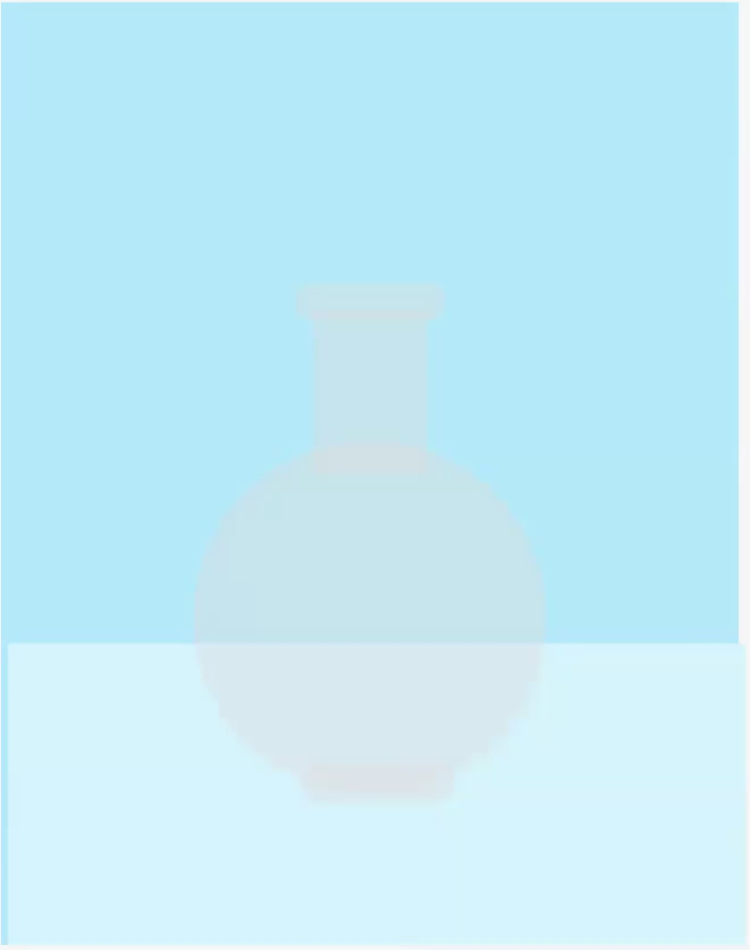} &
        \includegraphics[width=\linewidth,height=4cm,keepaspectratio]{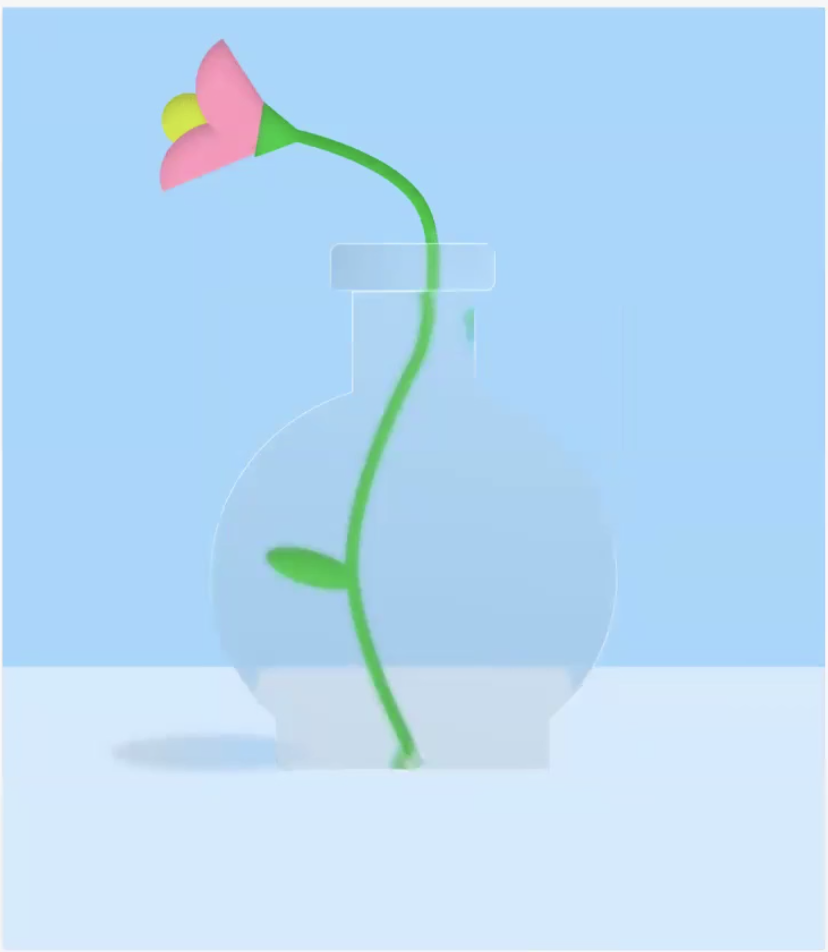} \\
        
        \midrule
        
        Blender & 0202 & C5L5 & 
        \includegraphics[width=\linewidth,height=4cm,keepaspectratio]{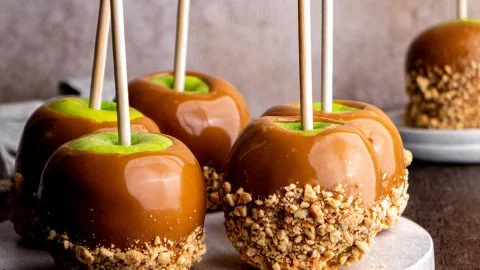} &
        \includegraphics[width=\linewidth,height=4cm,keepaspectratio]{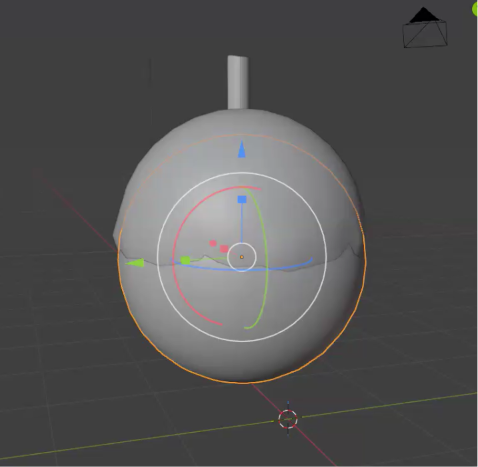} &
        \includegraphics[width=\linewidth,height=4cm,keepaspectratio]{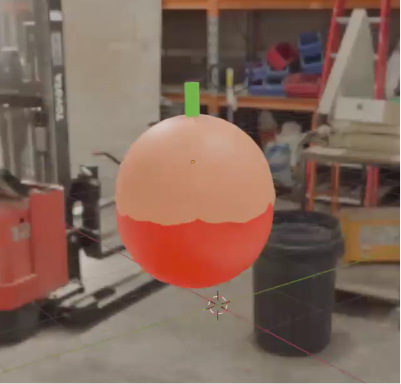} \\
        
        Blender & 0202 & C7L7 & 
        \includegraphics[width=\linewidth,height=4cm,keepaspectratio]{figures/learning_outcome/0202_target.png} &
        \includegraphics[width=\linewidth,height=4cm,keepaspectratio]{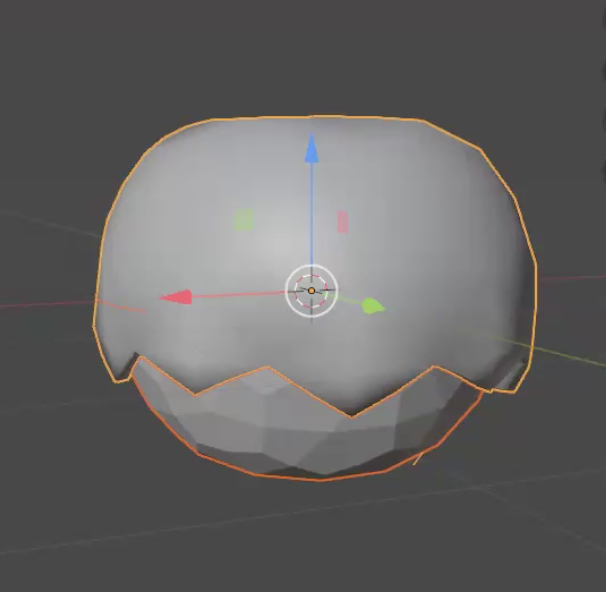} &
        \includegraphics[width=\linewidth,height=4cm,keepaspectratio]{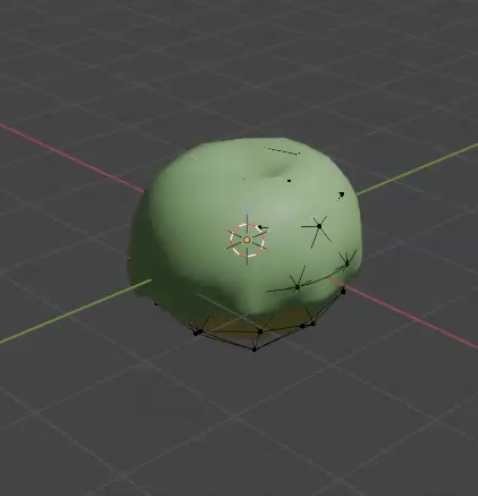} \\
        
        \midrule
        
        OnShape & 0302 & C8L8 & \includegraphics[width=\linewidth,height=4cm,keepaspectratio]{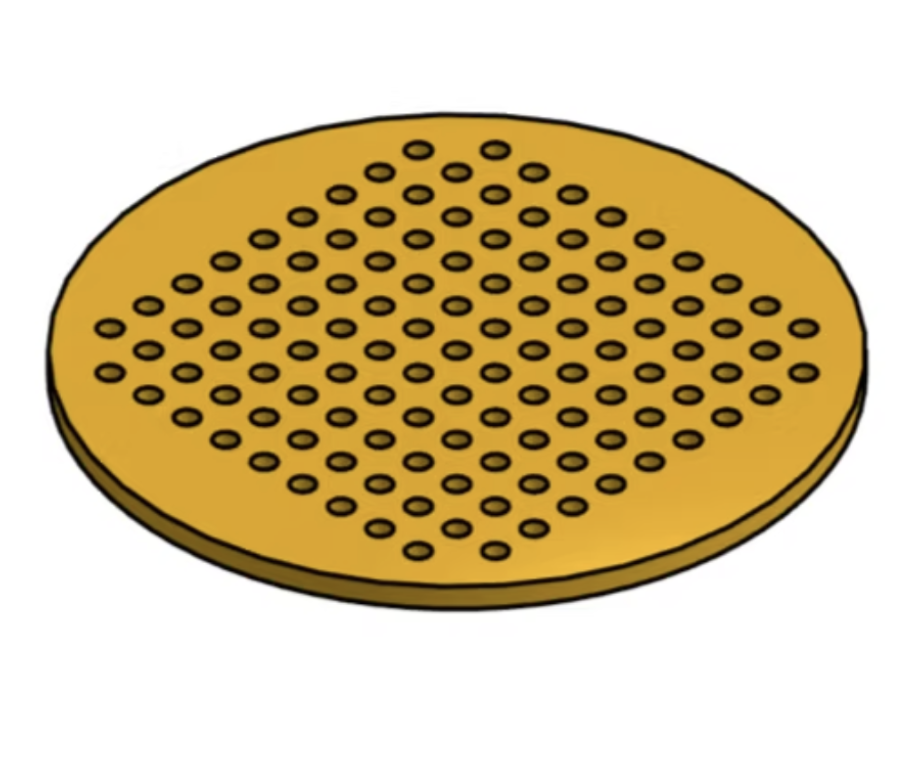} &
        \includegraphics[width=\linewidth,height=4cm,keepaspectratio]{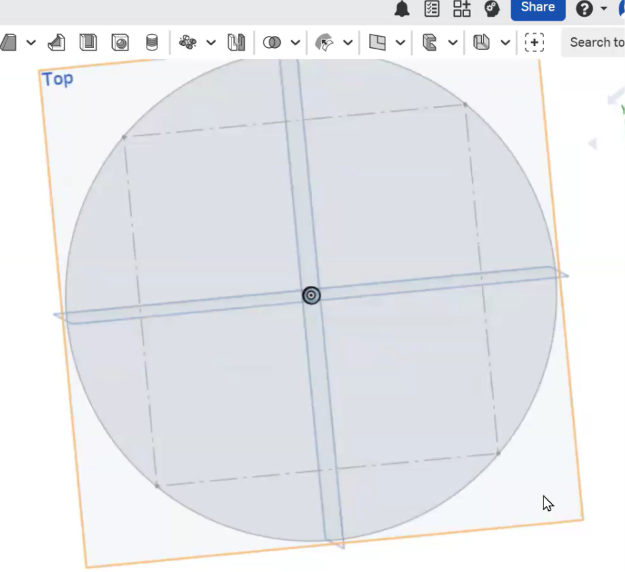} &
        \includegraphics[width=\linewidth,height=4cm,keepaspectratio]{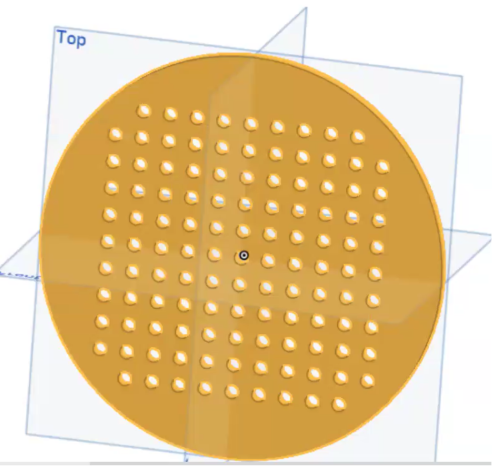} \\
        
        \bottomrule
    \end{tabularx}
    
    \caption{Examples of pre-task and post-task outcomes in visual tasks (Figma, Blender, OnShape).}
    \label{tab:learning_outcomes}
\end{table*}

\subsection{Subjective Learning Outcome}
\noindent Learners reported substantially higher confidence after completing the tasks.
Averaging confidence (7-scale likert score, 7 = very confident) increased from
\(\mu_{\mathrm{pre}}=2.49, \sigma_{\mathrm{pre}}=1.03\) to
\(\mu_{\mathrm{post}}=5.90, \sigma_{\mathrm{post}}=0.90\), with an average improvement of
\(\mu_{\mathrm{diff}}=3.41, \sigma_{\mathrm{diff}}=1.04\).
A non-parametric Wilcoxon signed-rank test showed the increase is statistical significant \(W=0, p<.001\).

Coaches rated whether learners had acquired the relevant software skill for the task (7-scale likert score, 7 = the learner was fully equipped to complete similar tasks independently). Coach-rated skill increased from
\(\mu_{\mathrm{pre}}=2.88, \sigma_{\mathrm{pre}}=2.06\)
to
\(\mu_{\mathrm{post}}=6.29, \sigma_{\mathrm{post}}=1.35\),
with an average improvement of
\(\mu_{\mathrm{diff}}=3.42, \sigma_{\mathrm{diff}}=1.97\).
A Wilcoxon signed-rank test showed the increase was statistically significant
\(W=0, p<.001\).
In total, 66 of 72 task instances received higher coach-rated skill scores after coaching, 6 stayed the same, and none decreased.

\subsection{Qualitative Feedback}
\noindent Across both learners (L1-L20) and coaches (C1-C20), participants described effective computer use coaching an activity beyond simply giving step-by-step answers. It was a balance between procedural and conceptual guidance, guided practice, and learner agency.

 Learners emphasized that practice was necessary for retention, especially after a dense coaching session. L10 noted that ``more practice would be helpful to re-enforce the things I learnt,'' because they were unsure ``for how long I will retain the knowledge if I am not practicing it regularly.'' Coaches echoed this point: C2 argued that ``more practice with new tools would help with retention,'' especially when learners first observe an effect and then reproduce it themselves with less guidance. This suggests that an effective digital coach should not only demonstrate procedures, but also create repeated opportunities for learners to apply newly learned operations.

Participants also emphasized that direct coaching is useful when learners lack basic orientation or do not know where to begin. L8 stated, ``I want direct teaching when I lack the basic knowledge of the software,'' while L20 found ``detailed step-by-step'' guidance helpful because each action was ``illustrated in a digestible way.'' Coaches similarly noted that in unfamiliar interfaces, direct pointing can reduce unnecessary friction. C13 explained that in Figma, many functions are difficult to find without prior familiarity, so ``it makes more sense to have those directly pointed out.'' However, participants also viewed direct instruction as only one part of effective teaching. L9 described a trade-off: direct answers are useful for ``simple procedural questions,'' but for design-related or conceptual tasks, they preferred ``guided exploration and explanation.'' Thus, digital coaches may need to dynamically shift between directive support and exploratory guidance depending on whether the learner’s immediate barrier is procedural, conceptual, or creative.

Another strong theme was that learners valued explanations of concepts, rationales, and tool functions, not only instructions about where to click. L17 said the coach first explained ``the function of a component'' and then showed ``how to operate,'' which strengthened both understanding and memory. C2 similarly argued that explanations and analogies help students understand ``what specific numbers or functions do, instead of just clicking mindlessly through the steps.'' Learners particularly valued explanations that made knowledge transferable. L10 appreciated learning ``strategies and thought processes'' in CAD that could transfer to other tasks. L8 described how understanding the difference between ``extrude vs. revolve'' helped them narrow down tool choices in a later post-task. L9 similarly said effective moments focused not only on how to complete the task, but also on why a certain approach works, making the knowledge transferable.

Participants also highlighted the importance of guided exploration, reflection, and learner agency. L9 described effective teaching as ``guidance that helped me think rather than directly giving the answer,'' using ``minimal but meaningful cues'' that allowed them to solve problems independently. C9 similarly described effective teaching as giving the learner ``time to explore and think about the concepts'' and then synthesize them individually. C20 preferred being able to ask learners questions rather than ``just tell them what to do,'' describing coaching as ``collaborative.'' Several learners valued moments where they tried, failed, and were corrected. L17 described ``practicing, failing, learning what is actually correct, and then doing it again'' as the best method because it forced them to think and understand what was wrong or right. L4 also found it most helpful when they tried to replicate what the coach taught, got stuck, and then received correction.

Visual grounding and screen annotation were repeatedly described as important for online computer use coaching. C12 said Zoom annotations were ``incredibly helpful'' because the coach could draw squares, arrows, and visual references without physically obstructing the screen. L13 said annotation made it easier to navigate the toolbar and find correct buttons, while L12 described screen annotations as ``very effective in visually guiding me through the activity.'' C2 described a useful pattern: first explain the concept, then annotate where to click with arrows or circles.

\section{Artifact Licenses, Terms of Use, and Intended Use}
\label{app:artifact-license-use}
\noindent Our work uses and creates several scientific artifacts: the \system{} dataset,
the software-task materials used to elicit coaching sessions, model outputs from
the evaluated multimodal language models, and annotation/evaluation code. The
\system{} dataset is intended for research on situated, multimodal
computer use coaching, human--AI collaboration, language grounding in graphical user
interfaces, and pedagogical dialogue. It is not intended for use in systems that
identify, profile, surveil, evaluate, or make consequential decisions about
individual participants.

For artifacts created by this work, we will release only de-identified research
data and accompanying documentation under a research-use license. The release
will include derived and anonymized metadata, dialogue annotations, task
metadata, model generations, and benchmark/evaluation scripts where permitted.
Raw participant-identifying information, including names, contact information,
and any uniquely identifying content observed in transcripts, screen recordings,
or files, will not be released. All transcripts are manually reviewed for
personally identifying information and redacted before release. Access to any
higher-risk modalities, such as screen recordings or file snapshots, will be
restricted to research use and subject to the dataset terms of use.

For existing artifacts, including the software applications, tutorials, APIs,
models, and evaluation packages used in this study, our use was limited to
research and evaluation. We followed the applicable licenses, API terms, and
platform policies for each artifact. We cite the creators of datasets,
benchmarks, models, and methods used in the paper where applicable. The use of
tutorial-derived task materials was limited to creating study tasks and was not
used to redistribute the original tutorials themselves. Model outputs are used
only for research evaluation of coaching behavior and are not presented as
authoritative instructions for end users.

\section{Experimental Settings and Package Parameters}
\label{app:package-settings}

\noindent All model generations used the default settings of the corresponding model
provider unless otherwise specified. We did not perform hyperparameter search
over decoding parameters. Instead, we compared prompting conditions, input
modalities, and context-window lengths as experimental factors. Specifically, we
compared vanilla, coach, and oracle prompts; text-only, image-only, and
text--visual context; and context windows of 1, 10, 30, and 60 seconds. The
default automatic-evaluation setting used text--visual input, a 30-second
context window, and the oracle prompt. The default interactive setting used
Gemini-3-Flash, the vanilla prompt, a 10-second context window, and visual
sampling at 1 frame per second.

For automatic evaluation, we used CLAIR, BLEU, METEOR, ROUGE-L, BERTScore,
Self-BLEU, Vendi Score, MAUVE, Cohen's $\kappa$, Fleiss' $\kappa$, and F1 as
reported in the paper. Unless otherwise stated, we used the default parameter
settings from the corresponding public implementations. The exact package names,
versions, and command-line arguments are provided in our released code and
configuration files. We used GPT-4.1 as the judge model for CLAIR scoring in the
automatic next-utterance evaluation.

\FloatBarrier

\section{Human Subjects, Consent, and Data Protection}
\label{app:consent}

\noindent All human data collection and evaluation protocols were reviewed and approved by
our institution's Institutional Review Board. Participants were recruited through
professional networks and Upwork. The study involved expert coaches, novice
learners, and expert evaluators. Before participating, all participants were
given study instructions describing the study purpose, the tasks they would
perform, the types of data collected, and how their data would be used for
research. Participants provided informed consent before data collection began.

The collected data included spoken dialogue, transcripts, screen recordings,
input events, task files or file snapshots, study metadata, annotations, and
pre-/post-task outcomes. Participants were informed that these data would be used
to study human expert--novice computer use coaching and to evaluate computational
models of situated coaching. Participants were also informed that released data
would be de-identified and that personally identifying information would not be
included in public artifacts. Participants could ask questions before beginning
the study and could stop participating according to the approved study protocol.

To protect participants, we manually reviewed transcripts and other releasable
materials for personally identifying information and redacted such information
before release. We do not release names, contact information, or other uniquely
identifying participant information. Because screen recordings and file snapshots
may contain incidental sensitive information, we apply additional review and
access controls to these modalities and release them only in de-identified or
restricted research-use form where permitted by the consent protocol and IRB
approval. Compensation was set based on the expected duration and complexity of
the tasks, participant expertise, and average market rates for comparable work.

The consent forms are included on the following pages.

\section{AI Use Disclosure}
\label{app:ai}

\noindent The authors used AI-based tools to assist with code generation, editing, and writing during the preparation of this paper. Specifically, AI assistance was used to help draft and revise portions of the manuscript for clarity, grammar, and organization, and to support the development, debugging, and refinement of code used in the research workflow. All AI-generated or AI-assisted content, code, analyses, and interpretations were reviewed, verified, and, where necessary, modified by the authors. The authors take full responsibility for the accuracy, integrity, originality, and final content of the paper, including any code or text developed with AI assistance.



\section*{Supplementary material}

[redacted]

**CONSENT TO PARTICIPATE IN RESEARCH**
***Multimodal Interaction in Collaborative Software Use***

**Key Information**

- You are being invited to participate in a research study. Participation in research is completely voluntary.
- The purpose of the study is to analyze language interaction and screen references in a collaborative software use.
- The study will take a total of 3 hours, and you will be asked to work with another participant to complete a 3D design creation activity (e.g., making a donut in Blender).
- Risks and/or discomforts may include potential fatigue of software using, a breach of confidentiality, and discomfort from being asked to reflect on the tasks they'll perform
- You can learn new skills of using the creative software. The results from the study may support building better collaborative and interactive systems, for a variety of tasks (e.g., UI design, video editing, 3D modeling).

## Introduction

My name is [redacted] I am planning to conduct a research study, which I invite you to take part in.

You are being invited to participate in this study because you indicated interest and availability via our recruitment form and arrived to participate in the study at the scheduled time.

## Purpose

The purpose of this study is to understand collaboration in complex software tasks. We aim to characterize how a more experienced collaborator helps a less experienced collaborator make progress during real software use while preserving the less experienced collaborator's long-term agency and independence. Based on the findings, we aim to develop and evaluate AI assistants that can collaborate with people by improving their abilities to perform tasks without assistance. We will analyze language and interactions to identify and quantify collaboration behaviors.

## Procedures

If you agree to be in this study, you will be asked to do the following:

1. Confirm that you are 18 years of age or older.
2. Complete the Media Records Release Form. If you do not agree to the uses of your data specified in this form, you will be excused from the study.
3. Complete an initial survey asking about your experience with relevant software tools and remote collaboration.

**During the main part of the study...**

At the scheduled time, you will join the study session remotely from your own location via Zoom or in an office room with a desk, chair, computer tower, keyboard, mouse, monitor, microphone, and headphones. Your partner will be placed similarly in another office room

Once both participants arrive:

1. You will complete the study while connected to your partner via Zoom. Zoom support remote collaboration by enabling a shared view of the active workspace (e.g., screen sharing or a streaming view) so your partner can follow progress and provide help in real time.
2. We will introduce the target software environment and provide brief instructions on how to set up software and Zoom for communication and collaboration.
3. We will begin recording, including audio recordings, screen recordings of the active workspace, and interaction logs (such as cursor position and keyboard/mouse events). We will not video record your face, and you will be asked to keep your camera off during the study.
4. Study personnel will join the Zoom session and will not participate in the collaboration unless technical issues occur.
5. You and your partner will complete collaborative activities using digital software. You will work together toward a specified goal (e.g., creating a 3D design in the target software) that requires communicating steps, making decisions, and resolving problems as you progress. You may communicate using voice, text, and on-screen cursor movement to provide guidance, explanations, and feedback while working toward the goal.
6. At the end of each activity, you will indicate completion to the researchers.
7. After each activity (or at the end of the session), we will ask brief questions to confirm your prior familiarity with the task content and to capture your reflections on what was difficult or helpful during the collaboration.
8. We will allow for a short break (e.g., up to 15 minutes) between activities.
9. The full session will last up to approximately 3 hours, including activities, breaks, and a short wrap-up as needed.

**Study time:** Your study participation will take a total of approximately 180 minutes (3 hours).

**Study location:** Study procedures will take place at Zoom or at [redacted]

## Benefits

You can learn new skills of using the creative software. The results from the study may support building better collaborative and interactive systems, for a variety of tasks (e.g., UI design, video editing, 3D modeling).

## Risks/Discomforts

- A possible, but unlikely, known risk from using software is experiencing eye strain like discomfort from intensive visual focus.
- Participants may feel discomfort when being asked to reflect on the task and their collaboration.
- Because software can be complex and have a steep learning curve, there is a risk that you could become fatigued when completing tasks.
- You are free to decline to answer any questions you do not wish to or to end participation at any time.

**Breach of confidentiality:** As with all research, there is a chance that confidentiality could be compromised; however, we are taking precautions to minimize this risk. Additionally, there is a special risk of breach of confidentiality with research involving group participants: once participants leave the

research setting, investigators have no control over information that participants choose to share, even if they have agreed to keep that information confidential.

## Confidentiality

Your safety and confidentiality are of utmost importance to us. We are taking the following steps to minimize the risks of a confidentiality breach:

1. Data we store may include survey responses, audio clips, and records of task activity, including screen recordings and recordings of mouse movements/keyboard presses, as well as derivative data such as transcripts. We will not collect video recordings of you (e.g., your face).
2. We may want to use portions of the audio recordings in public presentations related to the research. Please read the Media Records Release Form, which outlines several possible uses of the recordings and asks for your specific consent for each use. We will not use any recordings in future presentations without your consent. You can revoke consent and request deletion of this data at any time.
3. Your data will be stored on password-protected computers accessible only by the research team, and encrypted in the cloud (e.g., [redacted]).
4. Audio data will be stored in an encrypted format and will only be decrypted for analysis purposes.
5. After data is shared within the research team, we will delete the original copies on the machines used for recording.
6. Audio recordings and study data may be retained indefinitely on [redacted] for possible use in future research by the study team or others. Upon completion of the study, we will delete any local copies of the audio recordings stored on the computers of individual study personnel.
7. Audio recordings will be transcribed. The recordings will not be destroyed after transcription. We will remove identifiable information from the transcripts to the extent possible.
8. We will assign you a unique ID that will be associated with all files produced during the session. Other data we collect (e.g., screen recordings and interaction logs) and resulting analyses (e.g., annotated transcripts) will be associated only with the unique ID and will not be linkable to you without access to identifying information. Identifiers will be removed from the survey and transcripts to the extent possible.
9. Identifiers may be removed from identifiable private information. After such removal, the information may be used for future research studies or shared with other investigators for future research studies without additional informed consent.
10. This informed consent form, media release, and questionnaire may be printed and completed by pen. Hard copies will be stored in a locked filing cabinet in the PI's office.
11. We will keep information provided during the group session as confidential as possible and strongly encourage all participants to do likewise. However, we cannot guarantee that other participants will keep information confidential after the session is over. Please do not share anything during the session that you do not wish to be made public.

Your personal information may be released if required by law. Authorized representatives from the following organizations may review your research data for purposes such as monitoring or managing the conduct of this study:

[redacted]

## Compensation

In return for your time and effort, you will receive an Amazon gift card for taking part in this 3-hour study. The amount will be determined by years of relevant professional or instruction experiences. For professional experiences, we count the years when you hold a paid role (e.g., full-time designer,

contractor) where you regularly use the relevant software (e.g., Figma, Blender, Adobe Premiere, Maya) in their work). For instruction experience, we count the years when serve as an instructor or coach for the relevant software (e.g., Figma, Blender, Adobe Premiere, Maya).

- If you have <10 years relevant experiences, you will receive $20/hour.
- If you have 10-20 years relevant experiences, you will receive $40/hour.
- If you have 20+ years relevant experiences, you will receive $60/hour.

If you end the study early, compensation will be prorated based on the number of full hours completed, at $20/hour (e.g., 1.5 hours is compensated as 1 hour = $20).

## Rights

***Participation in research is completely voluntary***. You have the right to decline to participate or to withdraw at any point in this study without penalty or loss of benefits to which you are otherwise entitled.

## Questions

If you have any questions or concerns about this study, you may contact [redacted] at [redacted]

If you have any questions or concerns about your rights and treatment as a research subject, you may contact [redacted] or [redacted]

***************************************************

## CONSENT

Consenting to participate in this study will result in data collection. How your data will be used and released is contained in the Media Records Release Form.

You will be given a copy of this consent form to keep.

If you wish to participate in this study, please sign and date below.

________________________________ ______________
Participant's Name *(please print)* Date

________________________________ ______________
Participant's Signature Date

________________________________ ______________
Person Obtaining Consent Date

[redacted]

**CONSENT TO PARTICIPATE IN RESEARCH**
***Multimodal Interaction in Collaborative Software Use***

**Key Information**

- You are being invited to participate in a research study. Participation in research is completely voluntary.
- The purpose of the study is to analyze language interaction and screen references in a collaborative software use.
- The study will take a total of 3 hours, and you will be asked to work with another participant to complete a 3D design creation activity (e.g., making a donut in Blender).
- Risks and/or discomforts may include potential fatigue of software using, a breach of confidentiality, and discomfort from being asked to reflect on the tasks they'll perform
- You can learn new skills of using the creative software. The results from the study may support building better collaborative and interactive systems, for a variety of tasks (e.g., UI design, video editing, 3D modeling).

## Introduction

My name is [redacted]. I am planning to conduct a research study, which I invite you to take part in.

You are being invited to participate in this study because you indicated interest and availability via our recruitment form and arrived to participate in the study at the scheduled time.

## Purpose

The purpose of this study is to understand collaboration in complex software tasks. We aim to evaluate how current AI agents can help a less experienced user make progress during real software use while preserving the less experienced user's long-term agency and independence. Based on the findings, we aim to develop and evaluate AI assistants that can collaborate with people by improving their abilities to perform tasks without assistance. We will analyze language and interactions to identify and quantify collaboration behaviors.

## Procedures

If you agree to be in this study, you will be asked to do the following:

1. Confirm that you are 18 years of age or older.
2. Complete the Media Records Release Form. If you do not agree to the uses of your data specified in this form, you will be excused from the study.
3. Complete an initial survey asking about your experience with relevant software tools and remote collaboration.

**During the main part of the study...**

At the scheduled time, you will join the study session remotely from your own location via Zoom or in an office room with a desk, chair, computer tower, keyboard, mouse, monitor, microphone, and headphones.

Once you are ready to begin:

1. You will complete the study via Zoom. Zoom support remote collaboration by enabling a shared view of the active workspace (e.g., screen sharing or a streaming view) so the AI agent can follow progress and provide help in real time.
2. We will introduce the target software environment and provide brief instructions on how to set up software and Zoom for communication and collaboration.
3. We will begin recording, including audio recordings, screen recordings of the active workspace, and interaction logs (such as cursor position and keyboard/mouse events). We will not video record your face, and you will be asked to keep your camera off during the study.
4. Study personnel will join the Zoom session and will not participate in the collaboration unless technical issues occur.
5. You and the AI agent will complete collaborative activities using digital software. You will work together toward a specified goal (e.g., creating an 3D design in the target software) that requires communicating steps, making decisions, and resolving problems as you progress. You may communicate using voice, text, and on-screen cursor movement to provide guidance, explanations, and feedback while working toward the goal.
6. At the end of each activity, you will indicate completion to the researchers.
7. After each activity (or at the end of the session), we will ask brief questions to confirm your prior familiarity with the task content and to capture your reflections on what was difficult or helpful during the collaboration.
8. We will allow for a short break (e.g., up to 15 minutes) between activities.
9. The full session will last up to approximately 3 hours, including activities, breaks, and a short wrap-up as needed.

**Study time:** Your study participation will take a total of approximately 180 minutes (3 hours).

**Study location:** Study procedures will take place at Zoom [redacted]

### Benefits

You can learn new skills of using the creative software. The results from the study may support building better collaborative and interactive systems, for a variety of tasks (e.g., UI design, video editing, 3D modeling).

### Risks/Discomforts

- A possible, but unlikely, known risk from using software is experiencing eye strain like discomfort from intensive visual focus.
- You may feel discomfort when being asked to reflect on the task and their collaboration.
- Because software can be complex and have a steep learning curve, there is a risk that you could become fatigued when completing tasks.
- You are free to decline to answer any questions you do not wish to or to end participation at any time.

**Breach of confidentiality:** As with all research, there is a chance that confidentiality could be compromised; however, we are taking precautions to minimize this risk.

### Confidentiality

Your safety and confidentiality are of utmost importance to us. We are taking the following steps to minimize the risks of a confidentiality breach:

1. Data we store may include survey responses, audio clips, and records of task activity, including screen recordings and recordings of mouse movements/keyboard presses, as well as derivative data such as transcripts. We will not collect video recordings of you (e.g., your face).
2. We may want to use portions of the audio recordings in public presentations related to the research. Please read the Media Records Release Form, which outlines several possible uses of the recordings and asks for your specific consent for each use. We will not use any recordings in future presentations without your consent. You can revoke consent and request deletion of this data at any time.
3. Your data will be stored on password-protected computers accessible only by the research team, and encrypted in the cloud (e.g., [redacted]).
4. Audio data will be stored in an encrypted format and will only be decrypted for analysis purposes.
5. After data is shared within the research team, we will delete the original copies on the machines used for recording.
6. Audio recordings and study data may be retained indefinitely on [redacted] for possible use in future research by the study team or others. Upon completion of the study, we will delete any local copies of the audio recordings stored on the computers of individual study personnel.
7. Audio recordings will be transcribed. The recordings will not be destroyed after transcription. We will remove identifiable information from the transcripts to the extent possible.
8. We will assign you a unique ID that will be associated with all files produced during the session. Other data we collect (e.g., screen recordings and interaction logs) and resulting analyses (e.g., annotated transcripts) will be associated only with the unique ID and will not be linkable to you without access to identifying information. Identifiers will be removed from the survey and transcripts to the extent possible.
9. Identifiers may be removed from identifiable private information. After such removal, the information may be used for future research studies or shared with other investigators for future research studies without additional informed consent.
10. This informed consent form, media release, and questionnaire may be printed and completed by pen. Hard copies will be stored in a locked filing cabinet in the PI's office.

Your personal information may be released if required by law. Authorized representatives from the following organizations may review your research data for purposes such as monitoring or managing the conduct of this study:

[redacted] [redacted]

## Compensation

In return for your time and effort, you will receive an Amazon gift card for taking part in this 3-hour study. The amount will be determined by years of relevant professional or instruction experiences. For professional experiences, we count the years when you hold a paid role (e.g., full-time designer, contractor) where you regularly use the relevant software (e.g., Figma, Blender, Adobe Premiere, Maya) in their work). For instruction experience, we count the years when serve as an instructor or coach for the relevant software (e.g., Figma, Blender, Adobe Premiere, Maya).

- If you have <10 years relevant experiences, you will receive $20/hour.
- If you have 10-20 years relevant experiences, you will receive $40/hour.
- If you have 20+ years relevant experiences, you will receive $60/hour.

If you end the study early, compensation will be prorated based on the number of full hours completed, at $20/hour (e.g., 1.5 hours is compensated as 1 hour = $20).

## Rights

***Participation in research is completely voluntary***. You have the right to decline to participate or to withdraw at any point in this study without penalty or loss of benefits to which you are otherwise entitled.

## Questions

If you have any questions or concerns about this study, you may contact [redacted]

If you have any questions or concerns about your rights and treatment as a research subject, you may contact [redacted]

**************************************************

## CONSENT

Consenting to participate in this study will result in data collection. How your data will be used and released is contained in the Media Records Release Form.

You will be given a copy of this consent form to keep.

If you wish to participate in this study, please sign and date below.

______________________________ ____________
Participant's Name *(please print)* Date

______________________________ ____________
Participant's Signature Date

______________________________ ____________
Person Obtaining Consent Date

[redacted]

## Research Media Records Release Form

As part of this study, we will store audio recordings of your voice, screen recordings of the software use, and recordings of keyboard presses and cursor movements you make during gameplay, hereby referred to as "records". Below we describe the intended usages of these records. We will only use the records in the following usages. In any use of these records, your name will not be identified.

1. The records can be studied by the research team for use in the research project including the application of automatic transcription software to voice recordings.

2. The records can be shown to subjects in other experiments.

3. The records can be used for scientific publications.

4. The records can be shown at meetings of scientists interested in topics including natural language processing, human-computer interaction, machine learning, linguistics, computer vision, cognitive science and education.

5. The records can be shown in classrooms to students.

6. The records can be shown in public presentations to non-scientific groups.

7. Transcriptions of audio recordings including timestamps, screen recordings of software use, and recordings of keyboard presses and cursor movements can be released in a publicly available dataset (Dataset A).

8. Audio recordings can be released in a dataset (Dataset B) to research team personnel within and outside of the [redacted] to study to features of speech in collaborative interactions that can be only analyzed through audio recordings (e.g., prosody). Specifically, we will allow researchers access to Dataset B after they agree to two conditions: (a) to not share any parts of Dataset B with any other individual, and (b) to delete any audio recordings from their copy of Dataset B if requested by us.

I have read this form and give my consent for use of the records as indicated above.

Name_________________ Signature ______________________ Date _________________



\end{document}